\documentclass[12pt]{article}
\usepackage{amsmath,amssymb,amsthm,amsfonts,bm,mathrsfs,float,algorithm,algorithmic,caption,subcaption,setspace,relsize,xr}
\usepackage{avant,array,geometry,fontenc,inputenc,natbib,hyperref,multirow,enumerate,rotating,booktabs,color,latexsym,times,stmaryrd,xcolor}

\newtheorem{assump}{Assumption}
\newtheorem{lemma}{Lemma}
\newtheorem{theorem}{Theorem}

\newtheorem{example}{Example}

\def\eop{\hfill {\large $\Box$}}
\newcommand{\Mean}{{\mathbb{E}}}

\newcommand{\prob}{{\mathbb{P}}}

\DeclareMathOperator*{\argmax}{arg\,max}

\def\floor#1{\lfloor #1 \rfloor}
\newcommand{\independent}{\;\, \rule[0em]{.03em}{.67em} \hspace{-.27em}
	\rule[-.02em]{.7em}{.03em} \hspace{-.27em}
	\rule[0em]{.03em}{.67em}\;\,}

\newcommand{\blind}{1}

\makeatletter
\newcommand*{\addFileDependency}[1]{
  \typeout{(#1)}
  \@addtofilelist{#1}
  \IfFileExists{#1}{}{\typeout{No file #1.}}
}
\makeatother

\newcommand{\change}[1]{{\leavevmode\color{black}{#1}}}

\begin{document}

\if1\blind
{
\title{\Large{\textbf{Testing for the Markov Property in Time Series \\
via Deep Conditional Generative Learning}}}
\author{
\bigskip
Yunzhe Zhou$^\ddag$, Chengchun Shi$^\dag$, Lexin Li$^\ddag$, and Qiwei Yao $^\dag$\\
\normalsize{\textit{$^\dag$London School of Economics and Political Science, London, UK}} \\
\normalsize{\textit{$^\ddag$University of California at Berkeley, California, USA}}
}
\date{}
\maketitle
} \fi

\if0\blind
{
\title{\Large{\textbf{Testing for the Markov Property in Time Series \\
via Deep Conditional Generative Learning}}}
\author{
\bigskip
\vspace{0.5in}
}
\date{}
\maketitle
} \fi

\baselineskip=20pt
\begin{abstract}
The Markov property is widely imposed in analysis of time series data. Correspondingly, testing the Markov property, and relatedly, inferring the order of a Markov model, are of paramount importance. In this article, we propose a nonparametric test for the Markov property in high-dimensional time series via deep conditional generative learning. We also apply the test sequentially to determine the order of the Markov model. We show that the test controls the type-I error asymptotically, and has the power approaching one. Our proposal makes novel contributions in several ways. We utilize and extend state-of-the-art deep generative learning to estimate the conditional density functions, and establish a sharp upper bound on the approximation error of the estimators. We derive a doubly robust test statistic, which employs a nonparametric estimation but achieves a parametric convergence rate. We further adopt sample splitting and cross-fitting to minimize the conditions required to ensure the consistency of the test. We demonstrate the efficacy of the test through both simulations and the three data applications. 
\end{abstract}


\newpage
\baselineskip=22pt

\section{Introduction}
\label{sec:intro}

The Markov property is fundamental and is commonly imposed in time series analysis. For instance, in economics and reinforcement learning, the Markov property is the foundation of the Markov decision process that provides a general framework for modeling sequential decision making. In finance and marketing, the Markov property is widely assumed in most continuous time modeling. See \citet{Chen2012} for a review. Correspondingly, testing the Markov property, and relatedly, inferring the order of a Markov model, are of paramount importance in a broad range of applications. 

Such a testing problem, however, is highly nontrivial and poses many challenges, especially for high-dimensional time series. For the Markov property test, \cite{ait1996} proposed a nonparametric test based on the Chapman-Kolmogorov equation and smoothing kernels. \cite{Chen2012} tackled the testing problem based on the conditional characteristic function (CCF) estimated by local polynomial regressions. However, kernel smoothers, including local polynomial regressions, suffer from a poor estimation accuracy in moderate to high-dimensional settings, leading to an inflated type-I error or a low power for the tests. For the order determination in nonparametric autoregression, \citet{cheng1992, yao1994, vieu1994} developed some cross-validation based methods, and \citet{auestad1990, tschernig2000} proposed a final prediction error based criterion. 
But none of those order determination methods are based on hypothesis testing, and they all assume the dimension of the time series is fixed. More recently, \cite{Shi2020} developed a quantile random forest algorithm and a doubly robust procedure to test the Markov assumption in the context of reinforcement learning. But their method, as we show later in Section \ref{sec:sim}, would fail to control the type-I error in the time series setting.  

In this article, we propose a nonparametric testing procedure for the Markov property in high-dimensional time series via deep conditional generative learning. The proposed test can be sequentially applied for order selection of the Markov model as well. Our proposal makes unique and useful contributions in several ways. 

Particularly, we utilize some state-of-the-art deep conditional generative learning methods to address a classical yet challenging statistical inference problem in time series analysis. Deep conditional generative models include mixture density networks \citep{mdn}, conditional generative adversarial networks \citep{mirza2014conditional}, conditional variational autoencoders \citep{sohn2015learning}, and normalizing flow models \citep{kobyzev2020normalizing}. They provide a powerful set of tools to flexibly learn conditional probability distributions, and have been used in numerous applications, such as computer vision, imaging processing, and artificial intelligence \citep{yan2016attribute2image, shu2017bottleneck, wang2018cvpr, jo2021srflow}. Nevertheless, these tools are much less used and studied in the statistics literature. We employ this family of models to learn highly complex conditional distributions in a nonparametric fashion, and demonstrate their advantages over the more traditional kernel smoothers including local polynomial regressions, especially in a high-dimensional setting. 

Meanwhile, it is far from a simple application of some ready-to-use deep learning tools, but instead it requires both crucial modification of the methods and careful characterization of their theoretical properties. We build our testing procedure based upon mixture density networks \citep[MDN]{mdn}, combined with several crucial new components. First, we propose a new MDN architecture to model the conditional distribution of a multivariate response. Based on such an architecture, we learn two distributional generators, a forward generator and a backward generator, then properly integrate the two generators to construct the test statistic. Second, we derive the convergence rate of the MDN estimator in Theorem \ref{mdn_bound} , which is crucial to establish the consistency of our proposed test, but is not currently available in the MDN literature. In particular, we provide a sharp upper bound on the approximation error of MDN in Lemma \ref{lemma1} when the underlying conditional density function follows an infinite conditional Gaussian mixture model. We remark that, although it is possible to obtain a bound by directly applying Lemma 1 of \cite{barron1993universal}, it would only yield a very loose bound; see Section \ref{sec:mdn-bound} for more details. To our knowledge, we are among the first to systematically study the error bound of MDN, and our results are useful for the general theory of deep (generative) learning methods \citep[see e.g.,][]{farrell2019deepneural,liang2018well, Zhou2021deep,chen2020computation,zhou2022deep}. Third, we show the proposed test controls the type-I error in Theorem \ref{thm4}, and has the power approaching one in Theorem \ref{thm5}. We show that our test statistic achieves a parametric convergence rate and a parametric power guarantee while its components are estimated nonparametrically. This is made possible because the way in which we combine the two distribution generators yields a doubly robust estimator of the test statistic \citep{tsiatis2007semiparametric}. Thanks to this double robustness, the bias of our test statistic estimator decays to zero faster than the rate of the individual nonparametric distribution generator. Finally, to avoid the requirement of certain metric entropy conditions for the distribution generator estimators \change{\citep[Equation (1.6)]{chernozhukov2018double}}, we further employ the sample splitting and cross-fitting strategy \citep{romano2019multiple} to ensure the size control of the test.  

The rest of the article is organized as follows.  We formulate the hypotheses and propose a doubly robust test statistic in Section \ref{sec:formulation}. We develop the corresponding test, as well as a forward sequential procedure for order determination in Section \ref{sec:test}. We establish the theoretical guarantees in Section \ref{sec:theory}. We carry out simulations in Section \ref{sec:sim}, and illustrate with three real datasets in Section \ref{sec:real}. We relegate all technical proofs to the Supplementary Appendix.

\section{Hypotheses and Test Statistic}
\label{sec:formulation}

\subsection{Hypotheses}
\label{sec:hypotheses}

We first formulate the hypotheses of interest. Consider a strictly stationary $d$-dimensional time series, $X_t = (X_{t,1},X_{t,2},...,X_{t,d})^\top$, $ t\ge 1 $. We target the following pair of hypotheses:
\begin{align} \label{eqn:hypo}
\begin{split}
& H_0: \prob(X_{t+1} \leq x | I_t) = \prob(X_{t+1} \leq x | X_t), \; \textrm{ almost surely for all } x \in \mathbb{R}^d \textrm{ and } t >0; \\
& H_A: \prob(X_{t+1} \leq x | I_t) \neq \prob(X_{t+1} \leq x | X_t), \; \textrm{ for some } x \in \mathbb{R}^d \textrm{ and } t >0,
\end{split}
\end{align}
where $I_t$ denotes the data history $\{X_t,X_{t-1},...\}$. The Markov property holds under $H_0$. Intuitively, this property requires the past and future values to be independent, conditionally on the present. To test $H_0$, it suffices to test a sequence of conditional independences
\begin{eqnarray}\label{CIA}
\change{X_{t+q} \independent \{X_{j}\}_{t \leq j < t+q-1} \ | X_{t+q-1},}
\end{eqnarray}
for any time $t > 0$ and any lag $q\ge 2$, where $\independent$ denotes the conditional independence. 

We next characterize the conditional independence using the conditional characteristic function (CCF). A similar result is given in \change{\cite[Equation (2.6)]{Chen2012}}. For any vector $\mu \in \mathbb{R}^d$ of the same dimension as $X_t$, define the CCF of $X_{t+1}$ given $X_t$ as
\begin{align*}
\varphi^*(\mu|x)=\Mean \left\{ \exp(i \mu^\top X_{t+1})|X_{t}=x \right\}.
\end{align*}

\begin{theorem}\label{ccf_equ}
The conditional independence \eqref{CIA} holds if and only if 
\begin{eqnarray} \label{CIA2}
\change{\varphi^*(\mu|X_{t+q-1})\Mean [\exp(i\nu^\top X_{t})|\{X_j\}_{t<j<t+q}] = \Mean \left[ \exp(i\mu^\top X_{t+q}+i\nu^\top X_{t})|\{X_j\}_{t<j<t+q}\right]}
\end{eqnarray}
almost surely, for any $t>0$, $q\ge 2$, and $\mu, \nu \in \mathbb{R}^d$. 
\end{theorem}

\subsection{Doubly robust test statistic}
\label{sec:test-stat}

Theorem \ref{ccf_equ} suggests a possible test for the hypotheses in \eqref{eqn:hypo}. That is, under $H_0$, taking another expectation on both sides of \eqref{CIA2}, we obtain that
\begin{eqnarray*} 
\change{\Mean \Big[\left\{ \exp(i\mu^\top X_{t+q})-\varphi^*(\mu|X_{t+q-1}) \right\} \exp(i\nu^\top X_{t})\Big]=0,}
\end{eqnarray*}
for any $t,q,\mu,\nu$. This suggests the following test statistic,
\begin{eqnarray}\label{eqn:Stilde.qmunu}
\widetilde{S}(q, \mu, \nu) = \frac{1}{T-q} \sum_{t=1}^{T-q} \left\{ \exp(i\mu^\top X_{t+q})-  \widehat{\varphi}(\mu|X_{t+q-1}) \right\} \left\{ \exp(i\nu^\top X_{t})-\bar{\varphi}(\nu) \right\},
\end{eqnarray}
where $\widehat{\varphi}$ denotes some estimator of the CCF $\varphi^*$, and $\bar{\varphi}(\nu)=T^{-1}\sum_{1\le t\le T} \exp(i\nu^\top X_{t})$. Aggregating $\widetilde{S}(q, \mu, \nu)$ over different combinations of $(q,\mu,\nu)$ yields the test statistic proposed in \change{\citet[Equation (2.18)]{Chen2012}}.

Computing \eqref{eqn:Stilde.qmunu} requires a suitable estimator $\widehat{\varphi}$ for $\varphi^*$. \cite{Chen2012} proposed to use the local polynomial regression to estimate $\varphi^*$. However, the local polynomial regression tends to perform poorly when the dimension $d$ of $X_t$ increases \citep{taylor2013challenging}, and the corresponding test would fail to be consistent. More recently, deep conditional generative learning models have demonstrated an exceptional capacity of estimating complex conditional distributions \citep[e.g.,][]{sohn2015learning, kobyzev2020normalizing}. These tools can be potentially employed to estimate $\prob_{X_t|X_{t-1}}$, and subsequently the CCF $\varphi^*$. However, naively plugging in a deep conditional generative learning estimator for $\varphi^*$ would induce a heavy bias in  \eqref{eqn:Stilde.qmunu}, which would fail to guarantee a tractable limiting distribution for the test statistic. 

To address this issue, we propose to construct a doubly robust test statistic. Specifically, for any vector $\nu \in \mathbb{R}^d$ of the same dimension as $X_t$, define the CCF of $X_{t}$ given $X_{t+1}$ as
\begin{align*}
\psi^*(\nu | x) = \Mean \left\{ \exp(i\nu^\top X_{t})|X_{t+1}=x \right\}. 
\end{align*}
We introduce a doubly robust estimating equation in the next theorem.   

\begin{theorem} \label{thm2}
Under $H_0$, for any $t\ge 0$, $q\ge 2$, $\mu, \nu \in \mathbb{R}^{d}$, we have
\begin{eqnarray} \label{eqn:dr-ee}
\Mean \left\{ \exp(i\mu^\top X_{t+q})-\varphi^*(\mu|X_{t+q-1}) \right\} \left\{ \exp(i\nu^\top X_{t})-\psi^*(\nu|X_{t+1}) \right\} = 0.
\end{eqnarray}
In addition, \eqref{eqn:dr-ee} is doubly-robust, in that, for any CCFs $\varphi$ and $\psi$, as long as either $\varphi=\varphi^*$, or $\psi=\psi^*$, we have that $\Mean \{\exp(i\mu^\top X_{t+q})-\varphi(\mu|X_{t+q-1}) \} \{\exp(i\nu^\top X_{t})-\psi(\nu|X_{t+1}) \} = 0$. 
\end{theorem}

Motivated by \eqref{eqn:dr-ee}, we propose the following test statistic,
\begin{eqnarray}\label{eqn:S.qmunu}
S(q, \mu, \nu) = \frac{1}{T-q} \sum_{t=1}^{T-q} \{\exp(i\mu^\top X_{t+q}) - \widehat{\varphi}(\mu|X_{t+q-1}) \}  \{\exp(i\nu^\top X_{t})-\widehat{\psi}(\nu|X_{t+1})\},
\end{eqnarray}
where $\widehat{\varphi}$ and $\widehat{\psi}$ denote some estimators of $\varphi^*$ and $\psi^*$, respectively. This statistic, as suggested by Theorem \ref{thm2}, is doubly robust. A key advantage is that the bias of this test statistic can decay to zero at a faster rate than the convergence rate of the individual estimator $\widehat{\varphi}$ and $\widehat{\psi}$. By contrast, the bias of the test statistic in \eqref{eqn:Stilde.qmunu} has the same order of magnitude as that of $\widehat{\varphi}$; see Theorem \ref{thm35}. This double robustness property thus enables us to employ some highly flexible nonparametric estimators for $\varphi^*$ and $\psi^*$. In the next section, we extend mixture density networks \citep{mdn} to estimate the CCFs, and develop the corresponding testing procedure.

\section{Testing Procedure}
\label{sec:test}

\subsection{Mixture density networks} 
\label{mdn_detail}

The mixture density network is a classical deep generative model that combines the Gaussian mixture model with deep neural networks \citep{mdn}, and has shown promising performance in conditional density estimation \citep{koohababni2018nuclei, rothfuss2019conditional}. In effect it integrates the universal approximation property of the Gaussian mixture model to approximate any smooth density function \citep{nguyen2019approximations}, with the capacity of deep neural networks (DNNs) to approximate \change{both smooth and non-smooth conditional mean and variance functions in high dimension. See Assumption 2(iii) for the class of smooth functions, and \citet{imaizumi2019deep} for the class of non-smooth functions that can be well approximated by DNNs.} Next, we first introduce the standard MDN model, then propose a new MDN architecture to model the conditional distribution of a multivariate response.

\begin{figure}[t]
\centering
\includegraphics[width=11.5cm, height=7.5cm]{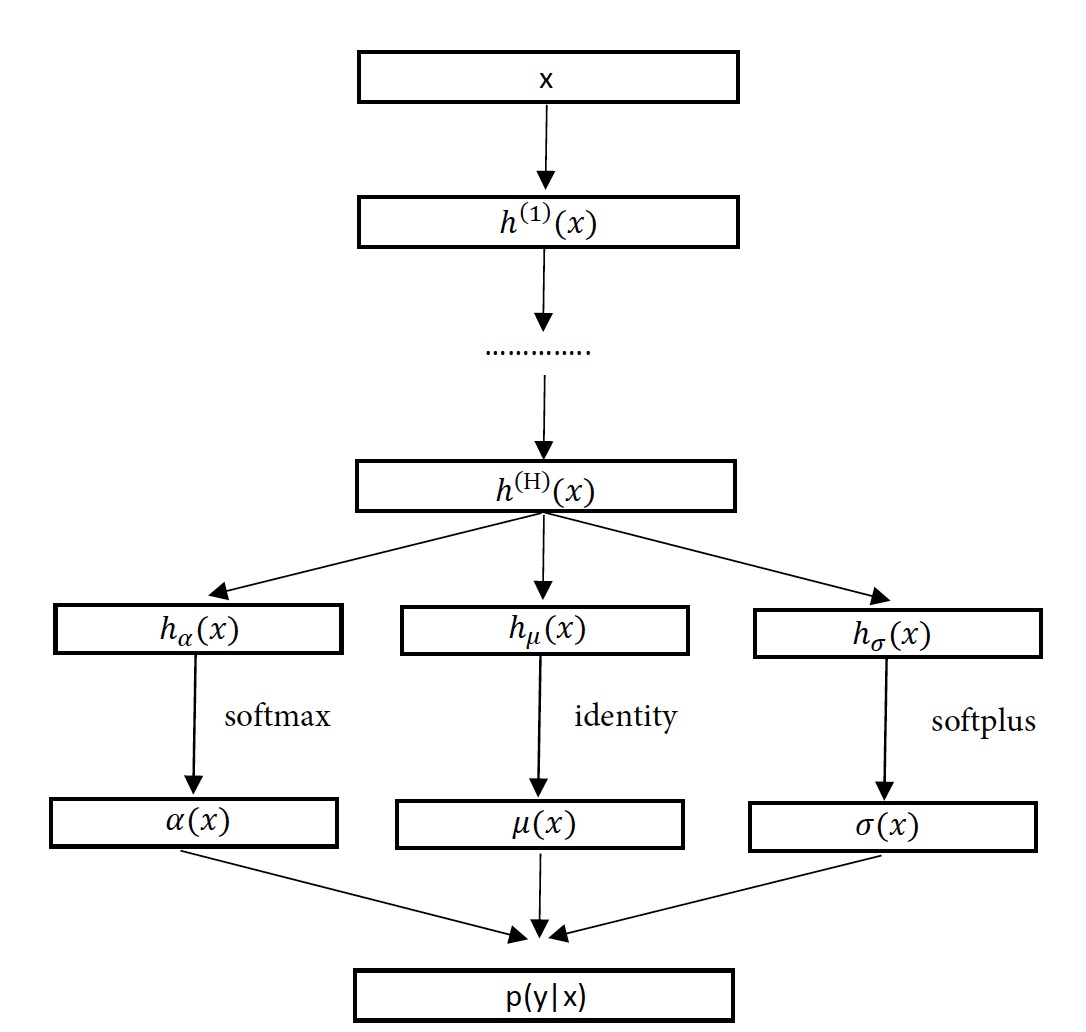}
\caption{Structure of the mixture density network.}
\label{fig1}
\end{figure}

We aim to estimate an unknown conditional probability density function of some univariate response $Y$ given a predictor vector \change{$X \in \mathbb{R}^{d_0}$ with $d_0$ being the input dimension.} Suppose the conditional density of $Y$ given $X$ follows a mixture density network model,  
\begin{eqnarray}\label{eqn:mdn}
f(y|x)=\sum_{g=1}^{G} \alpha_g(x) \frac{1}{\sqrt{2 \pi}\sigma_g(x)} \exp\left[ -\frac{\{ y-\mu_g(x) \}^2}{2 \sigma_g^2(x)} \right],
\end{eqnarray}
where $G$ is the number of mixture components, and deep neural networks are used to estimate the mean vector $\mu(x) = (\mu_1(x), \ldots, \mu_G(x))^\top$, the standard deviation vector $\sigma = (\sigma_1(x),...,\sigma_G(x))^\top$, and the weight vector $\alpha=(\alpha_1(x),...,\alpha_G(x))^\top$. Figure \ref{fig1} depicts the structure of the model. \change{The input layer is the $d_0$-dimension vector $x$.} Then there are $H$ hidden layers, each with a number of hidden units. \change{A hidden layer is between the input and output layers, which takes in a set of weighted inputs and produces an output through an activation function.} The last hidden layer outputs a $G$-dimensional vector $h^{(H)}(x)$, and is connected to three parallel layers whose outputs are given by 
\begin{align*}
h_{\alpha}(x)=\Theta_1^\top h^{(H)}(x), \quad h_{\mu}(x)=\Theta_2 h^{(H)}(x), \quad h_{\sigma}(x)=\Theta_3^\top h^{(H)}(x), 
\end{align*}
respectively, where $\Theta_j$ is a $G \times G$ coefficient matrix \change{that is to be trained via back propagation}, $j=1,2,3$. Next, two of those functions pass through activation functions, yielding
\begin{align*}
\alpha(x)=\text{softmax}(h_{\alpha}(x)), \quad \mu(x)= h_{\mu}(x), \quad \sigma(x)=\text{softplus}(h_{\sigma}(x)),
\end{align*}
respectively, where $\alpha(x)$, $h_{\alpha}(x)$, $\mu(x)$, $h_{\mu}(x)$, $\sigma(x)$ and $h_{\sigma}(x)$ are all $G$-dimensional vectors, and the activation functions are applied in an element-wise fashion. Finally, all these components are combined to parametrize $f(y|x)$ according to \eqref{eqn:mdn} \change{with a total of $W$ parameters.}  

Next, we propose a new MDN architecture to model the conditional density of a multivariate response variable $Y \in \mathbb{R}^{d_y}$. The main idea is to factorize the joint conditional density function $f(y|x)$ as the product of $d_y$ conditional densities, each with a univariate response, 
\begin{eqnarray}\label{eqn:factor}
f(y|x) = f_1(y_1|x) f_2(y_2|x,y_1) ... f_{d_y}(y_{d_y}|x,y_1,y_2,...,y_{d_y-1}).
\end{eqnarray}
It then suffices to model each $f_j(y_j|x,y_1,...,y_{j-1})$ separately. When the individual component of $Y$ is a continuous variable, we use the MDN model \eqref{eqn:mdn} to estimate the conditional density, whereas when it is a categorical variable, we use a supervised learning method, such as a random forest \citep{breiman2001random}, or a deep neural network \citep{lecun2015deep} to estimate the probability mass function. We briefly note that, \citet{mdn} also considered a version of MDN for the multivariate response, by extending \eqref{eqn:mdn} to a mixture of multivariate normal densities. However, such an extension does not work well when the components of the response have mixed type of continuous and categorical variables. 

We also comment that, most of the existing MDN literature study i.i.d.\ data. In our setting, the observed data are time-dependent. We later show that MDN is equally applicable, as long as the time series satisfies some mixing conditions such as $\beta$-mixing \citep{beta_mixing}.

\subsection{Testing Markov property}
\label{sec:test-proc}

Next, we develop a testing procedure for the hypotheses in \eqref{eqn:hypo}, where the key idea is to build upon the doubly robust test statistic \eqref{eqn:S.qmunu} and estimate the CCFs using MDN. Moreover, to avoid requiring the estimators of the CCFs to satisfy some restrictive metric entropy conditions, we employ the sample splitting and cross-fitting strategy. We first summarize our testing procedure in Algorithm \ref{alg1}, then discuss the main steps in detail.   

In Step 1 of the algorithm, we divide the time series into $L$ non-overlapping chunks of similar sizes. For simplicity, suppose the length $T$ of the observed 
time series is a multiple of $L$, and let $n = T/L$. Let $\mathcal{I}^{(\ell)} = \{(\ell-1)n+1,(\ell-1)n+2,..., \ell n\}$ denote the indices of the $\ell$th chunk of the time series, and let $\bar{\mathcal{I}}^{(\ell)}=\cup_{j=1}^{\ell} \mathcal{I}^{(j)}$ denote the union of indices of 
the first $\ell$ chunks, $\ell = 1, \ldots, L$. Data splitting allows us to use part of the data, i.e., the data $\bar{\mathcal{I}}^{(\ell)}$ up to chunk $\ell$, to train the MDN model, and another part, i.e., $\mathcal{I}^{(\ell+1)}$, to construct the test statistic. We then aggregate the estimates over all chunks to improve the estimation efficiency.

\begin{algorithm}[t!]
\caption{Testing procedure for the Markov property.}
\label{alg1}
\begin{algorithmic}
\item[]\normalsize
\begin{enumerate}
\item[\change{\textbf{Input}:}] \change{Data $\{X_t\}_{t=1,...,T}$, the number of data chunks $L$, the number of pairs $B$, the largest number of lags $Q$, and the number of samples from the generators $M$.}
\item[\textbf{Step 1}:] Divide the time series data into $L$ non-overlapping chunks, where $n = T/L$, $\mathcal{I}^{(\ell)}=\{(\ell-1)n+1,(\ell-1)n+2,..., \ell n\}$, and $\bar{\mathcal{I}}^{(\ell)}=\cup_{j=1}^{\ell} \mathcal{I}^{(j)}$, $\ell=1,\ldots,L$. 

\item[\textbf{Step 2}:] Deep conditional forward-backward generative learning.
\vspace{-1em}
\begin{enumerate}[({2}a)]
\item Obtain the estimators of a forward generator $\widehat{f}^{(\ell)}_{X_t|X_{t-1}}$, and a backward generator $\widehat{f}_{X_{t-1}|X_t}^{(\ell)}$, using the data $\bar{\mathcal{I}}^{(\ell)}$ up to chunk $\ell$, $\ell = 1, \ldots, L-1$. 

\item Randomly sample $M$ copies of $d$-dimensional time series observations $\{X_{m,f}^*\}_{m=1}^{M}$ and $\{X_{m,b}^*\}_{m=1}^{M}$ from each generator.
	
\item Randomly sample $B$ pairs $\{(\mu_b,\nu_b)\}_{1\le b\le B}$ from a multivariate normal distributions with zero mean and identity covariance matrix. 

\item Compute the CCF estimators $\widehat{\varphi}^{(\ell)}(\mu_b | x)$ and $\widehat{\psi}^{(\ell)}(\nu_b | x)$ according to \eqref{eqn:varp}, for $\ell=1, \ldots, L-1$, and $b = 1, \ldots,  B$.
\end{enumerate}
			
\item[\textbf{Step 3}:] Construct the test statistic.
\vspace{-1em}
\begin{enumerate}[({3}a)]
\item Compute $\widehat{S}(q,\mu_b,\nu_b)$ according to  \eqref{test_stat0}, for $q=2,...,Q$, $b=1,...,B$. 

\item Construct the test statistic $\widehat{S}$ according to \eqref{eqn:TS}. 
\end{enumerate}

\item[\textbf{Step 4}:] Compute the critical value.
\vspace{-1em}
\begin{enumerate}[({4}a)]
\item Compute the covariance matrix $\widehat{\Sigma}^{(q)}$ according to \eqref{sigmaq}, for $q=2,...,Q$.
			
\item Compute the critical value according to \eqref{eqn:critical-value}. 
\end{enumerate}

\item[\textbf{Step 5}:] Reject $H_0$ if $\widehat{S}$ is greater than $\widehat{c}_{\alpha}$.
\end{enumerate}
\end{algorithmic}
\end{algorithm}

In Step 2, we employ MDN to estimate the CCFs. Specifically, for each subset $\ell = 1, \ldots, L-1$, we first apply MDN to the data $\bar{\mathcal{I}}^{(\ell)}$ up to the $\ell$th chunk to obtain the estimates of two conditional probability density functions, a forward generator $\widehat{f}^{(\ell)}_{X_t|X_{t-1}}$, and a backward generator $\widehat{f}_{X_{t-1}|X_t}^{(\ell)}$. For the forward generator, the ``predictor" for the MDN model \eqref{eqn:factor} is $(X_1, X_2, \ldots, X_{\ell n - 1})^\top$ and the ``response" is $(X_2, X_3, \ldots, X_{\ell n})^\top$, whereas for the backward generator, the ``predictor" for \eqref{eqn:factor} is $(X_2, X_3, \ldots, X_{\ell n})^\top$ and the ``response" is $(X_1, X_2, \ldots,$ $X_{\ell n - 1})^\top$. Given the two estimated density functions $\widehat{f}^{(\ell)}_{X_t|X_{t-1}}$ and $\widehat{f}_{X_{t-1}|X_t}^{(\ell)}$, we then randomly sample $M$ copies of $d$-dimensional time series observations $\{X_{m,f}^*\}_{m=1}^{M}$ and $\{X_{m,b}^*\}_{m=1}^{M}$, respectively.  Next, we consider different combinations of $(\mu,\nu)$ for the test statistic $S(q, \mu, \nu)$ in \eqref{eqn:S.qmunu}. Toward that end, we randomly sample $B$ i.i.d.\ pairs of $\{(\mu_b,\nu_b)\}_{b=1}^{B}$ from a multivariate normal distribution with zero mean and identity covariance matrix. Finally, by noting that $\varphi^*(\mu | x) = \Mean \{\exp(i \mu^\top X_t)|X_{t-1}=x\}$ and $\psi^*(\nu | x) = \Mean \{\exp(i \mu^\top X_{t-1})|X_t=x\}$, we obtain the Monte Carlo estimators of $\varphi^*(\mu | x)$ and $\psi^*(\nu | x)$ for each pair of $(\mu_b,\nu_b)$ as 
\begin{eqnarray}\label{eqn:varp}
\widehat{\varphi}^{(\ell)}(\mu_b | x) = \frac{1}{M} \sum_{m=1}^M \exp(i \mu_b^\top X_{m,f}^*), \quad \widehat{\psi}^{(\ell)}(\nu_b | x) = \frac{1}{M} \sum_{m=1}^M \exp(i \nu_b^\top X_{m,b}^*). 
\end{eqnarray}
Due to the use of both forward and backward generators and deep neural networks, we refer to this step as deep conditional forward-backward generative learning. 

In Step 3, we construct our final composite test statistic given the estimates of $\widehat{\varphi}^{(\ell)}(\mu_b | x)$ and $\widehat{\psi}^{(\ell)}(\nu_b | x)$. We first compute $S(q, \mu, \nu)$ in \eqref{eqn:S.qmunu} using the cross-fitting strategy, i.e.,
\begin{align} \label{test_stat0}
\begin{split}
\widehat{S}(q, \mu_b, \nu_b) = \; & \frac{1}{T-n-(q-1)(L-1)} \sum_{\ell=1}^{L-1} \; \sum_{t=1}^{n-q+1} \Big\{ \exp(i \mu_b^\top X_{\ell  n+t+q-1}) \\
& - \widehat{\varphi}^{(\ell)}(\mu_b | X_{\ell n+t+q-2}) \Big\} \Big\{ \exp(i \nu_b^\top X_{\ell  n+t-1})-\widehat{\psi}^{(\ell)}(\nu_b | X_{\ell n+t}) \Big\},
\end{split}
\end{align}
for a given $q = 2, \ldots, Q$, and $Q$ denotes the largest number of lags to consider in the test. We note that, for any given $\ell = 1, \ldots, L-1$, the set of random variables $\{X_{\ell n+t}\}_{1\le t\le n}$ that appear in  \eqref{test_stat0} are from the $(\ell+1)$th chunk of the data, and are, under $H_0$, independent of $\widehat{\varphi}^{(\ell)}$ and $\widehat{\psi}^{(\ell)}$ given $X_{\ell n+1}$. \change{This allows us to avoid imposing certain entropy growth condition that limits the growth rate of the VC dimension of the MDN model with respect to the sample size \citep{chernozhukov2018double}.} A similar cross-fitting procedure has also been utilized by \citet{luedtke2016statistical} and \citet{shi2020statistical} for evaluation of an optimal policy, as well as by \citet{luedtke2018parametric} and \citet{shi2021statistical} for high-dimensional statistical inference. Next, since $\widehat{S}(q, \mu_b, \nu_b)$ is complex-valued, we use $\widehat{S}_R(q, \mu_b, \nu_b)$ and $\widehat{S}_I(q, \mu_b, \nu_b)$ to denote its real and imaginary part, respectively. We construct our final test statistic as
\begin{eqnarray} \label{eqn:TS}
\widehat{S} = \max_{b\in \{1,\ldots,B\}}\max_{q\in \{2,\ldots,Q\}} \sqrt{T-n-(q-1)(L-1)} \max\left( | \widehat{S}_R(q, \mu_b, \nu_b) |, | \widehat{S}_I(q, \mu_b, \nu_b) | \right).	
\end{eqnarray}
\change{In \eqref{eqn:TS}, we take the maximum absolute value over multiple combinations of $(q, \mu_b, \nu_b)$ to construct the test statistic, while we generate $\mu_b$ and $\nu_b$ from a Gaussian or uniform distribution. This way, we do not have to impose a bounded support for $(\mu_b, \nu_b)$, and avoid grid search that can be computationally intensive in a high-dimensional setting.}

In Step 4, we compute the critical value of the test statistic $\widehat{S}$. A key observation is that, under $H_0$, each $\widehat{S}_R(q,\mu_b,\nu_b)$ and $\widehat{S}_I(q,\mu_b,\nu_b)$ corresponds to a sum of martingale difference sequences. \change{Since the sum of martingale difference is a martingale \citep{hamilton2020time},} it follows from the high-dimensional martingale central limit theorem that $\widehat{S}$ converges in distribution to a maximum of some Gaussian random variables. This allows us to employ the high-dimensional multiplier bootstrap method of  \cite{belloni2018} to estimate the critical value. Specifically, we stack $\widehat{S}_R(q,\mu_b,\nu_b)$ and $\widehat{S}_I(q,\mu_b,\nu_b)$ for a given $q$ and all $b = 1, \ldots, B$ together to form a $2B$-dimensional vector, and estimate the covariance matrix of this vector by
\begin{eqnarray}\label{sigmaq}
\widehat{\Sigma}^{(q)}=\sum_{\ell=1}^{L-1} \sum_{t=1}^{n-q+1} \frac{(\lambda_{R,\mathbb{\ell},q,t}^\top,\lambda_{I,\mathbb{\ell},q,t}^\top)^\top  (\lambda_{R,\mathbb{\ell},q,t}^\top,\lambda_{I,\mathbb{\ell},q,t}^\top)}{(T-n-(q-1)(L-1))}, 
\end{eqnarray}
where $\lambda_{R,\mathbb{\ell},q,t}, \lambda_{I,\mathbb{\ell},q,t}$, $\ell = 1, \ldots, L-1$, $t = 1,...,n-q+1$, are both $B$-dimensional vectors, whose $b$th element is, respectively, the real and imaginary part of 
\begin{align*}
\Big\{ \exp(i\mu^\top X_{t+q-1+\ell n}) - \widehat{\varphi}^{(-\ell)}(\mu|X_{t+q-2+\ell n}) \Big\} \Big\{ \exp(i\nu^\top X_{t-1+ \ell n})-\widehat{\psi}^{(-\ell)}(\nu|X_{t+\ell n}) \Big\}.
\end{align*}  
We then compute the critical value $\widehat{c}_{\alpha}$ by simulating the upper $(\alpha/2)$th critical value of 
\begin{align} \label{eqn:critical-value}
\max_{q \in \{2,\ldots,Q\}} \left\| \{\widehat{\Sigma}^{(q)}\}^{1/2} Z_q \right\|_{\infty}, 
\end{align}
using Monte Carlo, where $Z_0, \ldots, Z_Q$ are i.i.d.\ $2B$-dimensional standard normal vectors. 

In Step 5, we reject $H_0$, if $\widehat{S} > \widehat{c}_{\alpha}$, under a given significance level $\alpha>0$. 

We make a few remarks. \change{First, in terms of the computational cost, step 2(a) is the most intensive step in Algorithm \ref{alg1}, as it involves fitting multiple MDN models.} Second, there are a number of hyper-parameters in our test, including the number of mixture components $G$, the number of data chunks $L$, the number of pairs $B$ of $(\mu,\nu)$, the number of samples $M$ from the forward and backward generators, and the largest number of lags $Q$ considered in the test. \change{We proposed to choose $G$ using cross-validation, and take the rest as the input parameters.} We further discuss their theoretical choices in Section \ref{sec:theory}, and their empirical choices in Section \ref{sec:sim}.

\subsection{Determining Markov order}
\label{modelselect}

The proposed test can be used to determine the order of the Markov model. Specifically, let $X_{t}^{(k)} = (X_t^\top, ..., X_{t+k-1}^\top)^\top$ denote the multivariate time series that concatenates the most recent $k$ observations at each time point. Suppose the data follows a $K$th order Markov model. Then the null hypothesis $H_0$ holds for the concatenated time series $X_t^{(k)}$ for any $k \ge K$, but does not hold for any $k< K$. This suggests we can sequentially test the Markov property on the concatenated time series $X_t^{(k)}$ for $k=1, 2, \ldots$. We set the estimated order to be the first integer $k$ by which we fail to reject $H_0$. We also briefly remark that $K$ is different from $Q$. The former denotes the largest possible order of the underlying Markov model, whereas the latter denotes the largest number of lags considered in our test for a series of conditional dependences.

\section{Theory} 
\label{sec:theory}

\subsection{Convergence rate of MDN} 
\label{sec:mdn-bound}

We first establish the error bound of the mixture density network estimator, then establish the consistency of the proposed test. We begin with some regularity conditions, 
and argue they are relatively mild and reasonable. 

Let $f_{X_{t+1}|X_t}^*(\cdot |x)$ and $f_{X_t|X_{t+1}}^*(\cdot |x)$ denote the true conditional density function of $X_{t+1}$ given $X_t=x$, and that of $X_t$ given $X_{t+1}=x$, respectively. A key observation is that $f_{X_{t+1}|X_t}^*=\argmax_f \mathbb{E}[\log \{f(X_{t+1}|X_t)\}]$, and $f_{X_{t}|X_{t+1}}^*=\argmax_f \mathbb{E}[\log \{f(X_{t}|X_{t+1})\}]$, \change{where $f$ belongs to a Sobolev ball with the smoothness $\gamma \in \mathbb{N}_{+} : \big\{ f: \max_{\nu,\|\nu\|_1 \leq \gamma} \sup_{\substack{x }}|D^{\nu}f(x)| < +\infty \big\}$.} Given the data $\bar{\mathcal{I}}^{(\ell)}$ up to chunk $\ell$, the estimated density functions are  
\begin{eqnarray*}
\widehat{f}^{(\ell)}_{X_{t+1}|X_t} = \argmax_f  \sum_{t=1}^{T-1} \log \{f(X_{t+1}|X_t)\}, \quad
\widehat{f}^{(\ell)}_{X_t|X_{t+1}} = \argmax_f \sum_{t=1}^{T-1} \log \{f(X_t|X_{t+1})\}, 
\end{eqnarray*}
based on the maximum likelihood. In the following, we focus on establishing the statistical properties of $\widehat{f}^{(\ell)}_{X_{t+1}|X_t}$. The properties of $\widehat{f}^{(\ell)}_{X_t|X_{t+1}}$ can be derived in similar manner. 

\begin{assump} \label{assump1} Suppose the following conditions hold for the time series $X_t$.
\vspace{-0.5em}
\begin{enumerate}[(i)]
\item Let $X_t$ be stationary, and its $\beta$-mixing coefficient satisfy the that $\beta(t) \le c_1\exp(-c_2 t)$ for some constants $c_1,c_2>0$. 

\vspace{-0.5em}
\item \change{Let $\mathcal{X}$ denote the support of $X_t$, and $\mathcal{X}$ be a compact subset of $\mathbb{R}^{d}$.}
\end{enumerate}
\end{assump}

\noindent 
Assumption \ref{assump1}(i) requires the $\beta$-mixing coefficient to decay exponentially with respect to $t$. Under the Markov property, it is equivalent to the geometric ergodicity condition \citep{Brad2005}. Such a condition is commonly imposed in the time series literature \citep[see, e.g.,][]{cline1999geometric, liebscher2005towards, beta_mixing}. \change{We also note that the $\beta$-mixing condition is not limited to a Markov process. For instance, \cite{neumann2011absolute} considered a class of observation-driven Poisson count process, which is $\beta$-mixing but non-Markovian.}

\begin{assump} \label{assump2}
Suppose the following conditions hold for the true density function $f_{X_{t+1}|X_t}^*$. 
\vspace{-0.5em}
\begin{enumerate}[(i)]
\item \change{Suppose $f_{X_{t+1}|X_t}^*(y|x)$ can be well-approximated by a conditional Gaussian mixture model with $G$ components, in that, there exists some constant $\omega_1>0$,  such that 
\begin{align*}
\left| f_{X_{t+1}|X_t}^*(y|x)-\sum_{g=1}^G\frac{\alpha^*_g(x)}{\sqrt{2 \pi}\sigma^*_g(x)} \exp\left\{-\frac{(y-\mu^*_g(x))^2}{2 {\sigma^*_g}^2(x)}\right\}\right| =O(G^{-\omega_1}),
\end{align*}
where the big-$O$ term is uniform in $x$ and $y$.}

\vspace{-0.5em}
\item Suppose $\{\mu^*_g\}_{g=1}^{G}$ and $\{\sigma^*_g\}_{g=1}^{G}$ are uniformly bounded away from infinity, and there exist a constant $C_0 > 0,\omega_2 \ge 0$, such that $\sigma^*_g(x) \ge C_0 G^{-\omega_2}$ for any $g$ and $x$. 

\vspace{-0.5em}
\item Suppose $\alpha^*_g(\cdot)$, $\mu^*_g(\cdot)$, and $\sigma^*_g(\cdot)$, $g=1,\ldots,G$, all lie in the Sobolev ball with the smoothness $\gamma \in \mathbb{N}_{+} : \big\{ f: \max_{\nu,\|\nu\|_1 \leq \gamma} \sup_{\substack{x }}|D^{\nu}f(x)| < +\infty \big\}$, where the maximum is taken over all $d$-dimensional non-negative integer-valued vectors $\nu$ the sum of whose elements is no greater than $\gamma$, and $D^{\alpha} f$ is the weak derivative \citep{gine_nickl_2015}. 

\vspace{-0.5em}
\item \change{Suppose $f_{X_{t+1}|X_t}^*(\cdot|\cdot)$ is uniformly bounded away from zero on $\mathcal{X}\times \mathcal{X}$.}
\end{enumerate}
\end{assump}

\noindent
Assumption \ref{assump2}(i) requires the true conditional density function $f_{X_{t+1}|X_t}^*$ can be well approximated by a conditional Gaussian mixture model, with a sufficiently large number of components $G$. This is reasonable, since the Gaussian mixture model can approximate any smooth density function, and the conditional Gaussian mixture model can approximate any smooth conditional density function \citep{dalal1983approximating}. Assumption \ref{assump2}(ii) to (iv) impose certain boundedness and smoothness conditions on the mean, variance, and weight functions used in the approximation of $f_{X_{t+1}|X_t}^*$, as well as on $f_{X_{t+1}|X_t}^*$ itself. All these conditions are reasonably mild and hold under numerous settings. We consider three examples to further illustrate. 
 
\begin{example}\label{exam1}
Suppose the true conditional density function $f_{X_{t+1}|X_t}^*$ follows a finite conditional Gaussian mixture model with bounded and smooth mean, variance, and weight functions. Then Assumption 2 trivially holds. 
\end{example}
 
\begin{example}\label{exam2}
Suppose $f_{X_{t+1}|X_t}^*$ follows an infinite conditional Gaussian mixture model, i.e.
\begin{eqnarray}\label{eqn:infgmix}
f^*_{X_{t+1}|X_t}(y|x) = \int g(y_0|x) \phi_{\sigma}(y-y_0)dy_0,
\end{eqnarray}
where $g$ denotes a certain conditional density function, and $\phi_{\sigma}$ denotes the probability density function of a Gaussian random variable with mean zero and variance $\sigma^2$. Then under some mild conditions on $g$, the next lemma show that Assumption 2 holds. 
 
\begin{lemma}\label{lemma1}
\change{Suppose \eqref{eqn:infgmix} holds, with a conditional density function $g$ bounded away from infinity. Suppose the support of $g(\cdot |x)$ is a subset of $[-C_1,C_1]$ for any $x$. It follows that
\begin{eqnarray*} \label{eqn:mixture}
\left| f^*_{X_{t+1}|X_t}(y|x)-\sum_{g=1}^{G} \alpha^*_g(x) \phi_{\sigma}\left(y+C_1-\frac{2C_1 (g-1)}{G}\right)\right| \leq c_4G^{-1}, 
\end{eqnarray*}
where $\alpha^*_g(x) = \mathop{\mathlarger{\int}}_{-C_1 + \frac{2C_1 (g-1)}{G}}^{-C_1 + \frac{2C_1 g}{G}} g(z|x) dz$, and $c_4$ is a positive constant independent of $x$ and $y$.}
\end{lemma}

\noindent
According to Lemma \ref{lemma1}, the mean $\{\mu^*_g(x)\}_{g=1}^{G}$ and variance $\{\sigma^*_g(x)\}_{g=1}^{G}$ are constant functions of $x$, which are equal to $2C_1(g-1)/K-C_1$ and $\sigma$. Then Assumption \ref{assump2}(i) holds with $\omega_1=1$, and Assumption \ref{assump2}(ii) holds with $\omega_2=0$. When $g$ lies in the Sobolev ball with the smoothness parameter $\gamma$, so are the weight functions $\{\alpha^*_g\}_{g=1}^{G}$, and Assumption \ref{assump2}(iii) holds. \change{Assumption \ref{assump2}(iv) holds as $g$ is bounded away from zero. Besides, the approximation error rate obtained in Lemma \ref{lemma1} is $O(G^{-1})$ in $L_\infty$ norm, which is shaper than the $O(G^{-1/2})$ rate in $L_2$ norm obtained in \citet[Lemma 1]{barron1993universal}, as we focus on the Gaussian mixture and one-dimensional case.}
\end{example}

\begin{example}\label{exam3}
Suppose $f_{X_{t+1}|X_t}^*$ satisfies Assumption \ref{assump2}(iv), and is Lipschitz continuous, i.e., $|f^*(y_1|x)-f^*(y_2|x)| = O(|y_1-y_2|)$ where the big-$O$-term is uniform in $x$. It follows from \citet[Theorem 9]{nguyen2019approximations} that $f^*$ can be well-approximated by an infinite conditional Gaussian mixture model specified in \eqref{eqn:infgmix} with $g=f^*$, with the approximation error $O(\sigma)$. In addition, similar to Lemma \ref{lemma1}, we can show that this infinite conditional Gaussian mixture model can be approximated by the finite conditional Gaussian mixture model, with the approximation error $O(\sigma^{-1} G^{-1})$. By setting $\sigma=G^{-1/2}$, Assumption \ref{assump2}(i) holds with $\omega_1=1/2$. The mean and variance are both constant functions of $x$, and the variance is lower bounded by $G^{-1/2}$. Assumption \ref{assump2}(ii) thus holds with $\omega_2=1/2$. When $f^*$ lies in the Sobolev ball with the smoothness parameter $\gamma$, so are the weight functions $\alpha^*_g$, and Assumption \ref{assump2}(iii) holds.
\end{example}
  
\begin{assump} \label{assump3} 
Suppose the following conditions hold for the MDN model. 
\vspace{-0.5em}
\begin{enumerate}[(i)]
\item Suppose the MDN function class is given by, for some sufficiently large constant $C_2$, 
\begin{eqnarray*}
\mathcal{F} = \Bigg\{ f(y|x)=\sum_{g=1}^{G} \frac{\alpha_g(x)}{\sqrt{2 \pi}\sigma_g(x)} \exp\left\{-\frac{(y-\mu_g(x))^2}{2 \sigma_g^2(x)}\right\}:  \mathop{\inf}_{x,y} f(y|x) \geq C_2^{-1}, \\
\sup_{x,g}|\mu_g(x)|\le C_2, C_2^{-1}G^{-\omega_2}\le \inf_{x,g}\sigma_g(x)\le\sup_{x,g}\sigma_g(x) \le C_2 \Bigg\},
\end{eqnarray*}
where $\alpha_g$, $\mu_g$ and $\sigma_g$ are parametrized via deep neural networks.  

\vspace{-0.5em}
\item The total number of parameters $W$ in the MDN model is proportional to $G^{(d+\gamma)/\gamma} T^{d/(2\gamma+d)}$ $\log (G T)$, where $\gamma$ is the smoothness parameter specified in Assumption \ref{assump2}(iii). 
\end{enumerate}
\end{assump}

\noindent
\change{Assumption \ref{assump3}(i) is mainly to simplify the technical proof, since the estimated functions are bounded when both the model parameters and the data support are bounded. It is easy to enforce Assumption \ref{assump3}(i) in practice, by imposing range constraints on the model parameters.} Assumption \ref{assump3}(ii) specifies the total number of parameters $W$, which represents a trade-off. On one hand, since we model $\{\alpha_g\}_{g=1}^{G}$, $\{\mu_g\}_{g=1}^{G}$ and $\{\sigma_g\}_{g=1}^{G}$ via deep neural networks, their approximation errors decay as $W$ increases. On the other hand, the estimation error of MDN increases with $W$. We require $W$ to be proportional to $G^{(d+\gamma)/\gamma} T^{d/(2\gamma+d)}\log (G T)$ to balance the bias-variance trade-off, and optimize the convergence rate of the MDN estimator. See the proof of Theorem \ref{mdn_bound} in  the Appendix for more details. 

Next, we establish the error bound of the MDN estimator $\widehat{f}^{(\ell)}_{X_{t+1}|X_t}$. The bound of $\widehat{f}^{(\ell)}_{X_t|X_{t+1}}$ is the same, and can be derived similarly.

\begin{theorem} \label{mdn_bound}
Suppose Assumptions \ref{assump1} and \ref{assump2} hold.  Then, there exist a certain MDN function class satisfying Assumption \ref{assump3}, such that the resulting MDN estimator $\widehat{f}_{X_{t+1}|X_t}^{(\ell)}$ satisfies that
\begin{eqnarray}\label{eqn:f2loss}
\begin{split}
\left\| \widehat{f}_{X_{t+1}|X_t}^{(\ell)}-f^*_{X_{t+1}|X_{t}} \right\|_2= \sqrt{\int_{x,y} |\widehat{f}_{X_{t+1}|X_t}^{(\ell)}(y|x)-f^*_{X_{t+1}|X_{t}}(y|x)|^2dxdy }
\\\leq c d \left\{ G^{-\omega_1}+  G ^{\frac{\gamma+d}{2\gamma}+4\omega_2}T^{-\frac{\gamma}{2\gamma+d}} \log^3 (T G) \right\},
\end{split}
\end{eqnarray}
for some constant $c>0$, and any $\ell = 1, \ldots, L$, with probability at least $1-O(T^{-1})$. 
\end{theorem}

We remark that the first term of the error bound in \eqref{eqn:f2loss} is due to the approximation error of the conditional Gaussian mixture model, while the second term is due to the approximation error of the deep neural networks and the estimation error of the MDN estimator. In general, the error bound increases with $d$ and $\omega_2$, and decreases with $\gamma$ and $\omega_1$.  We next revisit Examples \ref{exam1} to \ref{exam3}, and discuss the corresponding rate of convergence. 

\medskip
\noindent \textbf{Example \ref{exam1} revisited.} In this example, the finite conditional Gaussian mixture model holds. As a result, $G$ is finite and $\omega_1$ can be chosen arbitrarily large. The error bound is then of the same order of magnitude as $dT^{-\gamma/\{2\gamma+d\}} \log^3 (T)$. If the mean, variance, and weight functions are infinitely differentiable, i.e., $\gamma=+\infty$, then the MDN estimator achieves a convergence rate of $d T^{-1/2}$ up to some logarithmic term.  

\medskip
\noindent \textbf{Example \ref{exam2} revisited.} In this example, the infinite conditional Gaussian mixture model holds. As a result, $\omega_1=1$ and $\omega_2=0$. By setting $G$ to be proportional to $T^{2\gamma^2/\{(2\gamma+d)(3\gamma+d)\} }$, the error bound is minimized and is proportional to $dT^{-2\gamma^2 / \{(2\gamma+d)(3\gamma+d)\}} \log^3 (T)$. If $\gamma=+\infty$, then the MDN estimator achieves a convergence rate of $dT^{-1/3}$ up to some logarithmic term. 

\medskip
\noindent \textbf{Example \ref{exam3} revisited.} In this example, we have $\omega_1=\omega_2=1/2$. The error bound is minimized when $G$ is proportional to $T^{2\gamma^2/\{(2\gamma+d)(6\gamma+d)\} }$, and the resulting convergence rate is $dT^{-\gamma^2 / \{(2\gamma+d)(6\gamma+d)\}} \log^3 (T)$. If $\gamma=+\infty$, then the MDN estimator achieves a convergence rate of $dT^{-1/12}$ up to some logarithmic term. 

Finally, we remark on the problem of determining the order of a Markov model. In this case, we are interested in estimating the conditional density function of $X_{t+K}$ given $X_t^{(K)}$ and $X_{t-1}$ given $X_t^{(K)}$. Similar to Theorem \ref{mdn_bound}, we can show that the corresponding error bound is of the same order of magnitude as $$d [ G^{-\omega_1} + G ^{(\gamma+dK) / (2\gamma) + 4\omega_2} \ T^{-\gamma / \{2\gamma+dK\}} \log^3 (TG)].$$ We note that this upper bound depends on the order $K$ only through the exponents of $G$ and $T$.

\subsection{Consistency of the proposed test}
\label{sec:test-consistency}

Given the error bound of the MDN estimator, we now establish the consistency, i.e., the size and power properties of our proposed test. \change{We first show the bias of $\widehat{S}(q,\mu, \nu)$ converges at a faster rate than the forward and backward generators.  

\begin{assump} \label{assump4}
Suppose $\widehat{f}^{(\ell)}_{X_{t+1}|X_t}$ and $\widehat{f}^{(\ell)}_{X_t|X_{t+1}}$ converge at a rate of $O(T^{-\kappa_0})$ for some $\kappa_0>0$. More specifically, suppose 
\begin{eqnarray*}
\sqrt{\Mean \int_{x,y} |\widehat{f}_{X_{t+1}|X_t}^{(\ell)}(y|x)-f^*_{X_{t+1}|X_{t}}(y|x)|^2dxdy}=O(T^{-\kappa_0}),\\
\sqrt{\Mean \int_{x,y} |\widehat{f}_{X_{t}|X_{t+1}}^{(\ell)}(y|x)-f^*_{X_{t+1}|X_{t}}(y|x)|^2dxdy}=O(T^{-\kappa_0}),
\end{eqnarray*}
where the expectation is taken with respect to $\widehat{f}^{(\ell)}_{X_{t+1}|X_t}$ and $\widehat{f}^{(\ell)}_{X_t|X_{t+1}}$.
\end{assump}

\begin{theorem}\label{thm35}
Suppose Assumption \ref{assump4} holds. Then under the null hypothesis $H_0$, 
\begin{align*}
\sup_{q,\mu,\nu} \big| \Mean \widehat{S}(q, \mu, \nu) \big| = O\left( f_{\max}T^{-2\kappa_0} \right),
\end{align*}
where $f_{\max}=\sup_x\max_{1\le t\le T}f_{X_t}(x)$, and $f_{X_t}$ denotes the marginal density function of $X_t$. 
\end{theorem}

\noindent
We note that, when the marginal density functions are uniformly bounded, Theorem \ref{thm35} formally verifies the faster convergence rate of the bias of $\widehat{S}(q,\mu,
\nu)$.

Next, we establish the size property of the proposed test.  

\vspace{-0.5em}
\begin{assump} \label{assump5}
Suppose the following conditions hold. 
\vspace{-0.5em}
\begin{enumerate}[(i)]
\item The convergence rates for $\widehat{f}^{(\ell)}_{X_{t+1}|X_t}$ and $\widehat{f}^{(\ell)}_{X_t|X_{t+1}}$ are both $O(T^{-\kappa_0})$ for some $\kappa_0>1/4$.

\vspace{-0.5em}
\item Suppose there exists some $\epsilon>0$, such that the real and imaginary parts of $\big\{ \exp(i\mu^\top X_{t+q})$ $- \varphi^*(\mu|X_{t+q-1}) \big\} \big\{ \exp(i\nu^\top X_{t})-\psi^*(\nu|X_{t+1}) \big\}$ have their variances greater than $\epsilon$, for any $\mu,\nu$ and $q\in \{0,...,Q\}$.

\vspace{-0.5em}
\item Suppose $M = \kappa_1T^{\kappa_2}$ for some $\kappa_1 > 0, \kappa_2 \geq 1/2$, and $Q \le \max(\rho_0 T,T-2)$ for some constant $0< \rho_0<1$.

\vspace{-0.5em}
\item Suppose $B$ grows polynomially fast with respect to $T$.
\end{enumerate}
\end{assump}
}

\noindent 
Assumption \ref{assump5}(i) requires the convergence rates of $\widehat{f}^{(\ell)}_{X_{t+1}|X_t}$ and $\widehat{f}^{(\ell)}_{X_t|X_{t+1}}$ to be $o(T^{-1/4})$, which allows us to derive the size property of the test based upon Theorem \ref{mdn_bound}. This condition is reasonable. For instance, when the time series dimension $d$ is fixed, this corresponds to requiring that $\gamma>d/2$ for Example \ref{exam1}, and $\gamma>2.69d$ for Example \ref{exam2}. Meanwhile, we may also relax this condition, by using the theory of higher order influence functions \citep{robins2017minimax}. \change{Assumption \ref{assump5}(ii) is a technical condition to help simplify the theoretical analysis. Essentially, it is used to guarantee that the diagonal elements of the asymptotic covariance matrix are bounded away from zero. When the fitted MDN is consistent, it follows that the diagonal elements of the estimated covariance matrix are bounded away from zero as well, with probability tending to $1$. This allows us to apply Theorem 1 of \citet{chernozhukov2017detailed} to establish the size property. This condition automatically holds when the conditional density functions $f^*_{X_{t+1}|X_t}$, $f^*_{X_{t}|X_{t+1}}$, $\|\mu_b\|_2$s and $\|\nu_b\|_2$s are uniformly bounded away from zero. Meanwhile, if we truncate the diagonal elements of the estimated covariance matrix from below by some small positive constant, then this condition is not needed, and the subsequent test remains valid to control the type-I error. }  Finally, Assumption \ref{assump5}(iii) and (iv) impose some requirements on the parameters $M, Q$ and $B$. In particular, $B$ is allowed to diverge with $T$. Therefore, the classical weak convergence theorem is not applicable to show the asymptotic equivalence between the distribution of the test statistic and that of the bootstrap samples given the data. To overcome this issue, we employ the high-dimensional martingale central limit theorem recently developed by \citet{belloni2018}.

\begin{theorem}\label{thm4}
Suppose Assumptions \ref{assump1} and \ref{assump5} hold. Then, as $T\to \infty$, $\prob(\widehat{S} > \widehat{c}_{\alpha}) = \alpha+o(1)$ under the null hypothesis.
\end{theorem}

\change{Next, we establish the power property of the proposed test. 

\begin{assump} \label{assump6}
Suppose the following conditions hold. 
\vspace{-0.5em}
\begin{enumerate}[(i)]
\item Suppose $\sup_{q,\mu,\nu}S_0(q,\mu,\nu) \gg T ^{-1/2}$ $\log^{1/2} (T)$, where $S_0(q,\mu,\nu) = \big| \Mean \big\{ \exp(i\mu^\top X_{t+q})-\varphi^*(\mu|X_{t+q-1}) \big\} \big\{ \exp(i\nu^\top X_{t})-\psi^*(\nu|X_{t+1}) \big\} \big|$.

\vspace{-0.5em}
\item Suppose $B=\kappa_3 T^{\kappa_4}$ for some $\kappa_3 >0, \kappa_4 \geq 1/2$. 
\end{enumerate}
\end{assump}

\noindent
Assumption \ref{assump6}(i) measures the degree to which the alternative hypothesis deviates from the null. This is because, for $q=1,\ldots,Q$, the quantity
\begin{eqnarray}\label{eqn:weakcond}
\sup_{f,g} \left|\Mean \left[ f(X_{t+q})-\Mean \{f(X_{t+q})|X_{t+q-1}\} \right] \left[ g(X_t)-\Mean \{g(X_{t})|X_{t+1}\} \right] \right|
\end{eqnarray}
measures the weak conditional dependence between $X_{t+q}$ and $X_t$ given $X_{t+q-1}$ and $X_{t+1}$ \citep{daudin1980partial}. Here, the supremum is taken with respect to the class of all squared integrable functions of $X$, i.e., $L_2(X)$. According to the Weierstrass approximation theorem, the class of trigonometric polynomials are dense in $L_2(X)$. As such, \eqref{eqn:weakcond} is equal to zero if and only if $\sup_{\mu,\nu}S_0(q,\mu,\nu) = 0$. Therefore, $\sup_{q,\mu,\nu}S_0(q,\mu,\nu)$  measures the degree to which the alternative hypothesis deviates from the null, and we require it to be lower bounded. Assumption 6(ii) is mild as $B$ is user-specified.}

\begin{theorem}\label{thm5}
Suppose Assumptions \ref{assump1}, \ref{assump5}(i) to (iii), and \ref{assump6} hold. Then, as $T \to \infty$, $\prob(\widehat{S}>\widehat{c}_{\alpha}) \to 1$ under the alternative hypothesis.
\end{theorem}

\change{We remark that our proposed test is built on weak conditional independence, and thus is not consistent against \emph{all} alternatives. There are cases when \eqref{eqn:weakcond} equals zero but \eqref{CIA} does not hold, since weak conditional independence does not fully characterize conditional independence. In those cases, our test becomes powerless. A possible remedy is to consider an alternative doubly robust test statistics based on 
\begin{eqnarray*}
\Mean \Big[ \left\{ \exp(i\mu^\top X_{t+q})-\varphi^*(\mu|X_{t+q-1}) \right\} \left\{ \exp(i\nu^\top X_{t})-\psi^*(i\nu^\top X_{t+1}) \right\} \\
\exp\{i (X_{t+1}^\top, \cdots ,X_{t+q-1}^\top)\}\omega_q \Big].
\end{eqnarray*}
The above expectation equals zero for any $q\ge 2$, $\mu,\nu\in \mathbb{R}^{d}$, and $\omega_q\in \mathbb{R}^{d(q-1)}$, and the resulting supremum type test is consistent against all alternative hypotheses. However, it is computationally more expensive, since a large number of Monte Carlo samples $\{(\mu_b,\nu_b,\omega_{q,b})\}_b$ are needed to approximate the supremum over the space of $\mathbb{R}\times \mathbb{R}\times \mathbb{R}^{d(q-1)}$ when $q$ is large. In addition, our numerical analysis finds this test less powerful compared to our proposed test. This agrees with the observation in the literature that, even though the test based on weak conditional dependence is not consistent against all alternatives, it may benefit from a simple procedure, and thus a better power property \citep{Li2020nonparametric}. }

We also note that, Theorems \ref{thm4} to \ref{thm5} have suggested some theoretical choices of the parameters $L, B, M, Q$. In practice, we recommend to set $L$ fixed, and set $M$ to be proportional to the sample size. Besides, we choose a large value for $Q$ that is proportional to $T$, and also choose a large $B$. We discuss their empirical choices in the next section.

\section{Simulations} 
\label{sec:sim}

We study the empirical performance of our proposed test through simulations. We consider three different Markov time series models, each with order $K=3$, dimension $d=3$, and varying length $T = \{500, 1000, 1500, 2000\}$. We apply the proposed sequential testing procedure for $k = 1, 2, \ldots, 5$, and report the percentage of times out of 500 data replications when the null hypothesis is rejected. When $k < K$, this percentage reflects the empirical power of the test, and when $k \ge K$, it shows the empirical size.

We consider a linear type VAR model, a nonlinear type threshold model, and a nonlinear type GARCH model, all of which are commonly used in the time series literature \citep[e.g.,][]{auestad1990, cheng1992, tschernig2000}. 

\medskip
\noindent
\textbf{Model 1}: VAR Model \\
\vspace{-0.05in}
\begin{equation*}
A_1=
\begin{pmatrix}
0.5 & -0.2 & -0.2\\
-0.2 & 0.5 & -0.2\\
-0.2 & -0.2 & 0.5
\end{pmatrix}, \;
A_2=
\begin{pmatrix}
-0.5 & 0.2 & 0.2\\
0.2 & -0.5 & 0.2\\
0.2 & 0.2 & -0.5
\end{pmatrix}, \;
A_3=
\begin{pmatrix}
0.4 & -0.1 & -0.1\\
-0.1 & 0.4 & -0.1\\
-0.1 & -0.1 & 0.4
\end{pmatrix},
\end{equation*}
\smallskip
\begin{equation*} 
X_t = A_1 X_{t-1}+A_2 X_{t-2}+A_3 X_{t-3} + \varepsilon_t
\end{equation*}
where $X_t, \varepsilon_t \in \mathbb{R}^3$, and $\varepsilon_{t,1},\varepsilon_{t,2},\varepsilon_{t,3} 
\stackrel{iid}{\sim} \textrm{Normal}(0,0.5)$. 
	
\medskip
\noindent
\textbf{Model 2}: Threshold Model
\begin{equation*}
A_1=
\begin{pmatrix}
0.5 & -0.2 & -0.2\\
-0.2 & 0.5 & -0.2\\
-0.2 & -0.2 & 0.5
\end{pmatrix}, \;
A_2=
\begin{pmatrix}
-0.5 & 0.2 & 0.2\\
0.2 & -0.5 & 0.2\\
0.2 & 0.2 & -0.5
\end{pmatrix}, \;
A_3=
\begin{pmatrix}
0.4 & -0.1 & -0.1\\
-0.1 & 0.4 & -0.1\\
-0.1 & -0.1 & 0.4
\end{pmatrix},
\end{equation*}
\begin{equation*}
B_1=
\begin{pmatrix}
0.3 & -0.1 & -0.1\\
-0.1 & 0.3 & -0.1\\
-0.1 & -0.3 & 0.3
\end{pmatrix}, \;
B_2=
\begin{pmatrix}
-0.3 & 0.1 & 0.1\\
0.1 & -0.3 & 0.1\\
0.1 & 0.1 & -0.3
\end{pmatrix}, \;
B_3=
\begin{pmatrix}
0.25 & -0.05& -0.05\\
-0.05 & 0.25 & -0.05\\
-0.05 & -0.05 & 0.25
\end{pmatrix},
\end{equation*}
\begin{eqnarray*} 
\begin{cases}
X_t=A_1 X_{t-1}+A_2 X_{t-2}+A_3 X_{t-3} +\epsilon_t  & \text{ if }  \sum_{j=1}^3 X_{t-1,j} \le 0 \\ 
X_t=B_1 X_{t-1}+B_2 X_{t-2}+B_3 X_{t-3} +\epsilon_t  & \text{ if }  \sum_{j=1}^3 X_{t-1,j} > 0       
\end{cases}
\end{eqnarray*}
where $X_t, \varepsilon_t \in \mathbb{R}^3$, and $\varepsilon_{t,1},\varepsilon_{t,2},\varepsilon_{t,3} \stackrel{iid}{\sim} \textrm{Normal}(0,0.5)$. 

\medskip
\noindent
\textbf{Model 3}: Multivariate ARCH Model
\begin{equation*} \label{third}
\left\{
\begin{array}{lr}
X_t = A \tilde{X}_t, \;\; \tilde{X}_t = (\tilde{X}_{t,1}, \tilde{X}_{t,2}, \tilde{X}_{t,3})^\top, \;\; \tilde{X}_{t,j}=h_{t,j}^{\frac{1}{2}}  \varepsilon_{t,j}, \; j = 1,2,3 \\
h_{t,1}=0.1+0.6 \tilde{X}_{t-1,1}^2+0.35 \tilde{X}_{t-3,1}^2 \\
h_{t,2}=0.2+0.8 \tilde{X}_{t-1,2}^2+0.05 \tilde{X}_{t-2,2}^2+0.1 \tilde{X}_{t-3,2}^2 \\
h_{t,3}=0.1+0.3 \tilde{X}_{t-1,3}^2+0.65 \tilde{X}_{t-3,3}^2 \\
\end{array}
\right. 
A=
\begin{pmatrix}
1 & 0.2 & 0.2\\
0.2 & 1 & 0.2\\
0.2 & 0.2 & 1
\end{pmatrix},
\end{equation*}
where $X_t, \varepsilon_t \in \mathbb{R}^3$, and $\varepsilon_{t,1},\varepsilon_{t,2},\varepsilon_{t,3} 
\stackrel{iid}{\sim} \textrm{Normal}(0,0.5)$. 

\change{We apply the proposed test. For the hyper-parameters, we propose to select the number of mixture components $G$ using cross-validation, as its choice is important to the empirical performance. When $G$ is small, the fitted MDN model may suffer from a large bias, leading to an inflated type-I errors, whereas when $G$ is large, the model may be overfitted, yielding a more variable test statistic. For the number of pairs $B$, a larger value of $B$ generally improves the power of the test, but also increases the computational cost. We thus fix it at $B=1000$ to achieve a trade-off between the power and the computational cost. For the rest of parameters, including the number of data chunks $L$, the number of pseudo samples $M$, and the largest number of lags $Q$, we conduct a sensitivity analysis in Section \ref{sec:appex_sensi} of the Supplementary Appendix. We find that the proposed test is not overly sensitive to the choice of these parameters, as long as they are in a reasonable range. We thus set $L=3$, $M=100$, and $Q=10$ in our numerical studies. For MDN, we fix the number of layers $H = 1$, and vary the number of nodes $U$ per hidden layer to vary the total number of parameters, and correspondingly the overall complexity of MDN. We carry out another sensitivity analysis for $U$ in Section \ref{sec:appex_sensi}, and again find a similar performance of the test in a range of values of $U$, so we fix $U = 20$ for the first two models, and $U = 40$ for the last model, as the last one is more complex. We estimate the parameters of MDN through maximum likelihood, where the derivative of the likelihood function with respect to each parameter is derived and the back-propagation is employed. In our implementation, we employ the Adam algorithm \citep{kingma2015adam}, and use \texttt{Python} and \texttt{Tensorflow} \citep{dillon2017tensorflow}. We publish our code on GitHub\footnote{\url{https://github.com/yunzhe-zhou/markov_test}}.}

We compare our proposed test with two baseline tests for the Markov property, including the test by \citet{Chen2012}, which used local polynomial regressions (LPF) to estimate the CCFs, and a version of the random forest-based test by \citet{Shi2020}, which was designed for reinforcement learning, and is modified and adapted to our setting. In addition, \citet{Chen2012} suggested two methods to compute the $p$-value for their test. The first method estimates the asymptotic variance of the test and uses a normal approximation. The second method employs bootstrap. In our settings, we find that the bootstrap procedure is extremely slow for a large $T$. As such, we calculate the $p$-value based on the  normal approximation. 

\begin{table}[t!] 
\centering
\caption{Percentage of times out of 500 data replications when the null hypothesis is rejected under the significance level $\alpha=0.05$. The true order of the Markov model is $K=3$ in all examples. Three methods are compared: our proposed test (MDN), \cite{Shi2020}'s method (RF), and \cite{Chen2012}'s method (LPF).}
\label{tab:d3K3}
\begin{tabular}{|c|c|c|c|c|c|c|c|c|c|} \hline
\multicolumn{10}{|l|}{Model 1: VAR Model} \\ \hline
& \multicolumn{3}{c|}{T = 500} & \multicolumn{3}{c|}{T = 1000} & \multicolumn{3}{c|}{T = 1500} \\ \hline
$k$ & MDN & RF & LPF & MDN & RF & LPF & MDN & RF & LPF\\ \hline
1 & 0.952 & 0.980 & 0.010 & 1.000 & 1.000 & 0.280 & 1.000 & 1.000 &0.722 \\ \hline
2 & 0.258 & 0.508 & 0.016 & 0.856 & 0.954 & 0.116 & 0.992 & 1.000 &0.204 \\ \hline
3 & 0.052 & 0.422 & 0.020 & 0.042 & 0.762 & 0.132 & 0.060 & 0.934 &0.200 \\ \hline
4 & 0.042 & 0.060 & 0.020 & 0.044 & 0.048 & 0.112 & 0.058 & 0.048 &0.200 \\ \hline
5 & 0.056 & 0.052 & 0.032 & 0.044 & 0.050 & 0.134 & 0.048 &  0.044 &0.220 \\ \hline\hline
\multicolumn{10}{|l|}{Model 2: Threshold Model} \\ \hline
& \multicolumn{3}{c|}{T = 500} & \multicolumn{3}{c|}{T = 1000} & \multicolumn{3}{c|}{T = 1500} \\ \hline
$k$ & MDN & RF & LPF & MDN & RF & LPF & MDN & RF & LPF\\ \hline
1 & 0.614 & 0.704 & 0.000 & 0.998 & 0.998 & 0.168 & 1.000 &  1.000 & 0.484 \\ \hline
2 & 0.160 & 0.246 & 0.028 & 0.716 & 0.692 & 0.122 & 0.976 & 0.966 & 0.278 \\ \hline
3 & 0.062 & 0.126 & 0.026 & 0.056 & 0.128 & 0.118 & 0.066 & 0.234 & 0.170 \\ \hline
4 & 0.040 & 0.070 & 0.028 & 0.036 & 0.042 & 0.112 & 0.048 & 0.052 & 0.188 \\ \hline
5 & 0.060 & 0.068 & 0.030 & 0.056 &0.038 & 0.096 & 0.034 & 0.038 & 0.146 \\ \hline\hline
\multicolumn{10}{|l|}{Model 3:  Multivariate ARCH Model} \\ \hline
& \multicolumn{3}{c|}{T = 1000} & \multicolumn{3}{c|}{T = 1500} & \multicolumn{3}{c|}{T = 2000} \\ \hline
$k$ & MDN & RF & LPF & MDN & RF & LPF & MDN & RF & LPF\\ \hline
1  & 0.368 & 0.842 & 0.244 & 0.648 & 0.966 & 0.552 & 0.846 & 1.000 & 0.840 \\ \hline
2  & 0.332 & 0.826 & 0.240 & 0.642 & 0.960 & 0.528 & 0.838 & 1.000 & 0.734 \\ \hline
3  & 0.064 & 0.398 & 0.098 & 0.044 & 0.520 & 0.210 & 0.058 &  0.794 & 0.284 \\ \hline
4  & 0.042 & 0.286 & 0.090 & 0.050 & 0.356 & 0.202 & 0.054 & 0.554 & 0.236 \\ \hline
5  & 0.064 & 0.252 & 0.094 & 0.058 & 0.328 & 0.154 & 0.064 & 0.484 & 0.228 \\ \hline
\end{tabular}
\end{table}

Tables \ref{tab:d3K3} reports the empirical rejection rate of each test under the significance level $\alpha=0.05$, aggregated over 500 data replications. It can be seen that the proposed test effectively controls the type-I error when $k \ge 3$, and is very powerful when $k < 3$. To the contrary, both the two baseline tests suffer from inflated type-I errors for large $T$. For instance, when $T\ge 1000$, the type-I error of the test of \cite{Chen2012} exceeds $0.09$ in all cases. \change{This is probably due to that the local polynomial regression tends to suffer with a larger dimension in the multivariate setting \citep{taylor2013challenging}.} The test of \cite{Shi2020} has considerably large type-I errors when applied to the multivariate ARCH model. This is likely due to the fact that their test was not designed for time series data. 

Finally, we report the computation time of the proposed test. We ran all simulations on savio2 htc node of the UC Berkeley Computing Platform, with 12 CPUs and 128 GB RAM, and it took around 2 minutes on average for a single data replication. \change{We also run an example on a regular laptop computer with a single CPU and 8 GB memory RAM, and it took around 20 minutes on average for one data replication.}

\section{Real Data Applications}
\label{sec:real}

We illustrate our method with three datasets: the temperature dataset \cite[Example 1 of][]{yao}, the PM2.5 dataset \cite[Example 4 of][]{yao}, and the diabetes dataset \citep{marling2018ohiot1dm}. 

The first dataset consists of the monthly temperature of seven cities in Eastern China from January 1954 to December 1998. To remove the seasonal trend, we subtract the average across the same month of the year. This ensures that the resulting time series is stationary. The resulting time series has dimension $d = 7$ and length $T = 528$. 

The second dataset consists of the daily average PM2.5 concentration readings, in the logarithmic scale, at 74 monitoring stations in Beijing and nearby areas of China from January 1, 2015 to December 31, 2016. PM2.5 refers to the mix of solid and liquid particles whose diameters are smaller than 2.5 micrometers, and is a key measure of air quality and pollution. We again subtract the average across the same day of the year. The resulting times series has dimension $d = 74$ and length $T = 731$.

The third dataset consists of measurements, recorded every 5 minutes, involving blood glucose level, meal, exercise and insulin treatment from six patients with type-I diabetes over eight weeks. We divide each day into one-hour intervals, and compute the average blood glucose level, the carbohydrate estimate for the meal, the exercise intensity, and the amount of insulin received during the one-hour interval. For each patient, the resulting time series has dimension $d=4$ and length $T=1100$. 

We note that the third data example is different from the other two examples as well as the setting of our problem in several ways. First, for each $d_0$-dimensional time series, there are $N=6$ replications corresponding to 6 patients. Second, for the $d_0=4$ variables, it is of interest to test the Markov property for three of them, but not the insulin amount, because the amount of insulin is determined by the patients themselves. In addition, the insulin amount should be included in the conditioning set, because it directly affects the blood glucose level. Finally, for the carbohydrate estimate of the meal and the exercise intensity, a good portion of the measurements are zero, because no meal or exercise was taken in those time intervals. We modify the test in Algorithm \ref{alg1} to accommodate these differences. Specifically, in Step 1, to tackle multiple replications, instead of splitting a single time series into multiple chunks, we now randomly split $N$ replications into multiple chunks of similar sizes. In Step 2, to test the Markov property of a subset of variables of the multivariate time series, instead of estimating $\widehat{f}^{(\ell)}_{X_t|X_{t-1}}$, we now estimate the forward generator $\widehat{f}^{(\ell)}_{\widetilde{X}_t|X_{t-1}}$, where $\widetilde{X}_t$ only includes those variables to test about. Meanwhile, we still estimate the backward generator $\widehat{f}^{(\ell)}_{X_t|X_{t-1}}$ as before. Also in Step 2, to tackle the issue that some observed time series involve many zeros, we fit a logistic regression to estimate the conditional densities, while we still use MDN for other continuous time series. The rest of steps remain essentially the same as in Algorithm \ref{alg1}.

\begin{table}[t!]
\centering
\caption{The $p$-values of the sequential tests for $k = 1, 2, ..., 12$ for the three datasets: the temperature data, the PM2.5 data, and the diabetes data, by the three methods: our proposed test (MDN), \cite{Shi2020}'s method (RF), and \cite{Chen2012}'s method (LPF).} 
\label{tab:realdata}
\resizebox{16cm}{!} 
{ 
\begin{tabular}{|l|c|c|c|c|c|c|c|c|c|c|c|c|} \hline
Order $k$ & 1  & 2 & 3 & 4 & 5  & 6 & 7 & 8 & 9  & 10 & 11 & 12 \\ \hline
\multicolumn{13}{|l|}{MDN} \\ \hline
Temperature data & 0.110 & 0.187 & 0.371 & 0.591 & 0.454 & 0.282 & 0.186 & 0.049 & 0.206 & 0.117 & 0.780 & 0.027 \\ \hline
PM2.5 data & 0.394 & 0.365 & 0.259 & 0.467 & 0.706 & 0.140 &  0.288 & 0.437 & 0.312 & 0.168 & 0.355 & 0.470 \\ \hline
Diabetes data & 0 & 0.010 & 0.030 & 0.240 & 0.243 & 0.421 & 0.436 & 0.485 & 0.360  & 0.338 & 0.485 & 0.411 \\ \hline\hline
\multicolumn{13}{|l|}{RF} \\ \hline
Temperature data & 0 & 0.097& 0.154&0.063& 0.023& 0.052& 0.052& 0.026& 0.025 &0.037& 0.019& 0.031 \\ \hline
PM2.5 data & 0.052& 0.004& 0.067& 0.047& 0.056& 0.044& 0.029& 0.006& 0.052&0.119& 0.137& 0.119 \\ \hline
Diabetes data & 0 & 0.001& 0.003& 0.097& 0.084& 0.092& 0.066& 0.069& 0.091& 0.103& 0.124& 0.096 \\ \hline\hline
\multicolumn{13}{|l|}{LPF} \\ \hline
Temperature data & 0.805 & 0.847& 0.513 &0.807& 0.250&  0.754& 0.705& 0.144&  0.448&0.214& 0.948& 0.315 \\ \hline
PM2.5 data & 0.201& 0.645& 0.522& 0.336 &0.493 &0.265& 0.245& 0.035& 0.676& 0.091 &0.857& 0.491 \\ \hline
Diabetes data & 0 & 0.225& 0.036& 0.001& 0.915& 0.131& 0.668& 0.866& 0.135&0.068& 0.935& 0.013 \\ \hline
\end{tabular}
}
\end{table}

\change{We apply the proposed test, as well as the two alternative tests of \citet{Chen2012} and \citet{Shi2020}, for $k = 1, 2, ..., 12$ sequentially, to the three datasets. Table \ref{tab:realdata} reports the corresponding $p$-values. For both the temperature and PM2.5 datasets, our test suggests the Markov property holds. This result is consistent with the findings in the literature, as a simple vector autoregressive model of order 1 is sufficient to model these high-dimensional datasets \citep[see, e.g.,][]{yao}. For the diabetes data, the test suggests the order of the Markov model is 4, which is consistent with the finding of \cite{Shi2020}. By contrast, the test of \citet{Chen2012} yields a large $p$-value when $k=2$ then a very small $p$-value when $k=4$ for the diabetes dataset. The test of \citet{Shi2020} tends to select a large value of $k$ for both the temperature dataset and the PM2.5 dataset.}

\baselineskip=21pt
\bibliographystyle{chicago}
\bibliography{ref-markov}

\appendix
In this supplement, we first present the proofs of all the theoretical results in the paper, and some useful auxiliary lemmas. We then present some additional numerical results.

\subsection{Proofs}

\subsubsection{Proof of Lemma \ref{lemma1}}

Denote a $G$-partition of the interval $[-C_1,C_1]$ as $a_1 = -C_1,a_2= -C_1 + 2C_1 / G, \ldots,a_g= - C_1 + 2C_1 (g-1) / G, \ldots, a_{G+1} = C_1$. Since the support of $g$ belongs to the interval $[-C_1,C_1]$, 
\allowdisplaybreaks
\begin{align}\label{eqn0}
	f^*(y|x) 
	= \int_{-C_1}^{C_1} g(y_0|x) \phi_{\sigma}(y-y_0)dy_0  
	= \sum_{g=1}^G \int_{a_g}^{a_{g+1}} g(y_0|x) \phi_{\sigma}(y-y_0) dy_0.
\end{align}
It follows from Taylor's theorem that 
\vspace{-0.05in}
\begin{eqnarray*}
	\phi_{\sigma}(y-y_0)=\phi_{\sigma}(y-a_g)-\phi_{\sigma}^{'}(y-\eta_{g,y_0})(y_0-a_g)\le \phi_{\sigma}(y-a_g)+c(y_0-a_g),
\end{eqnarray*}
for some $\eta_{g_{y_0}}$ that lies between $y_0$ and $a_g$, \change{where the constant $c = 2C_1/ \sqrt{2\pi\sigma^3}$} is uniform in $(g,y_0)$. This, together with \eqref{eqn0}, yields that
\begin{align*}
	f^*(y|x)
	\leq \sum_{g=1}^G \int_{a_g}^{a_{g+1}} g(y_0|x)  dy_0  \; \phi_{\sigma}(y-a_g) + c\sum_{g=1}^G \int_{a_g}^{a_{g+1}} g(y_0|x)(y_0-a_g)dy_0. 
\end{align*}
\change{Given that $g$ is uniformly bounded away from infinity,} the second term on the right-hand-side is of the order of magnitude $O(G^{-1})$. This completes the proof of Lemma \ref{lemma1}. 
\eop

\subsubsection{Proof of Theorem \ref{ccf_equ}}
\label{sec:append-thm1}

\change{The proof is similar to that of Theorem 1 of \citet{Shi2020}, and we outline the key steps here. 
	
	Specifically, if the conditional independence \eqref{CIA} holds, it follows that 
	\begin{eqnarray*} 
		\varphi^*(\mu|X_{t+q-1})\Mean [\exp(i\nu^\top X_{t})|\{X_{j}\}_{t < j \leq t+q-1}] = \Mean \left[ \exp(i\mu^\top X_{t+q}+i\nu^\top X_{t})|\{X_{j}\}_{t < j \leq t+q-1} \right],
	\end{eqnarray*}
	almost surely, for any $t>0$, $q\ge 2$ and the vectors $\mu, \nu \in \mathbb{R}^d$. 
	
	When \eqref{CIA2} holds, we begin with $q=2$. We have that
	\begin{eqnarray*} 
		\varphi^*(\mu|X_{t+1})\Mean \{\exp(i\nu^\top X_{t})|X_{t+1}\} = \Mean \left[ \exp(i\mu^\top X_{t+2}+i\nu^\top X_{t})|X_{t+1} \right],
	\end{eqnarray*}
	almost surely, for any $t>0$ and $\mu, \nu \in \mathbb{R}^d$. By Lemma 3 of \cite{Shi2020}, we obtain that
	\begin{eqnarray*} 
		X_{t+2} \independent X_{t} \mid X_{t+1}, \quad \textrm{ for any } \; t > 0.
	\end{eqnarray*} 
	For $q = 3$,  by \eqref{CIA2}, we have that
	\begin{eqnarray} \label{eq_induct}
		\varphi^*(\mu|X_{t+2})\Mean \{\exp(i\nu^\top X_{t})|\{X_{j}\}_{t < j \leq t+2}\} = \Mean \left[ \exp(i\mu^\top X_{t+q}+i\nu^\top X_{t})|\{X_{j}\}_{t < j \leq t+2} \right],
	\end{eqnarray}
	for any $t>0, \mu, \nu \in \mathbb{R}^{d}$. For any $v \in \mathbb{R}^{d}$, multiplying both sides of \eqref{eq_induct} by $\exp \left(i v^{\top} X_{t+1}\right)$ and taking the  expectation with respect to $X_{t+1}$ conditional on $X_{t+2}$, we obtain that
	\begin{align*}
		&\mathbb{E}\left\{\exp \left(i \mu^{\top} X_{ t+3}\right) \mid X_{t+2}\right\} \mathbb{E}\left\{\exp \left(i v^{\top} X_{t}+i \nu^{\top} X_{t+1}\right) \mid X_{ t+2}\right\} \\
		= \; &\mathbb{E}\left\{ \exp \left(i \mu^{\top} X_{t+3}+i v^{\top} X_{t}+i \nu^{\top} X_{t+1}\right) \mid X_{t+2} \right\}.    
	\end{align*}
	By Lemma 3 of \cite{Shi2020} again, we obtain that 
	\begin{eqnarray*} 
		X_{t+3} \independent X_{t},X_{t+1} \mid X_{t+2}, \quad \textrm{ for any } \; t > 0.
	\end{eqnarray*} 
	Similarly, we can show that, for any $q>4$,
	\begin{eqnarray*} 
		X_{t+q} \independent \{X_{j}\}_{t \leq j < t+q-1} \ | X_{t+q-1} , \quad \textrm{ for any } \; t > 0.
	\end{eqnarray*} 
	This completes the proof of Theorem \ref{ccf_equ}. 
	\eop }

\subsubsection{Proof of Theorem \ref{thm2}}

When $\varphi=\varphi^*$, we have that, 
\begin{eqnarray*}
	\Mean \left[ \exp(i\mu^\top X_{t+q+1})-\varphi^*(\mu|X_{t+q})|\{X_{j}\}_{ j\le t+q} \right] = 0,
\end{eqnarray*}
under the Markov property. Therefore, Equation \eqref{eqn:dr-ee} in Theorem \ref{thm2} holds. In addition, we have $X_{t-1}\independent \{X_{j}\}_{j>t}|X_{t}$ for any $t>1$. When $\psi=\psi^*$, we can similarly show that,
\begin{eqnarray*}
	\Mean \left[ \exp(i\nu^\top X_{t-1})-\psi^*(\nu|X_{t})|\{X_{j}\}_{j>t} \right] = 0.
\end{eqnarray*}
The doubly-robustness property thus follows. This completes the proof of Theorem \ref{thm2}.
\eop

\subsubsection{Proof of Theorem \ref{mdn_bound}}
\label{sec:proofmdn}

We begin with some definitions. Let $\mathcal{F}_{\text{DNN}}$ denote the deep neural network for modeling each unit of MDN. The total number of parameters of $\mathcal{F}_{\text{DNN}}$ is $W$, and the number of hidden layers is $H$.  For a given conditional density estimator $\widehat{f}$, define the following norms,
\begin{align*}
	\|f\|_{\infty} = \sup_{x,y} |f(y|x)|,   \quad\quad
	\|\widehat{f}-f^*\|_n = \sqrt{\frac{1}{T} \sum_{i=1}^T |\widehat{f}(y_i|x_i)-f^*(y_i|x_i)|^2}.
\end{align*}
where $x_i=X_i$ and $y_i=X_{i+1}$. Since $L$ is fixed, the convergence rate of $\widehat{f}_{X_{t+1}|X_t}^{(\ell)}$ is the same as that of $\widehat{f}_{X_{t+1}|X_t}$, which is trained based on the entire dataset. We thus focus on establishing the upper error bound for $\widehat{f}_{X_{t+1}|X_t}$ in the rest of the proof. 

\change{
	We divide the proof into three major steps. In Step 1, we temporarily ignore the dependence over time and apply the machinery of \cite{farrell2019deepneural} to derive the error bounds, 
	\begin{align}
		\begin{split}
			\mathbb{E}_{T}\left( \log f^*-\log f_T \right) \leq \; & \frac{1}{2C^2} \left\{ C_2' G^{4\omega_2} \epsilon+O(G^{-\omega_1}) \right\}^2 \\
			& + \left\{ C_2' G^{4\omega_2} \epsilon+O(G^{-\omega_1}) \right\} \sqrt{\frac{2\widetilde{r}}{C^2  (T-1)}}+\frac{7 A G^{\omega_2} \widetilde{r}}{C (T-1)}, 
		\end{split} \\
		\begin{split}
			(\mathbb{E}-\mathbb{E}_{T}) (\log f^*-\log \widehat{f}) \leq \; & 3C_4r_0 {G}^{\omega_2}  \sqrt{\frac{G \cdot \text{Pdim}(\mathcal{F_{\text{DNN}}})}{T}\log T} \\
			& + 2r_0 {G}^{\omega_2}\sqrt{\frac{\widetilde{r}}{C^2 T}}+\frac{23A G^{\omega_2} \widetilde{r}}{CT}
		\end{split}
	\end{align}
	under the i.i.d.\ setting. In this step, compared to \cite{farrell2019deepneural}, our main contribution lies in analyzing a new deep generative learning model architecture.
	
	We further divide this step into three sub-steps.
	
	In Step 1.1, we show that the MDN model satisfies Condition (2.1) of \cite{farrell2019deepneural}. Then similar to \cite{farrell2019deepneural}, we decompose the error bound into the sum of the approximation error and the estimation error, i.e., 
	\begin{align*}
		\mathbb{E}_T\left( \log f^*_{X_{t+1,1}|X_t}-\log f_{T,X_{t+1,1}|X_t} \right), \;\; \textrm{ and } \;\; (\mathbb{E}-\mathbb{E}_T) \left( \log f^*_{X_{t+1,1}|X_t}-\log \widehat{f}_{X_{t+1,1}|X_t} \right),
	\end{align*}
	where $X_{t+1,1}$ denotes the first element of $X_{t+1}$, and $f_T\in \mathcal{F}$ denotes the best MDN approximation whose explicit form is given in Lemma \ref{lemma4} in Section \ref{sec:aux}. Here we consider the first element of $X_{t+1}$ as an example of one dimension case. We will generalize the result to high-dimension case in Step 3. The expectation $\mathbb{E}$ is taken with respect to the stationary distribution of $(X_t,X_{t+1})$, and $\mathbb{E}_T$ denotes the empirical mean operator, i.e., $\mathbb{E}_T[g]  = (T-1)^{-1} \sum_{t=1}^{T-1} g(X_{t+1}|X_t)$. 
	
	In Step 1.2, We obtain Lemma \ref{lemma4} in Section \ref{sec:aux}, then combine it with the arguments in Section A.1 of \cite{farrell2019deepneural} to upper bound our approximation error.
	
	In Step 1.3, we follow Section A.2 of \cite{farrell2019deepneural} to upper bound the Rademacher complexity of the MDN function class, which in turn upper bounds our estimation error. 
	
	Next, in Step 2, we extend our results to the time-dependent setting based on Lemma \ref{lemma6} in Section \ref{sec:aux}, which itself is based on Berbee's lemma \citep{Berbee1979}. We note that most existing deep learning theories are derived under the i.i.d. setting, whereas our arguments can potentially be used to help establish other deep learning theories under the time-dependent setting. This is another contribution of our theoretical analysis. 
	
	Finally, in Step 3, we combine  the first two steps and establish the finite-sample convergence rate of $\widehat{f}_{X_{t+1}|X_t}$. We also note that our analysis applies to high-dimensional time series. 
	
	To simplify the notation, we omit the subscript $X_{t+1,1}|X_t$  in Steps 1and 2, and write $\widehat{f}_{X_{t+1,1}|X_t}$, $f^*_{X_{t+1,1}|X_t}$, $f_{T,X_{t+1,1}|X_t}$ as $\widehat{f}$, $f^*$ and $f_T$, whenever there is no confusion.

	\paragraph{Step 1.1} We start with the i.i.d.\ setting. We first check that Conditions (2.1) of \cite{farrell2019deepneural} holds. Letting $A = C_2 / \sqrt{2 \pi}$, a uniform upper bound for any $f\in \mathcal{F}$ is given by,
	\begin{eqnarray}\label{eqn:upperbound}
		\|f\|_{\infty} \leq \sum_{g=1}^G \frac{\alpha_g(x)}{\sqrt{2 \pi}\sigma_g(x)} \leq \sum_{g=1}^G \frac{\alpha_g(x)C_2G^{\omega_2}}{\sqrt{2 \pi}} = \frac{C_2G^{\omega_2}}{\sqrt{2 \pi}} = A G^{\omega_2}.
	\end{eqnarray}
	Under Assumption \ref{assump3}(i), the function class $\mathcal{F}$ is uniformly lower bounded. By Taylor's expansion, for any $f,g \in \mathcal{F}$, there exists a constant $C = \inf _{x, y} f(y \mid x) \wedge \inf _{x, y} g(y \mid x)>0$, such that
	\begin{eqnarray*} 
		|\log f(y|x) - \log g(y|x)| & \leq & \frac{1}{C}|f(y|x)-g(y|x)|. 
	\end{eqnarray*}
	Similar to \eqref{eqn:upperbound}, we can obtain a lower bound for $|\log f(y|x) - \log g(y|x)| $. It follows that,
	\begin{eqnarray*}
		\frac{1}{2A^2 G^{2\omega_2}} \mathbb{E} \{ (f-f^*)^2 \} & \leq & \mathbb{E}(\log f^*) - \mathbb{E}(\log f) \leq \frac{1}{2C^2} \mathbb{E}\{ (f-f^*)^2 \}.
	\end{eqnarray*}
	where we use the Taylor expansion for the derivation of the inequalities.  Specifically,  we can show that there $\exists \widetilde{f} \in \mathcal{F}$ such that $\mathbb{E}(\log f^*) - \mathbb{E}(\log f) =\mathbb{E} [\frac{1}{2\widetilde{f}^2}(f-f^*)^2]$ since the expectation for the gradient of $\log f^*$ is 0.  Then the inequalities can be derived according to the lower and upper bound of $\widetilde{f}$.  Recall that $\widehat{f}= \underset{f}{\arg \max } \; \mathbb{E}_T \log f$. We apply the decomposition in Section A.1 of \cite{farrell2019deepneural} and obtain that
	\begin{align} \label{error_bound}
		\begin{split}
			\frac{1}{2A^2 G^{2\omega_2}} \mathbb{E} \left\{ (\widehat{f}-f^*)^2 \right\} & \leq \mathbb{E}(\log f^*) - \mathbb{E}( \log \widehat{f} ) \\
			& \leq \mathbb{E}(\log f^*) - \mathbb{E}(\log \widehat{f}) -\mathbb{E}_T(\log f_T) + \mathbb{E}_T(\log \widehat{f}) \\ 
			&= (\mathbb{E}-\mathbb{E}_T)( \log f^*-\log \widehat{f} ) + \mathbb{E}_T(\log f^*-\log f_T).
		\end{split}
	\end{align}
	Recall that $f_T$ is the best MDN approximation.

	\paragraph{Step 1.2} We upper bound the approximation error. Applying Lemma \ref{lemma4} in Section \ref{sec:aux} yields an upper error bound for $\epsilon=\|f^*-f_T\|_{\infty}$. This together with the Bernstein's inequality leads to an upper bound for $\mathbb{E}_T(\log f^*-\log f_T)$. Combining these two bounds, for any constant $\widetilde{r} > 0$, we obtain that, with probability at least $1-e^{-\widetilde{r}}$, 
	\begin{align} \label{PCK inequality}
		\begin{split}
			\mathbb{E}_{T}\left( \log f^*-\log f_T \right) \leq \; &  \frac{1}{2C^2} \left\{ C_2' G^{4\omega_2} \epsilon+O(G^{-\omega_1}) \right\}^2 \\
			& + \left\{ C_2' G^{4\omega_2} \epsilon+O(G^{-\omega_1}) \right\} \sqrt{\frac{2\widetilde{r}}{C^2  (T-1)}}+\frac{7 A G^{\omega_2} \widetilde{r}}{C (T-1)}. 
		\end{split}
	\end{align}
	where the constant term  $\frac{1}{2C^2}$ comes from the use of the inequality $\mathbb{E}(\log f^*) - \mathbb{E}(\log f) \leq \frac{1}{2C^2} \mathbb{E}\{ (f-f^*)^2 \}$ derived from Step 1.1.  Note that we assume $\mathcal{F}$ is a bounded function class, and that $f^*$ is upper bounded.  As a result, the Bernstein inequality's conditions are satisfied.

	\paragraph{Step 1.3} We upper bound the estimation error. Suppose there exists a constant $r_0>0$ such that $\|f-f^*\|_{2} \leq r_0 G^{\omega_2}$. This holds true if we set $r_0=3A$, where $A$ is defined in Step 1.1. It then follows that, 
	\begin{align*}
		\mathbb{E}\left\{ |\log (f)-\log (f^*)|^2 \right\} \leq \frac{\mathbb{E}(f-f^*)^2}{C^2} \leq \frac{r^2_0 G^{2\omega_2}}{C^2}.
	\end{align*}
	Applying Theorem 2.1 of \cite{bartlett2005local} to the function class $\mathcal{G} = \big\{g = \log{f}-\log{f^*}: f \in \mathcal{F}, \|f-f^*\|_{2} \leq r_0G^{\omega_2} \big\}$, we obtain that, with probability at least $1-2e^{-\widetilde{r}}$, 
	\begin{align} \label{emp_sum_bound}
		(\mathbb{E}-\mathbb{E}_T)( \log f^*-\log \widehat{f} )  \leq    3 \mathbb{E}_{\eta} R_{n} \mathcal{G} + \sqrt{\frac{2 r^2_0  G^{2\omega_2} \widetilde{r} }{C^2 (T-1)}}+\frac{23A  G^{\omega_2} \widetilde{r}}{C(T-1)}
	\end{align}
	where the empirical Rademacher complexity $\mathbb{E}_{\eta} R_{n} \mathcal{G}$ is defined by,  
	\begin{align*}
		\mathbb{E}_{\eta} \mathop{\sup}_{g \in \mathcal{G}} \left\{ \frac{1}{(T-1)} \sum_{t=1}^{(T-1)} \eta_{t}  g(x_{t+1}|x_t)  \right\},
	\end{align*}
	in which $\{\eta_{\tau}\}_{\tau}$ are i.i.d.\ samples from the Rademacher distribution, and $\mathbb{E}_{\eta}$ is the expectation over this distribution.
	
	Next, by contraction and Dudley's chaining properties \citep{mendelson2003few}, we obtain that, with probability at least $1-e^{-\widetilde{r}}$,
	\begin{align*}
		\mathbb{E}_{\eta} R_{n} \mathcal{G} & = \mathbb{E}_{\eta}  \left[ R_{n} \left\{ g : g=\log(f)-\log(f^*),f \in \mathcal{F}, \|f-f^*\|_{L_2} \leq r_0 G^{\omega_2} \right\} \right] \\
		&\leq \frac{2}{ C} \mathbb{E}_{\eta} \left[ R_{n} \left\{ f-f^*,f \in \mathcal{F}, \|f-f^*\|_{L_2} \leq r_0 G^{\omega_2} \right\} \right]  \\
		&\leq \frac{2}{ C} \mathbb{E}_{\eta} R_{n} \{f-f^*,f \in \mathcal{F}, \|f-f^*\|_n \leq 2r_0 G^{\omega_2}\}   \\
		&\leq \frac{2}{ C} \mathop{\inf}_{0<\alpha<2r_0 G^{\omega_2}} \Big \{4\alpha + \frac{12}{\sqrt{(T-1)}}\int_{\alpha}^{2r_0 G^{\omega_2}}\sqrt{\log \mathcal{T}(\delta,\mathcal{F},\|\cdot\|_n)} d\delta \Big \} \\ \label{metric_entropy}
		&\leq \frac{2}{ C} \mathop{\inf}_{0<\alpha<2r_0  G^{\omega_2}} \Big \{4\alpha + \frac{12}{\sqrt{(T-1)}}\int_{\alpha}^{2r_0  G^{\omega_2}}\sqrt{G \log  \mathcal{T}(\frac{ \delta}{C_3 G^{3/2 + 3\omega_2}} ,\mathcal{F_{\text{DNN}}},\|\cdot\|_n)} d\delta \Big \}  \\ 
		&\leq \frac{2}{C} \mathop{\inf}_{0<\alpha<2r_0 G^{\omega_2}} \Big \{4\alpha + \frac{12}{\sqrt{(T-1)}}\int_{\alpha}^{2r_0 G^{\omega_2}}\sqrt{G \log  \mathcal{T}(\frac{ \delta}{C_3 G^{3/2 + 3\omega_2}} ,\mathcal{F_{\text{DNN}}},\|\cdot\|_{\infty})} d\delta \Big \},
	\end{align*}
	where $\mathcal{F}$ is the MDN function class, and $\mathcal{F}_{\text{DNN}}$ is the DNN class defined at the beginning. The second last inequality is due to Lemma \ref{lemma5} in Section \ref{sec:aux}, and $\alpha$ is some positive constant that will be specified later.  
	
	Next, applying Theorems 12.2 and 14.1 in \cite{anthony1999neural}, we further upper bound the entropy integral using the pseudo VC dimension of the neural network function class, denoted by $\text{Pdim}(\mathcal{F}_{\text{DNN}})$. Specifically, we have,
	\begin{align*}
		\mathbb{E}_{\eta} R_{n} \mathcal{G} & \leq  \frac{2}{C} \mathop{\inf}_{0<\alpha<2r_0 G^{\omega_2}} \left\{ 4\alpha + \frac{12}{\sqrt{(T-1)}}\int_{\alpha}^{2r_0 G^{\omega_2}}\sqrt{G \cdot \text{Pdim}(\mathcal{F_{\text{DNN}}})\log \frac{eAT \cdot C_3   G^{3/2 + 4\omega_2}}{\delta \cdot \text{Pdim}(\mathcal{F_{\text{DNN}}})}} d\delta \right\}   \\ 
		& \leq \frac{64r_0 G^{\omega_2}}{ C} \left\{  \sqrt{\frac{G \cdot \text{Pdim}(\mathcal{F}_{\text{DNN}})}{T-1}(\log \frac{eAC_3  G^{3/2 + 4\omega_2}}{r_0}+\frac{3}{2}\log T)} \right\},
	\end{align*}
	by setting $\alpha=2r_0G^{\omega_2}\sqrt{G \cdot \text{Pdim}(\mathcal{F}_{\text{DNN}})/(T-1)}$. 
	
	Therefore, whenever $r_0 \geq 1/(T-1)$, and $T = O(G)$, there exists a constant $C_4>0$, such that, with probability at least $1-e^{-\widetilde{r}}$, 
	\begin{equation} \label{estimation_error}
		\mathbb{E}_{\eta} R_{n} \mathcal{G} \leq  C_4 r_0 G^{\omega_2}  \sqrt{\frac{G  \cdot \text{Pdim}(\mathcal{F}_{\text{DNN}})}{T}\log T}
	\end{equation}
	
	Plugging \eqref{estimation_error} into \eqref{emp_sum_bound}, we obtain that,, with probability at least $1-6e^{-\widetilde{r}}$, 
	\begin{align} \label{final_estimation_bound0}
		\begin{split}
			&(\mathbb{E}-\mathbb{E}_{T}) (\log f^*-\log \widehat{f}) \\ 
			& \leq 3C_4r_0 {G}^{\omega_2}  \sqrt{\frac{G \cdot \text{Pdim}(\mathcal{F_{\text{DNN}}})}{T}\log T} +2r_0 {G}^{\omega_2}\sqrt{\frac{\widetilde{r}}{C^2 T}}+\frac{23A G^{\omega_2} \widetilde{r}}{CT}
		\end{split}
	\end{align}

	\paragraph{Step 2.}
	The data observations in our setting are time-dependent. Nevertheless, thanks to the exponential $\beta$-mixing condition in Assumption \ref{assump1}, and the decoupling lemma, i.e., Lemma \ref{lemma6} in Section \ref{sec:aux}, the estimation error is of the same order of magnitude as that under the i.i.d. setting, up to some logarithmic factors. More specifically, by Lemma \ref{lemma6}, we can construct two i.i.d.\  sequences, $\{U_{2i}^0\}_{i\ge 0}$ and $\{U_{2i+1}^0\}_{i\ge 0}$, such that $U_i^0=(X^0_{ip+1},X^0_{ip+2},...,X^0_{ip+p})$ equals $U_i = (X_{ip+1},X_{ip+2},...$ $,X_{ip+p})$ with probability at least $1-\beta(p)$. Without loss of generality, suppose $T-1$ is divisible by $2p$. Then with probability at least $1- T \beta(p) / p$, we have that, 
	\begin{align} \label{stationary_correct2}
		\mathbb{E}_T( \log f^*-\log f_T ) = \mathbb{E}^0_{T}(\log f^*-\log f_T),
	\end{align}
	where $\mathbb{E}_T^0 \log f = (T-1)^{-1} \sum_{t=1}^{T-1} \log f(X^0_{t+1}|X^0_t)$. 
	
	We next bound the approximation error \eqref{stationary_correct2}. By construction, the sequence $\{U^0_{2t}\}_{t \geq 0}$ is i.i.d., and so is  $\{U^0_{2t+1}\}_{t \geq 0}$. We thus divide the original empirical sum into multiple pieces, and apply the empirical process theory to derive the upper error bound for each of these pieces separately. 
	
	More specifically, we have that,
	\begin{align}\label{eqn:00}
		\mathbb{E}^0_{T}\left( \log f^*-\log f_T \right) = \frac{1}{2p} \left\{ \frac{1}{(T-1)/(2p)} \sum_{\tau=1}^{(T-1)/(2p)} \zeta_{p,2\tau-1}   +  \frac{1}{(T-1)/(2p)} \sum_{\tau=1}^{(T-1)/(2p)} \zeta_{p,2\tau} \right\}
	\end{align}
	where $\zeta_{p,\tau} = \sum_{t=(\tau-1)p+1}^{\tau p} \left\{ \log f^* (X^0_{t+1,1}|X^0_t)-\log f_T (X^0_{t+1,1}|X^0_t) \right\}$. 
	
	Following the same procedure of Step 1.2, for any constant $\widetilde{r} > 0$, we have that, with probability at least $1-e^{-\widetilde{r}}$,
	\begin{align} \label{PCK inequality1}
		\begin{split}
			\mathbb{E}^0_{T}\left( \log f^*-\log f_T \right) \leq \; &  \frac{1}{2C^2} \left\{ C_2' G^{4\omega_2} \epsilon+O(G^{-\omega_1}) \right\}^2 \\
			& + \left\{ C_2' G^{4\omega_2} \epsilon+O(G^{-\omega_1}) \right\} \sqrt{\frac{4p\widetilde{r}}{C^2  (T-1)}}+\frac{14 p A G^{\omega_2} \widetilde{r}}{C (T-1)}. 
		\end{split}
	\end{align}
	
	Similarly, we have that, with probability at least $1- T \beta(p) / p$, 
	\begin{align} \label{stationary_correct1}
		(\mathbb{E}-\mathbb{E}_T)(\log f^*-\log \widehat{f}) = (\mathbb{E}-\mathbb{E}_T^0) (\log f^*-\log \widehat{f}).
	\end{align}
	This allows us to extend the result of Step 1.3 to obtain that, with probability at least $1-6e^{-\widetilde{r}}$, 
	\begin{align} \label{final_estimation_bound}
		\begin{split}
			&(\mathbb{E}-\mathbb{E}^0_{T}) (\log f^*-\log \widehat{f}) \\ 
			& \leq 3C_4r_0 {G}^{\omega_2}  \sqrt{\frac{p G \cdot \text{Pdim}(\mathcal{F_{\text{DNN}}})}{T}\log T} +2r_0 {G}^{\omega_2}\sqrt{\frac{p \widetilde{r}}{C^2 T}}+\frac{23pA G^{\omega_2} \widetilde{r}}{CT}
		\end{split}
	\end{align}
	
	By Lemma 6 of \cite{farrell2019deepneural}, the pseudo VC dimension $\text{Pdim}(\mathcal{F_{\text{DNN}}})$ is of the order $O(WH\log W)$, where $H$ is the number of layers in DNN. Using (\ref{stationary_correct1}), (\ref{stationary_correct2}), plugging (\ref{PCK inequality1}) and (\ref{final_estimation_bound}) into (\ref{error_bound}), and setting $p=2/c_2 \log T$, we obtain that, with probability at least $1-7e^{-\widetilde{r}}-T^{-1}c_1$, 
	\begin{align*}
		\begin{split}
			\|\widehat{f}-f^*\|_2 \leq \; & 
			C_5 \log T \Bigg [ r_0 {G}^{\omega_2}  \sqrt{\frac{G  W H \log W}{T}} + r_0 {G}^{\omega_2}  \sqrt{\frac{   \widetilde{r}}{T}}+\frac{G^{\omega_2}\widetilde{r}}{T} + \epsilon^2 G^{8\omega_2} \\
			& + \epsilon G^{4\omega_2} \left\{ \sqrt{\frac{\widetilde{r}}{T}}+O(G^{-\omega_1}) \right\} \Bigg ] + O(G^{-2\omega_1}),
		\end{split}
	\end{align*}
	for some constant $C_5>0$, and $c_1$, $c_2$ are as defined in Assumption \ref{assump1}.
	
	We next employ the techniques in Appendices A.2.3 and A.2.4 of \cite{farrell2019deepneural} to recursively improve the upper bound $r_0$, which leads to that, for any constant $\widetilde{r}>0$, with probability at least $1-7e^{-\widetilde{r}}-O (T^{-1})$,
	\begin{align*}
		\|\widehat{f}-f^*\|_2 &\leq 
		C_6 \left\{ \left( \sqrt{\frac{G W H \log W}{T}\log^2 T} + \log T \sqrt{\frac{\log \log T +\widetilde{r}}{T}}+  \epsilon  \log T \right) G^{4\omega_2} +O(G^{-\omega_1}) \right\},
	\end{align*}
	for some constant $C_6>0$. 
	
	Recall that the approximation error $\epsilon$ of DNN depends on the model structure. By Lemma \ref{lemma4}, there exists a DNN function class with the approximation error $\epsilon$, such that
	\begin{align*}
		H \leq C_1 \{ \log(G/\epsilon) +1 \}, \quad 
		W \leq C_1 G^{\frac{d}{\gamma}} \ \epsilon ^{-\frac{d}{\gamma}}\{ \log (G / \epsilon)+1 \},
	\end{align*}
	for some constant $C_1>0$. 
	Setting $\epsilon = T^{-\gamma/(2\gamma+d)}$, with probability at least $1-7e^{-\widetilde{r}}-O(T^{-1})$,
	\begin{align*}
		& \|\widehat{f}-f^*\|_2 \\
		\leq \; & C_7 \left( \left[ \sqrt{\frac{G^{\frac{\gamma+d}{\gamma}} \epsilon^{-\frac{d}{\gamma}} \{\log (G/\epsilon)+1\}^3}{T}\log^2 T} + \log T \sqrt{\frac{\log \log T+\widetilde{r}}{T}} + \epsilon \log T \right]  G^{4\omega_2} + O(G^{-\omega_1}) \right) \\
		\leq \; & C_{8} \left[ \left\{ G ^{\frac{\gamma+d}{2\gamma}}T^{-\frac{\gamma}{2\gamma+d}} \log^3 (TG) +\log T  \sqrt{\frac{\log \log T +\widetilde{r}}{T}} \right\} G^{4\omega_2} +O(G^{-\omega_1}) \right].
	\end{align*}
	Setting $\widetilde{r}=\log T$, we obtain that, with probability at least $1- O(T^{-1})$,
	\begin{align*}
		\|\widehat{f}-f^*\|_2 \leq O (G ^{\frac{\gamma+d}{2\gamma} + 4\omega_2}T^{-\frac{\gamma}{2\gamma+d}} \log^3 (TG)) +O(G^{-\omega_1}).
	\end{align*}

	\paragraph{Step 3.}
	According to the factorization rule, the true joint conditional density can be decomposed as: $f_{X_{t+1}|X_t}^*=\prod_{i=1}^d f_{X_{t+1,i}|X_t,X_{t+1,1},\cdots,X_{t+1,i-1}}^*$. By Steps 1 and 2, we have obtained the upper bound for $\|\widehat{f}_{X_{t+1,1}|X_t}-f_{X_{t+1,1}|X_t}^*\|_2$. Following similar arguments, we can show that the same bound holds for $\|\widehat{f}_{X_{t+1,i}|X_t,X_{t+1,1},\cdots,X_{t+1,i-1}}-f_{X_{t+1,i}|X_t,X_{t+1,1},\cdots,X_{t+1,i-1}}^*\|_2$ for each $i=1,\ldots,d$. Then applying triangle inequality iteratively, we have, with probability at least $1- O(T^{-1})$,
	\begin{align*}
		& \quad \; \|\widehat{f}_{X_{t+1}|X_t}-f^*_{X_{t+1}|X_t}\|_2 \\
		&= \left\| \prod_{i=1}^d \widehat{f}_{X_{t+1,i}|X_t,X_{t+1,1},\cdots,X_{t+1,i-1}}-\prod_{i=1}^d f_{X_{t+1,i}|X_t,X_{t+1,1},\cdots,X_{t+1,i-1}}^* \right\|_2 \\
		&\leq \left\| \prod_{i=2}^d \widehat{f}_{X_{t+1,i}|X_t,X_{t+1,1},\cdots,X_{t+1,i-1}}-\prod_{i=2}^d f_{X_{t+1,i}|X_t,X_{t+1,1},\cdots,X_{t+1,i-1}}^* \right\|_2  \\
		&+ O(1)\left\| \widehat{f}_{X_{t+1,1}|X_t,X_{t+1,1},\cdots,X_{t+1,0}}- f_{X_{t+1,1}|X_t,X_{t+1,1},\cdots,X_{t+1,0}}^* \right\|_2  \\
		& \leq \cdots \\
		& \leq \sum_{i=1}^d O(1)\left\| \widehat{f}_{X_{t+1,i}|X_t,X_{t+1,1},\cdots,X_{t+1,i-1}}-f_{X_{t+1,i}|X_t,X_{t+1,1},\cdots,X_{t+1,i-1}}^* \right\|_2\\
		& \leq  \left[ O\left( G ^{\frac{\gamma+d}{2\gamma} + 4\omega_2}T^{-\frac{\gamma}{2\gamma+d}} \log^3 (TG) \right) +O(G^{-\omega_1}) \right] d,
	\end{align*}
	where $O(1)$ denotes some positive constant. 
	This completes the proof of Theorem \ref{mdn_bound}. 
	\eop
}

\subsubsection{Proof of Theorem \ref{thm35}}

The proof is similar to Step 1 of the Proof of Theorem \ref{thm4} below, and is omitted. 
\eop

\subsubsection{Proof of Theorem \ref{thm4}}
\label{sec:append-proof-thm5}

\change{The proof follows that of Theorem 3 of \cite{Shi2020}, and is presented here for completeness. Essentially, we adopt similar arguments and apply to our setting of multivariate time series where we employ MDN instead of random forest to estimate $\widehat{\varphi}$ and $\widehat{\psi}$. Moreover, we remark that Theorem \ref{thm4} is built upon Theorem \ref{mdn_bound}, but cannot be directly deduced from Theorem \ref{mdn_bound}. More specifically, Theorem \ref{mdn_bound} is about the rate of convergence of the MDN estimator, whereas Theorem \ref{thm4} establishes the size property of the proposed test. While Theorem \ref{thm4} requires the convergence rate result of Theorem \ref{mdn_bound}, its proof also requires a number of additional techniques, including the high-dimensional Gaussian approximation theory \citep{chernozhukov2013gaussian}, and the Neyman orthogonality \citep{chernozhukov2018double}. As a result, the proof of Theorem \ref{thm4} is considerably different from that of Theorem \ref{mdn_bound}.}

\change{Define
	\begin{eqnarray*}
		\Gamma^*(q,\mu,\nu)=\frac{1}{T-n-(q-1)(L-1)} \sum_{\ell=1}^{L-1} \; \sum_{t=1}^{n-q+1}\{\exp(i\mu^\top X_{\ell \cdot n+t+q+1}) \\
		- \varphi^{*}(\mu|X_{\ell \cdot n+t+q}) \}  \{\exp(i\nu^\top X_{\ell \cdot n+t-1})-\psi^*(\nu|X_{\ell \cdot n+t})\}.
	\end{eqnarray*}
	for any $q,\mu,\nu$. And $\Gamma^*_R$ and $\Gamma^*_I$ are the real and imaginary part of $\Gamma^*$ respectively.}  \change{Also denote}
\begin{eqnarray*}
	\change{\hat{\Gamma}(q,\mu,\nu)=\frac{1}{T-n-(q-1)(L-1)} \sum_{\ell=1}^{L-1} \; \sum_{t=1}^{n-q+1} \big\{ \exp(i\mu^\top X_{\ell \cdot n+t+q+1})}  \\
	\change{- \hat{\varphi}(\mu|X_{\ell \cdot n+t+q}) \big\}  \big\{ \exp(i\nu^\top X_{\ell \cdot n+t-1})-\hat{\psi}(\nu|X_{\ell \cdot n+t}) \big\}.}
\end{eqnarray*}

We divide the proof into three steps. 

In step 1, we show that 
\begin{eqnarray*} 
	\max_{b\in \{1,\cdots,B\}}\max_{q\in \{2,\cdots,Q\}} \sqrt{T-n-(q-1)(L-1)} |\widehat{\Gamma}(q,\mu_b,\nu_b)-\Gamma^*(q,\mu_b,\nu_b)|=o_p(\log^{-1/2} (T)),
\end{eqnarray*}
which can be obtained by Lemma \ref{lemma7}, by requiring $\widehat{\varphi}$ and $\widehat{\psi}$ to satisfy certain uniform convergence rates. Letting $S^*=\max_{b\in \{1,\cdots,B\}} \max_{q\in \{2,\cdots,Q\}} \sqrt{T-n-(q-1)(L-1)}\max(|\Gamma_R^*(q,\mu_b,\nu_b)|,$ $|\Gamma_I^*(q,\mu_b,\nu_b)|)$, we can further obtain that, 
\begin{eqnarray}\label{eqstep1star}
	\widehat{S}=S^*+o_p(\log^{-1/2} (T)).
\end{eqnarray}

In Step 2, we show that, for any $z\in \mathbb{R}$ and any sufficiently small $\varepsilon>0$, 
\begin{align*}
	\prob(S^*\le z) &\ge \prob\left( \|N(0,V_0)\|_{\infty} \le z-\varepsilon\log^{-1/2} (T) \right) - o(1),\\
	\prob(S^*\le z) &\le \prob\left( \|N(0,V_0)\|_{\infty} \le z+\varepsilon\log^{-1/2} (T) \right) + o(1),
\end{align*}
where the matrix $V_0$ is defined later. Combining it with \eqref{eqstep1star}, we obtain that, 
\begin{align} \label{eqstep2}
	\begin{split}
		\prob(\widehat{S}\le z) & \ge \prob\left( \|N(0,V_0)\|_{\infty}\le z-2\varepsilon\log^{-1/2} (T) \right) - o(1),\\
		\prob(\widehat{S}\le z) & \le \prob\left( \|N(0,V_0)\|_{\infty}\le z+2\varepsilon\log^{-1/2} (T) \right) + o(1).
	\end{split}
\end{align}

In Step 3, we show that $\|V_0-\widehat{V}\|_{\infty,\infty}=O((T)^{-c^{**}})$ for some $c^{**}>0$ with probability tending to $1$, where $\|\cdot\|_{\infty,\infty}$ denotes the element-wise max-norm. This, together with \eqref{eqstep2}, yields that, for any sufficiently small $\varepsilon>0$, with probability tending to $1$,
\begin{align*}
	\prob(\widehat{S}\le z) & \ge \prob\left( \|N(0,\widehat{V})\|_{\infty}\le z-2\varepsilon\log^{-1/2} (T)|\widehat{V} \right) - o(1),\\
	\prob(\widehat{S}\le z) & \le \prob\left( \|N(0,\widehat{V})\|_{\infty}\le z+2\varepsilon\log^{-1/2} (T)|\widehat{V} \right) + o(1), 
\end{align*}
where $\prob(\cdot|\widehat{V})$ is the conditional probability given $\widehat{V}$.  Setting $z=\widehat{c}_{\alpha}$, \change{which is defined in Equation \eqref{eqn:critical-value} of the paper}, we obtain that, with probability tending to $1$,
\begin{align}\label{eqstep3.1}
	\begin{split}
		\prob(\widehat{S} & \le \widehat{c}_{\alpha})\ge \prob\left( \|N(0,\widehat{V})\|_{\infty}\le \widehat{c}_{\alpha}-2\varepsilon\log^{-1/2} (T)|\widehat{V} \right) - o(1),\\
		\prob(\widehat{S} & \le \widehat{c}_{\alpha})\le \prob\left( \|N(0,\widehat{V})\|_{\infty}\le \widehat{c}_{\alpha}+2\varepsilon\log^{-1/2} (T)|\widehat{V} \right) + o(1),
	\end{split}
\end{align}
We next show that, with probability tending to $1$, the diagonal elements in $\widehat{V}$ are bounded away from zero, since the diagonal elements in $V_0$ are bounded away from zero given the conditions in Theorem \ref{thm4}. It follows from Theorem 1 of \cite{chernozhukov2017detailed} that, conditioning on $\widehat{V}$, with probability tending to $1$,
\begin{eqnarray*}
	\prob(\|N(0,\widehat{V})\|_{\infty}\le \widehat{c}_{\alpha}+2\varepsilon\log^{-1/2} (T)|\widehat{V})-\prob(\|N(0,\widehat{V})\|_{\infty}\le \widehat{c}_{\alpha}-2\varepsilon\log^{-1/2} (T)|\widehat{V}) \\
	\le o(1) \varepsilon \log^{1/2}(BQ) \log^{-1/2}(T),
\end{eqnarray*}
where $o(1)$ denotes some positive constant that is independent of $\varepsilon$. Under the given conditions on $B$ and $Q$, we obtain that, with probability tending to $1$, 
\begin{eqnarray*}
	\prob(\|N(0,\widehat{V})\|_{\infty}\le \widehat{c}_{\alpha}+2\varepsilon\log^{-1/2} (T)|\widehat{V})-\prob(\|N(0,\widehat{V})\|_{\infty}\le \widehat{c}_{\alpha}-2\varepsilon\log^{-1/2} (T)|\widehat{V})\le C^*\varepsilon,
\end{eqnarray*}
for some constant $C^*>0$. Combining with \eqref{eqstep3.1}, we obtain that, with probability tending to $1$, 
\begin{eqnarray*}
	\left| \prob(\widehat{S}\le \widehat{c}_{\alpha})-\prob\left( \|N(0,\widehat{V})\|_{\infty}\le \widehat{c}_{\alpha}|\widehat{V} \right) \right| \le C^*\varepsilon+o(1).
\end{eqnarray*}
This proves the validity of our test because $\varepsilon$ can be made arbitrarily small. 

In the following, we present each step in detail. We denote $\widehat{\varphi}_R^{(\ell)}$, $\widehat{\varphi}_I^{(\ell)}$ as the real and imaginary part of $\widehat{\varphi}^{(\ell)}$,  respectively. \change{According to the definition, we have that the absolute values of $\widehat{\varphi}_R^{(\ell)}$, $\widehat{\varphi}_I^{(\ell)}$ are uniformly bounded by $1$, and we treat $\{\mu_b,\nu_b\}_{1\le b\le B}$ as fixed throughout the proof. }

\paragraph{Step 1.}
Consider the decomposition that, for any $q,\mu,\nu$,
\begin{eqnarray*}
	\widehat{S}(q,\mu,\nu)=S^*(q,\mu,\nu)+R_1(q,\mu,\nu)+R_2(q,\mu,\nu)+R_3(q,\mu,\nu),
\end{eqnarray*}
where the remainder terms $R_1,R_2$ and $R_3$ are,
\begin{align*}
	R_1(q,\mu,\nu) = \; & \frac{1}{T-n-(q-1)(L-1)} \sum_{\ell=1}^{L-1}  \sum_{t=1}^{n-q+1}\{
	\varphi^{*}(\mu|X_{\ell \cdot n+t+q})- \\ &\widehat{\varphi}^{(\ell)}(\mu|X_{\ell \cdot n+t+q}) \}\{\psi^*(\nu|X_{\ell \cdot n+t})-\widehat{\psi}^{(\ell)}(\nu|X_{\ell \cdot n+t})\},\\
	R_2(q,\mu,\nu) = \; & \frac{1}{T-n-(q-1)(L-1)} \sum_{\ell=1}^{L-1}  \sum_{t=1}^{n-q+1}\{
	\exp(i\mu^\top X_{\ell \cdot n+t+q+1})
	-  \\ & \varphi^{*}(\mu|X_{\ell \cdot n+t+q}) \}\{\psi^*(\nu|X_{\ell \cdot n+t})-\widehat{\psi}^{(\ell)}(\nu|X_{\ell \cdot n+t})\},\\
	R_3(q,\mu,\nu) = \; & \frac{1}{T-n-(q-1)(L-1)} \sum_{\ell=1}^{L-1}  \sum_{t=1}^{n-q+1}\{
	\varphi^{*}(\mu|X_{\ell \cdot n+t+q})- \\ &\widehat{\varphi}^{(\ell)}(\mu|X_{\ell \cdot n+t+q}) \}\{\exp(i\nu^\top X_{\ell \cdot n+t-1})-\psi^*(\nu|X_{\ell \cdot n+t})\}.
\end{align*}
So it suffices to show that, 
\begin{eqnarray}\label{eqstep1eq1}
	\max_{b\in \{1,\cdots,B\}}\max_{q\in \{2,\cdots,Q\}} \sqrt{T-n-(q-1)(L-1)} |R_m(q,\mu_b,\nu_b)|=o_p(\log^{-1/2} (T)),
\end{eqnarray}
for $m=1,2,3$. In the following, we show \eqref{eqstep1eq1} holds with $m=1$ and $m=2$, respectively. When $m=3$, it can be shown similarly.  

First, we show that \eqref{eqstep1eq1} holds when $m=1$. Since $L$ is fixed, it suffices to show that, 
\begin{eqnarray}\label{eqstep1eq2}
	\max_{b\in \{1,\cdots,B\}} \max_{q\in \{2,\cdots,Q\}} \sqrt{T-n-(q-1)(L-1)}|R_{1,\ell}(q,\mu_b,\nu_b)|=o_p(\log^{-1/2} (T)),
\end{eqnarray}
where $R_{1,\ell}(q,\mu_b,\nu_b)$ is of the form,  
\begin{eqnarray*}
	\frac{1}{T-n-(q-1)(L-1)}   \sum_{t=1}^{n-q+1}\{
	\varphi^{*}(\mu_b|X_{\ell \cdot n+t+q})-  \widehat{\varphi}^{(\ell)}(\mu_b|X_{\ell \cdot n+t+q}) \}\{\psi^*(\nu_b|X_{\ell \cdot n+t})-\widehat{\psi}^{(\ell)}(\nu_b|X_{\ell \cdot n+t})\}.
\end{eqnarray*}
Similarly, let $\varphi^*_R$ and $\varphi^*_I$ denote the real and imaginary part of $\varphi^*$, respectively. We can rewrite $R_{1,\ell}(q,\mu_b,\nu_b)$ as $R_{1,\ell}^{(1)}(q,\mu_b,\nu_b)-R_{1,\ell}^{(2)}(q,\mu_b,\nu_b)+iR_{1,\ell}^{(3)}(q,\mu_b,\nu_b)+iR_{1,\ell}^{(4)}(q,\mu_b,\nu_b)$, where
\begin{align*}
	R_{1,\ell}^{(1)}(q,\mu_b,\nu_b) = \; & \frac{1}{(T-n-(q-1)(L-1))}   \sum_{t=1}^{n-q+1}\{
	\varphi^{*}_R(\mu_b|X_{\ell \cdot n+t+q})-\\ & \widehat{\varphi}^{(\ell)}_R(\mu_b|X_{\ell \cdot n+t+q}) \}\{\psi^*_R(\nu_b|X_{\ell \cdot n+t})-\widehat{\psi}^{(\ell)}_R(\nu_b|X_{\ell \cdot n+t})\},\\
	R_{1,\ell}^{(2)}(q,\mu_b,\nu_b) = \; & \frac{1}{(T-n-(q-1)(L-1))}   \sum_{t=1}^{n-q+1}\{
	\varphi^{*}_I(\mu_b|X_{\ell \cdot n+t+q})- \\ & \widehat{\varphi}^{(\ell)}_I(\mu_b|X_{\ell \cdot n+t+q}) \}\{\psi^*_I(\nu_b|X_{\ell \cdot n+t})-\widehat{\psi}^{(\ell)}_I(\nu_b|X_{\ell \cdot n+t})\},\\
	R_{1,\ell}^{(3)}(q,\mu_b,\nu_b) = \; & \frac{1}{(T-n-(q-1)(L-1))}   \sum_{t=1}^{n-q+1}\{
	\varphi^{*}_R(\mu_b|X_{\ell \cdot n+t+q})-\\ 
	& \widehat{\varphi}^{(\ell)}_R(\mu_b|X_{\ell \cdot n+t+q}) \}\{\psi^*_I(\nu_b|X_{\ell \cdot n+t})-\widehat{\psi}^{(\ell)}_I(\nu_b|X_{\ell \cdot n+t})\},\\
	R_{1,\ell}^{(4)}(q,\mu_b,\nu_b) = \; & \frac{1}{(T-n-(q-1)(L-1))}   \sum_{t=1}^{n-q+1}\{
	\varphi^{*}_I(\mu_b|X_{\ell \cdot n+t+q})-\\ & \widehat{\varphi}^{(\ell)}_I(\mu_b|X_{\ell \cdot n+t+q}) \}\{\psi^*_R(\nu_b|X_{\ell \cdot n+t})-\widehat{\psi}^{(\ell)}_R(\nu_b|X_{\ell \cdot n+t})\}.
\end{align*}
To prove \eqref{eqstep1eq2}, it suffices to show that, 
\begin{eqnarray}\label{eqstep1eq3}
	\max_{b\in \{1,\cdots,B\}} \max_{q\in \{2,\cdots,Q\}} \sqrt{T-n-(q-1)(L-1)}|R_{1,\ell}^{(s)}(q,\mu_b,\nu_b)|=o_p(\log^{-1/2} (T)),
\end{eqnarray}
for $s=1,2,3,4$. In the following, we show that \eqref{eqstep1eq3} holds when $s=1$. When $s=2,3,4$, it can be shown similarly. 

By Cauchy-Schwarz inequality, it suffices to show that, 
\begin{align}\label{eqstep1eq4}
	\begin{split}
		\max_{b\in \{1,\cdots,B\}} \max_{q\in \{2,\cdots,Q\}}\frac{1}{\sqrt{T-n-(q-1)(L-1)}} 
		\sum_{t=1}^{n-q+1} & \{ 
		\varphi^{*}_R(\mu_b|X_{\ell \cdot n+t}) - \widehat{\varphi}^{(\ell)}_R(\mu_b|X_{\ell \cdot n+t}) \}^2 \\ 
		& = o_p(\log^{-1/2} (T)),
	\end{split}
	\\ \label{eqstep1eq5}
	\begin{split}
		\max_{b\in \{1,\cdots,B\}} \max_{q\in \{2,\cdots,Q\}}\frac{1}{\sqrt{T-n-(q-1)(L-1)}}  
		\sum_{t=1}^{n-q+1} & \{\psi^*_R(\nu_b|X_{\ell \cdot n+t}) - \widehat{\psi}^{(\ell)}_R(\nu_b|X_{\ell \cdot n+t})\}^2 \\ 
		& = o_p(\log^{-1/2} (T)).
	\end{split}
\end{align}
In the following, we prove \eqref{eqstep1eq4}. The proof of \eqref{eqstep1eq5} is similar and is thus omitted. 

By Assumption \ref{assump1}, $\{X_{t}\}_{t\ge 0}$ is exponentially $\beta$-mixing, with the coefficient $\beta(t) = O(\rho^t)$. Let $\phi_{\ell,t,b}$ denote $\varphi^{*}_R(\mu_b|X_{\ell \cdot n+t})-\widehat{\varphi}^{(\ell)}_R(\mu_b|X_{\ell \cdot n+t})$. We have that, 
\begin{eqnarray}\label{eq1}
	\max_{ t,b}\Mean^{X_{\ell \cdot n+t}} \phi_{ t,b}^4\le 4\max_{b\in \{1,\cdots,B\}} \int_x \{\varphi^{*}_R(\mu_b|x)-\widehat{\varphi}^{(\ell)}_R(\mu_b|x)\}^2\mathbb{F}(dx)\equiv \Delta,
\end{eqnarray}
where the expectation $\Mean^{X_{\ell \cdot n+t}}$ is taken with respect to $X_{\ell \cdot n+t}$. Note that $\Delta$ is a random variable that depends on $\{\mu_b,\nu_b\}_{1\le b\le B}$ and $\{X_{t}\}_{t\in \mathcal{I}^{(\ell)}}$. By \eqref{eq1}, we have, 
\begin{eqnarray*}
	\max_{t,b}\Mean^{X_{\ell \cdot n+t}} (\phi_{\ell,t,b}^2-\Mean^{X_{\ell \cdot n+t}} \phi_{\ell,t,b}^2)^2\le \Delta.
\end{eqnarray*}
\change{By the boundedness assumption, we have $|\phi_{\ell,t,b}|\le 2$. Therefore, $|\phi_{\ell,t,b}^2-\Mean^{X_{\ell \cdot n+t}} \phi_{\ell,t,b}^2|\le \text{max}\{\phi_{\ell,t,b}^2,\Mean^{X_{\ell \cdot n+t}} \phi_{\ell,t,b}^2\}  \leq 4$. }

\change{Similar to the idea of Step 2 of the proof for Theorem \ref{mdn_bound} using Berbee’s lemma,} we have that, for any integers $\tau\ge 0$ and $1<p<T/2$,
\begin{eqnarray*}
	\prob\left(\left.\left| \sum_{t=1}^{n-q+1} (\phi_{\ell,t,b}^2-\Mean^{X_{0}} \phi_{\ell,t,b}^2) \right|\ge 6\tau\right|\Delta \right)\le \frac{T}{p}\beta(p)+\prob\left( \left.\left|\sum_{t\in \mathcal{I}_r}(\phi_{\ell,t,b}^2-\Mean^{X_{0}} \phi_{\ell,t,b}^2) \right|\ge \tau\right|\Delta \right)\\
	+ 4\exp\left( -\frac{\tau^2/2}{(n-q+1)p\Delta+4p\tau/3} \right),
\end{eqnarray*}
where $\mathcal{I}_r$ denotes the last $T-p\floor{T/p}$ elements in the list $\{X_t\}_{t \in \mathcal{I}^{(l)}}$, and $\floor{z}$ denotes the largest integer that is smaller than or equal to $z$. Suppose $\tau\ge 4p$. Noting  that $|\mathcal{I}_r|\le p$, we have, 
\begin{eqnarray*}
	\prob\left( \left.\left|
	\sum_{t\in \mathcal{I}_r}(\phi_{\ell,t,b}^2-\Mean^{X_{0}} \phi_{\ell,t,b}^2) \right| 
	\ge \tau \right|\Delta\right)=0.
\end{eqnarray*}
Since $\beta(t)=O(\rho^t)$, setting $p=-(c^*+3)\log ( T)/\log \rho$, we obtain that $T\beta(p)/p=O(T^{-2}B^{-1})=O(B^{-1}Q^{-1}T^{-2})$, because $Q\le T$, $B=O((T)^{c_*})$. Here, the big-$O$ notation is uniform in $b\in \{1,\cdots,B\}$, and $q\in \{2,\cdots,Q\}$. Setting $\tau=\max\{3\sqrt{\Delta Tp \log (BT)}, 11p\log(BT)\}$, we obtain that, 
\begin{eqnarray*}
	\frac{\tau^2}{4}\ge 2(n-q+1)p\Delta\log (BT), \quad \frac{\tau^2}{4}\ge  8p\tau\log (BT)/3, \quad \tau \ge 4p,
\end{eqnarray*}
as $T\to \infty$. It follows that $\tau^2/(2(n-q+1)p\Delta+8p\tau/3)\ge 2\log (BT)$. Therefore, 
\begin{eqnarray*}
	\max_{b\in \{1,\cdots,B\}}\max_{q\in \{2,\cdots,Q\}}\prob\left(\left.\left| \sum_{t=1}^T (\phi_{\ell,t,b}^2-\Mean^{X_{0}} \phi_{\ell,t,b}^2) \right|\ge 6\tau\right|\Delta \right)=O(B^{-1}Q^{-1}T^{-1}).
\end{eqnarray*}
By Bonferroni's inequality, we obtain that, 
\begin{eqnarray*}
	\prob\left(\left.\max_{b\in \{1,\cdots,B\}}\max_{q\in \{2,\cdots,Q\}}\left| \sum_{t=1}^T (\phi_{\ell,t,b}^2-\Mean^{X_{0}} \phi_{\ell,t,b}^2) \right|\ge 6\tau\right|\Delta \right)=O(T^{-1}).
\end{eqnarray*}
Therefore, with probability $1-O(T^{-1})$, we have that, 
\begin{eqnarray}\label{eq2}
	\max_{b\in \{1,\cdots,B\}}\max_{q\in \{2,\cdots,Q\}}\left| \sum_{t=1}^T (\phi_{\ell,t,b}^2-\Mean^{X_{0}} \phi_{\ell,t,b}^2) \right|=O(\sqrt{\Delta T}\log (BT),\log^2 (BT)).
\end{eqnarray}
Under the given condition on $Q$, $T-n-(q-1)(L-1)$ is proportional to $T$ for any $q\le Q$. Since we assume the convergence rate for $ \widehat{\varphi}$ and $ \widehat{\psi}$ is both $O(T^{-\kappa_0})$ for some $\kappa_0>1/4$, combining condition on $B$ with \eqref{eq2} yields \eqref{eqstep1eq4}. Specifically,  because of the convergence rate condition for $ \widehat{\varphi}$ and $ \widehat{\psi}$,  we know that $\Delta$ converges to 0 in the rate of $T^{-2\kappa_0}\ll T^{-1/2}$.  This shows that \eqref{eq2} convergences in the rate of at least $o_p(\log^{-1/2}(T))$.  In addition,  such convergence rate condition also implies that $\Mean^{X_{0}} \phi_{\ell,t,b}^2$ convergences in the rate of $T^{-2\kappa_0}$.  As a result, this term itself is negligible and  \eqref{eqstep1eq4} is thus proved.

Next, we show that \eqref{eqstep1eq1} holds when $m=2$. Similar to the proof of \eqref{eqstep1eq2}, it suffices to show that one of the following holds,
\begin{align*}
	\max_{q,b} \sqrt{T-n- (q-1)(L-1)} |R_{2,\ell}(q,\mu_b,\nu_b)| & = o_p(\log^{-1/2} (T)), \textrm { or } \\
	\max_{q,b} \sqrt{T-n-(q-1)(L-1)}|R_{2,\ell}^{(r)}(q,\mu_b,\nu_b)| & = o_p(\log^{-1/2} (T)),
\end{align*}
for any $\ell=1,\cdots,L$ and $r=1,2,3,4$, where
\vspace{-0.01in}
\begin{eqnarray*}
	R_{2,\ell}(q,\mu,\nu)=\frac{1}{T-n-(q-1)(L-1)}   \sum_{t=1}^{n-q+1}\{
	\exp(i\mu^\top X_{\ell \cdot n+t+q-1}) \\
	-\varphi^{*}(\mu|X_{\ell \cdot n+t+q-2}) \}  \{\psi^*(\nu|X_{\ell \cdot n+t})-\widehat{\psi}^{(\ell)}(\nu|X_{\ell \cdot n+t})\},\\
	R_{2,\ell}^{(1)}(q,\mu,\nu)=\frac{1}{T-n-(q-1)(L-1)}   \sum_{t=1}^{n-q+1}\{
	\cos(\mu^\top X_{\ell \cdot n+t+q-1})  \\
	-\varphi^{*}_R(\mu|X_{\ell \cdot n+t+q-2}) \} \{\psi^*_R(\nu|X_{\ell \cdot n+t})-\widehat{\psi}^{(\ell)}_R(\nu|X_{\ell \cdot n+t})\},\\
	R_{2,\ell}^{(2)}(q,\mu,\nu)=\frac{1}{T-n-(q-1)(L-1)}   \sum_{t=1}^{n-q+1}\{
	\sin(\mu^\top X_{\ell \cdot n+t+q-1})  \\
	-\varphi^{*}_I(\mu|X_{\ell \cdot n+t+q-2}) \} \{\psi^*_I(\nu|X_{\ell \cdot n+t})-\widehat{\psi}^{(\ell)}_I(\nu|X_{\ell \cdot n+t})\},\\
	R_{2,\ell}^{(3)}(q,\mu,\nu)=\frac{1}{T-n-(q-1)(L-1)}   \sum_{t=1}^{n-q+1}\{
	\cos(\mu^\top X_{\ell \cdot n+t+q-1}) \\
	-\varphi^{*}_R(\mu|X_{\ell \cdot n+t+q-2}) \}  \{\psi^*_I(\nu|X_{\ell \cdot n+t})-\widehat{\psi}^{(\ell)}_I(\nu|X_{\ell \cdot n+t})\},\\
	R_{2,\ell}^{(4)}(q,\mu,\nu)=\frac{1}{T-n-(q-1)(L-1)}   \sum_{t=1}^{n-q+1}\{
	\sin(\mu^\top X_{\ell \cdot n+t+q-1}) \\
	-\varphi^{*}_I(\mu|X_{\ell \cdot n+t+q-2}) \}  \{\psi^*_R(\nu|X_{\ell \cdot n+t})-\widehat{\psi}^{(\ell)}_R(\nu|X_{\ell \cdot n+t})\}.
\end{eqnarray*}
In the following, we only show $\max_{q,b} \sqrt{T-n-(q-1)(L-1)}|R_{2,\ell}^{(1)}(q,\mu_b,\nu_b)|=o_p(\log^{-1/2} (T))$. The proofs of the rest of the cases are similar, and are thus omitted. 

Let $\mathcal{F}^{(0)}_q=\{X_{1+n},X_{2+n},\cdots,X_{q-1+n}\}\cup \{X_{t}:t \in \bar{\mathcal{I}}^{(\ell)} \}\cup \{\mu_1,\cdots,\mu_B,\nu_1,\cdots,\nu_B\}$. Then we recursively define $\mathcal{F}^{(g)}_q$ as $\mathcal{F}^{(g)}_q=\mathcal{F}^{(g-1)}_q\cup \{X_{g+q-1+\ell n}\}$ for any $1\le g\le n-q+1$. Let $\phi_{\ell,g,q,b}^*=\{\cos(\mu_b^\top X_{g+q-1+\ell n}) - \varphi^{*}_R(\mu_b|X_{g+q-2+\ell n}) \}\{\psi^*_R(\nu_b|X_{g+\ell n})-\widehat{\psi}^{(\ell)}_R(\nu_b|X_{g+\ell n})\}$.  Under the Markov property, $R_{2,\ell}^{(1)}(q,\mu_b,\nu_b)$ can be rewritten as $\{T-n-(q-1)(L-1)\}^{-1}\sum_{g=1}^{n-q+1} \phi^*_{\ell,g,q,b}$, and forms a sum of martingale difference sequence with respect to the filtration $\{\sigma(\mathcal{F}^{(g)}_q):g\ge 0\}$, where $\sigma(\mathcal{F}^{(g)}_q)$ denotes the $\sigma$-algebra generated by the variables in $\mathcal{F}^{(g)}_q$. In the following, we apply concentration inequalities for martingales to bound $\max_{q,b} |R_{2,\ell}^{(1)}(q,\mu_b,\nu_b)|$. 

Under the boundedness condition, we have $|\phi_{\ell,g,q,b}^*|^2\le 4 \{\psi^*_R(\nu_b|X_{g+\ell n})-\widehat{\psi}^{(\ell)}_R(\nu_b|X_{g+\ell n})\}^2$. In addition, by the Markov property, we have that,
\begin{eqnarray*}
	\Mean\{(\phi_{g+1,q,b}^{*})^2|\sigma(\mathcal{F}^{(g)}_q)\}=\Mean [\{\cos(\mu_b^\top X_{g+q-1+\ell n})
	-\varphi^{*}_R(\mu_b|X_{g+q-2+\ell n}) \}^2|X_{g+q-2+\ell n}]\\ \times \{\psi^*_R(\nu_b|X_{g+\ell n})-\widehat{\psi}^{(\ell)}_R(\nu_b|X_{g+\ell n})\}^2
	\le 4\{\psi^*_R(\nu_b|X_{g+\ell n})-\widehat{\psi}^{(\ell)}_R(\nu_b|X_{g+\ell n})\}^2. 
\end{eqnarray*}
It follows from Theorem 2.1 of \citet{Bercu2008} that, for any $y$ and $\tau$, 
\begin{eqnarray*}
	\prob\left(\left|\sum_{g=1}^{n-q+1} \phi^*_{\ell,g,q,b}\right|\ge\tau, \sum_{g=1}^{n-q+1} 4\{\psi^*_R(\nu_b|X_{g+\ell n}) -\widehat{\psi}^{(\ell)}_R(\nu_b|X_{g+\ell n})\}^2\le y \right) \le 2\exp\left(-\frac{\tau^2}{2y}\right).
\end{eqnarray*}
Therefore, for any $y$ and $\tau$, 
\begin{eqnarray*}
	\prob\left( \left|\sum_{g=1}^{n-q+1} \phi^*_{\ell,g,q,b}\right|\ge\tau, \max_{b\in\{1,\cdots,B\}}  \sum_{g=1}^{n-q+1} \{\psi^*_R(\nu_b|X_{g+\ell \cdot n})- \widehat{\psi}^{(\ell)}_R(\nu_b|X_{g+\ell \cdot n})\}^2\le \frac{y}{4} \right)  \le 2\exp\left(-\frac{\tau^2}{2y}\right).
\end{eqnarray*}
By Bonferroni's inequality, we obtain that, for any $y$ and $\tau$, 
\begin{eqnarray*}
	\prob \Bigg( \max_{\substack{q\in \{2,\cdots,Q\} \\ b\in \{1,\cdots,B\} }}\left|\sum_{g=1}^{n-q+1} \phi^*_{\ell,g,q,b}\right|\ge\tau, \max_{b\in\{1,\cdots,B\}}  \sum_{t=1}^{n-q+1} \{\psi^*_R(\nu_b|X_{\ell \cdot n+t}) \\ 
	- \widehat{\psi}^{(\ell)}_R(\nu_b|X_{\ell \cdot n+t})\}^2\le \frac{y}{4} \Bigg) \le 2BQ\exp\left(-\frac{\tau^2}{2y}\right).
\end{eqnarray*}
Setting $y=4\sqrt{T}$, we obtain that, 
\begin{eqnarray*}
	\prob\Bigg( \max_{\substack{q\in \{2,\cdots,Q\} \\ b\in \{1,\cdots,B\} }}\left|\sum_{g=1}^{n-q+1} \phi^*_{\ell,g,q,b}\right|\ge\tau, \max_{b\in\{1,\cdots,B\}}  \sum_{t=1}^{n-q+1} \{\psi^*_R(\nu_b|X_{\ell \cdot n+t})  \\ 
	- \widehat{\psi}^{(\ell)}_R(\nu_b|X_{\ell \cdot n+t})\}^2\le \sqrt{T} \Bigg) \le 2BQ\exp\left(-\frac{\tau^2}{8\sqrt{T}} \right),
\end{eqnarray*}
For any sufficiently small $\varepsilon>0$, it follows from \eqref{eqstep1eq5} that, 
\begin{eqnarray}\label{eq3}
	\prob\left(\max_{\substack{q\in \{2,\cdots,Q\} \\ b\in \{1,\cdots,B\} }}\left|\sum_{g=1}^{n-q+1} \phi^*_{\ell,g,q,b}\right|\ge\tau\right)\le 2BQ\exp\left(-\frac{\tau^2}{8 \sqrt{T}}\right)+o(1).
\end{eqnarray}
\change{Setting $\tau=T^{1/4}\sqrt{8\log (BQT)}$, the right-hand-side of \eqref{eq3} is $o(1)$. Under the given conditions on $B$ and $Q$, we obtain that}
\begin{align*}
	\max_{q,b} \sqrt{T-n-(q-1)(L-1)}|R_{2,\ell}^{(1)}(q,\mu_b,\nu_b)|=o_p(\log^{-1/2} (T)). 
\end{align*}

\paragraph{Step 2.}
For any  $0< t<n-q+1$, define the vectors $\lambda_{R,\ell,q,t}^*,\lambda_{I,\ell,q,t}^*\in \mathbb{R}^{B}$, whose $b$-th element corresponds to the real and imaginary part of 
\begin{eqnarray*}
	\frac{1}{\sqrt{T-n-(q-1)(L-1)}}\{\exp(i\mu_b^\top X_{\ell \cdot n+t+q-1})
	-\varphi^{*}(\mu_b|X_{\ell \cdot n+t+q-2}) \} \\ \{\exp(i\nu_b^\top X_{\ell \cdot n+t-1})-\psi^{*}(\nu_b|X_{\ell \cdot n+t})\},
\end{eqnarray*}
respectively. Let $\lambda^*_{q,t}$ denote the (2B)-dimensional vector $(\lambda_{R,\ell,q,t}^{*\top},\lambda_{I,\ell,q,t}^{*\top})^\top$. In addition, define the (2B(Q+1))-dimensional vector $\lambda_{t}^*$ as $(\lambda_{0,t}^{*\top}, \lambda_{1,t-1}^{*\top}\mathbb{I}(t>1), \cdots,\lambda_{Q,t-Q}^{*\top}\mathbb{I}(t>Q))^\top$.

For any $1\le g\le T-n-(q-1)(L-1)$, let $\mathcal{F}^{(0)}=\{X_{1}\}\cup \{\mu_1,\cdots,\mu_B,\nu_1,\cdots,\nu_B\}$ and recursively define $\mathcal{F}^{(g)}=\mathcal{F}^{(g-1)}\cup \{X_g\}$. The vector $M_{n,T}=\sum_{g=1}^{T-n-(L-1)} \lambda_g^*$ forms a sum of martingale difference sequence with respect to the filtration $\{\sigma(\mathcal{F}^{(g)}):g\ge 0 \}$. Note that $S^*=\|\sum_{g=1}^{T-n-(L-1)} \lambda_g^*\|_{\infty}$. In this step, we apply the high-dimensional martingale central limit theorem of \citet{belloni2018} to establish the limiting distribution of $S^*$. \change{A similar result as that in \citet{belloni2018} is also given in \cite{chernozhukov2013gaussian}.}

For $1\le g\le T-n-(L-1)$, let
\begin{eqnarray*}
	\Sigma_g=\sum_{g=1}^{T-n-(L-1)} \Mean \left(\left.\lambda_g^* \lambda_g^{*\top}\right|\mathcal{F}^{(g-1)} \right), \quad \textrm{ and } \quad V^*=\sum_{g=1}^{T-n-(L-1)} \Sigma_g. 
\end{eqnarray*}
Using similar arguments as in proving \eqref{eq2}, we can show that, with probability $1-O(T^{-1})$, $\|V^*-V_0\|_{\infty,\infty}=O((T)^{-1/2}\log (BT))+O((T)^{-1}\log^2 (BT))$, where $V_0=\Mean(V^*)$. Under the given conditions on $B$, we have $\|V^*-V_0\|_{\infty,\infty}\le \kappa_{B,T}$, for some $\kappa_{B,T}=O((T)^{-1/2}\log (T))$, with probability $1-O(T^{-1})$. In addition, under the boundedness assumption, all the elements in $V^*$ and $V_0$ are uniformly bounded by some constant. It then follows that, 
\begin{align*}
	\Mean \|V^*-V_0\|_{\infty,\infty}\le \kappa_{B,T}+\prob (\|V^*-V_0\|_{\infty,\infty}>\kappa_{B,T})=O((T)^{-1/2}\log (T)).
\end{align*}
By Theorem 3.1 of  \citet{belloni2018}, we have that, for any Borel set $\mathcal{R}$ and any $\delta>0$, 
\begin{align}\label{importantassertion}
	\begin{split}
		&\prob(S^*\in \mathcal{R})\le \prob (\|N(0,V_0)\|_{\infty}\in \mathcal{R}^{C\delta})| \\
		\le \; & C\left(\frac{1}{T}+\frac{\log (BT)\log (BQ)}{\delta^2\sqrt{T}}+\frac{\log^3 (BQ)}{\delta^3 \sqrt{T}}+\frac{\log^3(BQ)}{\delta^3}\sum_{g=1}^{T-n-(L-1)}\Mean \|\eta_g\|_{\infty}^3 \right),
	\end{split}
\end{align}
for some constant $C>0$. 

Under the boundedness assumption, the absolute value of each element in $\Sigma_g$ is uniformly bounded by $16 (T-n-(q-1)(L-1))^{-1}=O(T^{-1})$. With some direct calculation, we can show that $\sum_{g=1}^{T-n-(L-1)}\Mean \|\eta_g\|_{\infty}^3=O((T)^{-1/2}\log^{3/2} (BQ))$. In addition, we have $Q=O(T)$, and $B=O((T)^{c_*})$. Combining these together with \eqref{importantassertion} yields that
\begin{eqnarray}\label{importantassertion2}
	\prob(S^*\in \mathcal{R})\le \prob (\|N(0,V_0)\|_{\infty}\in \mathcal{R}^{C\delta})|+ O(1)\left(\frac{1}{T}+\frac{
		\log^2 (T)}{\delta^2\sqrt{T}}+\frac{\log^{9/2}(T)}{\delta^3\sqrt{T}} \right),
\end{eqnarray}
where $O(1)$ denotes some positive constant. 

Setting $\mathcal{R}=(z,+\infty)$ and $\delta=\varepsilon \log^{-1/2} (T)/C$, we obtain that,
\begin{eqnarray*}
	\prob(S^*\le z)\ge \prob(\|N(0,V_0)\|_{\infty}\le z-\varepsilon \log^{-1/2} (T))-o(1).
\end{eqnarray*}
Setting $\mathcal{R}=(-\infty,z]$, we can similarly show that, 
\begin{eqnarray*}
	\prob(S^*\le z)\le \prob(\|N(0,V_0)\|_{\infty}\le z+\varepsilon \log^{-1/2} (T))+o(1).
\end{eqnarray*}

\paragraph{Step 3.}
We break this step into two parts. First, we show $V_0$ is a block diagonal matrix. Specifically, let $V_{0,q_1,q_2}$ denote the $(2B)\times (2B)$ submatrix of $V_0$ formed by the rows in $\{2q_1B+1,2q_1B+2,\cdots,2(q_1+1)B\}$ and the columns in  $\{2q_2B+1,2q_2B+2,\cdots,2(q_2+1)B\}$. For any $q_1\neq q_2$, we show $V_{0,q_1,q_2}=O_{(2B)\times (2B)}$. Next, letting $\Sigma^{(q)}$ denote $V_{0,q,q}$, we establish an upper bound for $\max_{q\in \{2,\cdots,Q\}}\|\Sigma^{(q)}-\widehat{\Sigma}^{(q)}\|_{\infty,\infty}$. Let $\widehat{V}$ be a block diagonal matrix where the diagonal blocks are given by $\widehat{\Sigma}^{(0)},\widehat{\Sigma}^{(1)},\cdots,\widehat{\Sigma}^{(Q)}$, we obtain the bound for $\|V_0-\widehat{V}\|_{\infty,\infty}$

First, let $\lambda_{R,q,t,b}^*$ and $\lambda_{I,q,t,b}^*$ denote the $b$-th element of $\lambda_{R,\ell,q,t}^*$ and $\lambda_{I,q,t}^*$, respectively. Each element in $V_{0,q_1,q_2}$ equals $\Mean (\sum_{t} \lambda_{Z_1,q_1,t,b_1}^*) (\sum_{t} \lambda_{Z_2,q_2,t,b_2}^*)$, for some $b_1,b_2\in \{1,\cdots,B\}$ and $Z_1,Z_2\in \{R,I\}$. In the following, we show that,
\begin{eqnarray*}
	\Mean \left(\sum_{t} \lambda_{R,q_1,t,b_1}^*\right) \left(\sum_{t} \lambda_{R,q_2,t,b_2}^*\right)=0,\,\,\,\,\,\,\,\,\forall q_1\neq q_2.
\end{eqnarray*}
Similarly, we can show that $\Mean (\sum_{t} \lambda_{R,q_1,t,b_1}^*) (\sum_{t} \lambda_{I,q_2,t,b_2}^*)=0$, and $\Mean (\sum_{t} \lambda_{I,q_1,t,b_1}^*) (\sum_{t} \lambda_{I,q_2,t,b_2}^*)=0$, for any $q_1\neq q_2$.

Toward our goal, since the observations in different time series are i.i.d., it suffices to show,
\begin{eqnarray*}
	\sum_j \Mean \left(\sum_{t} \lambda_{R,q_1,t,b_1}^*\right) \left(\sum_{t} \lambda_{R,q_2,t,b_2}^*\right)=0,\,\,\,\,\,\,\,\,\forall q_1\neq q_2,
\end{eqnarray*}
or equivalently,
\begin{eqnarray}\label{eq4.5}
	\Mean \left(\sum_{t} \lambda_{R,q_1,0,t,b_1}^*\right) \left(\sum_{t} \lambda_{R,q_2,0,t,b_2}^*\right)=0,\,\,\,\,\,\,\,\,\forall q_1\neq q_2,
\end{eqnarray}
By definition, we have that, 
\begin{eqnarray*}
	\lambda_{R,q,0,t,b}^*=\frac{1}{\sqrt{T-n-(q-1)(L-1)}}\{\cos(\mu_b^\top X_{t+q-1+n})
	-\varphi_R^{*}(\mu_b|X_{t+q-2+n}) \} \\ \{\cos(\nu_b^\top X_{t-1+n})-\psi_R^{*}(\nu_b|X_{t+n})\}.
\end{eqnarray*}
Since $q_1\neq q_2$, for any $t_1,t_2$, we have either $t_1+q_1\neq t_2+q_2$, or $t_1\neq t_2$. Suppose $t_1+q_1>t_2+q_2$. Under the Markov property, we have that, for any $b$, 
\begin{eqnarray*}
	\Mean [\{\cos(\mu_b^\top X_{t_1+q_1-1+n}) - \varphi_R^{*}(\mu_b|X_{t_1+q_1-2+n})\}|\{X_{j}\}_{j\le t_1+q_1-2+n}]=0.
\end{eqnarray*}
Therefore, for any $b_1, b_2$, 
\begin{eqnarray}\label{eq4}
	\Mean \lambda_{R,q_1,0,t_1,b_1}^* \lambda_{R,q_2,0,t_2,b_2}^*=0.
\end{eqnarray}
Similarly, when $t_1+q_1<t_2+q_2$, we can show \eqref{eq4} holds as well. 

Suppose $t_1<t_2$, under $H_0$, we have that, for any $b$, 
\begin{eqnarray*}
	\Mean [\{\cos(\nu_b^\top X_{t_1-1+n}) - \varphi_R^{*}(\nu_b|X_{t_1+n})\}|\{X_{j}\}_{j\ge t_1+n}]=0,
\end{eqnarray*}
and hence \eqref{eq4} holds. Similarly, when $t_1>t_2$, we can show \eqref{eq4} holds as well. This yields \eqref{eq4.5}. 

Next, for any $q\in \{2,\cdots,Q\}$, we can represent $\widehat{\Sigma}^{(q)}-\Sigma^{(q)}$ by
\begin{eqnarray}\label{eq5}
	\sum_{\ell=1}^{L-1}  \sum_{t=1}^{n-q+1} \frac{(\lambda_{R,\ell,q,t}^\top,\lambda_{I,q,t}^\top)^\top  (\lambda_{R,\ell,q,t}^\top,\lambda_{I,q,t}^\top)-(\lambda_{R,\ell,q,t}^{*\top},\lambda_{I,q,t}^{*\top})^\top  (\lambda_{R,\ell,q,t}^{*\top},\lambda_{I,q,t}^{*\top})}{T-n-(q-1)(L-1)}.
\end{eqnarray}
Using similar arguments as in Step 1 of the proof, we can show that, with probability tending to $1$, the absolute value of each element in \eqref{eq5} is upper bounded by $c_0^*(T)^{-c^{**}}$, for any $q\in \{2,\cdots,Q\}$ and some constants $c_0,c^* > 0$. Therefore, we obtain that, with probability tending to $1$, $\max_{q\in \{2,\cdots,Q\}} \|\widehat{\Sigma}^{(q)}-\Sigma^{(q)}\|_{\infty,\infty}=O((T)^{-c^{**}})$. This completes the proof of Theorem \ref{thm4}. 
\eop

\subsubsection{Proof of Theorem \ref{thm5}}

Under the condition that $\sup_{q,\mu,\nu} S_0(q,\mu,\nu) \gg T ^{-1/2} \log^{1/2} (T) $, there exist some $q_0$, $\mu_0$ and $\nu_0$, such that $S_0(q,\mu_0,\nu_0) \gg T ^{-1/2}\log^{1/2} (T) $. Note that the objective function $S_0(q,\mu_0,\nu_0)$ is Lipschitz continuous in $\mu$ and $\nu$. As such, for any $\mu$ and $\nu$ within the interval $[\mu_0 - T ^{-1/2}\log^{1/2} (T)  ,\mu_0 + T ^{-1/2}\log^{1/2} (T)  ]$ and $[\nu_0 - T ^{-1/2}\log^{1/2} (T)  ,\nu_0 + T ^{-1/2} \log^{1/2} (T) ]$, we have that
\begin{align*}
	\sup_{q} S_0(q,\mu,\nu) \gg T ^{-1/2}\log^{1/2} (T). 
\end{align*}
Since each $\mu,\nu$ is independently normally distributed, the probability that $\mu,\nu$ falls into this interval is lower bounded by $c T ^{-1/2} \log^{1/2} (T)$, for some constant $c \geq 0$. Since we randomly generate $B$ pairs of $\mu,\nu$, the probability that at least one pair of $\mu,\nu$ falls into this interval is lower bounded by 
\begin{align*}
	1 - \{1-cT ^{-1/2}\log^{1/2} (T) \}^B \geq 1 - \exp\{-cB T ^{-1/2}\log^{1/2} (T) \}   
\end{align*}
The above probability tends to $1$ under the condition that $B=\kappa_1 T^{\kappa_2}$, for some $\kappa_2 \geq 1/2$. As a result, we obtain that,
\begin{align*}
	\max_{b \in \{1,2,....,B\}}  \max_{q\in \{2,\cdots,Q\}} S_0(q,\mu,\nu) \gg T ^{-1/2} \log^{1/2} (T)    
\end{align*}

Following similar arguments as in the proof of Theorem \ref{thm4}, we can show that, 
\begin{align*}
	\max_{b\in \{1,\cdots,B\}}\max_{q\in \{2,\cdots,Q\}} \sqrt{T-n-(q-1)(L-1)} |\widehat{S}(q,\mu_b,\nu_b)-S^*(q,\mu_b,\nu_b)|=o_p(\log^{-1/2} (T)). \\
	\max_{b\in \{1,\cdots,B\}}\max_{q\in \{2,\cdots,Q\}} \sqrt{T-n-(q-1)(L-1)} | S_0(q,\mu_b,\nu_b)-S^*(q,\mu_b,\nu_b)|=O_p(\log^{1/2}(T)).    
\end{align*}
It then follows that,
\begin{align*}
	\begin{split}
		\max_{b\in \{1,\cdots,B\}}\max_{q\in \{2,\cdots,Q\}} \sqrt{T-n-(q-1)(L-1)} |\widehat{S}(q,\mu_b,\nu_b)- S_0(q,\mu_b,\nu_b)| =O_p(\log^{1/2}(T)).
	\end{split}
\end{align*}
In addition, with some direct calculation, we have that, 
\begin{align*}
	& \max_{b\in \{1,\cdots,B\}}\max_{q\in \{2,\cdots,Q\}} \sqrt{T-n-(q-1)(L-1)} |\widehat{S}(q,\mu_b,\nu_b)- S_0(q,\mu_b,\nu_b)| \\
	\geq \; & \max_{b\in \{1,\cdots,B\}}\max_{q\in \{2,\cdots,Q\}} \sqrt{T-n-(q-1)(L-1)} \Big [ |S_0(q,\mu_b,\nu_b)| - |\widehat{S}(q,\mu_b,\nu_b)|\Big ]  \\
	\geq \; & \max_{b\in \{1,\cdots,B\}}\max_{q\in \{2,\cdots,Q\}} \sqrt{T-n-(q-1)(L-1)} |S_0(q,\mu_b,\nu_b)| \\ 
	& - \sqrt{T-n-(q_0+1)(L-1)} |\widehat{S}(q_0,\mu_{b_0},\nu_{b_0})|, \quad \text{for any } q_0,b_0\\
	\geq \; & \max_{b\in \{1,\cdots,B\}}\max_{q\in \{2,\cdots,Q\}} \sqrt{T-n-(q-1)(L-1)} |S_0(q,\mu_b,\nu_b)| \\ 
	& - \max_{b\in \{1,\cdots,B\}}\max_{q\in \{2,\cdots,Q\}} \sqrt{T-n-(q-1)(L-1)} |\widehat{S}(q,\mu_b,\nu_b)|
\end{align*}
To summarize, we have that, 
\begin{align*}
	& \max_{b\in \{1,\cdots,B\}}\max_{q\in \{2,\cdots,Q\}} \sqrt{T-n-(q-1)(L-1)} |\widehat{S}(q,\mu_b,\nu_b)|\\
	\geq \; & \max_{b\in \{1,\cdots,B\}}\max_{q\in \{2,\cdots,Q\}} \sqrt{T-n-(q-1)(L-1)} |S_0(q,\mu_b,\nu_b)| + O_p(\log^{1/2} T)
\end{align*}

Under the condition that $\max_{b \in \{1,2,....,B\}} \max_{q\in \{2,\cdots,Q\}} S_0(q,\mu,\nu) \gg  T ^{-1/2}\log^{1/2} (T)  $, we have $\widehat{S}  =  \max_{b\in \{1,\cdots,B\}}\max_{q\in \{2,\cdots,Q\}}$ $\sqrt{T-n-(q-1)(L-1)} \widehat{S}(q,\mu_b,\nu_b) \gg \log^{1/2}{(T)}$. 

Using similar arguments as in the proof of Theorem \ref{thm4}, we obtain that $\max_{q\in \{2,\cdots,Q\}} \|\widehat{\Sigma}^{(q)}-\Sigma^{(q)}\|_{\infty,\infty}=O(T^{-c^{**}})$, with probability tending to $1$, for some constant $c^{**}>0$. Therefore, there exist some constants $C,C'>0$, such that $\widehat{c}_{\alpha} < C \log^{1/2}(B) = C'\log^{1/2}(T)$. Since $\widehat{S} \gg \log^{1/2}{(T)}$, we have that $\prob(\widehat{S} > \widehat{c}_{\alpha}) \to 1$, as $T \to \infty$. This completes the proof of Theorem \ref{thm5}. 
\eop

\subsubsection{Auxiliary Lemmas}\label{sec:aux}

We present a set of auxiliary lemmas that are useful for our proofs.

The first lemma establishes the approximation bound of DNN, which is used to calculate the approximation error of the mixture density network. It follows directly from Lemma 7 in \cite{farrell2019deepneural}, and its proof is omitted. 

\begin{lemma} \label{lemma2}
	There exists a DNN class $\mathcal{F}_{\text{DNN}}$ with the ReLU activation, such that, for any $\epsilon>0$, 
	\begin{enumerate}[(i)]
		\item $\mathcal{F}_{\text{DNN}}$ approximates $W^{\gamma,\infty}([-1,1]^d)$, in the sense that, for any $g^* \in W^{\gamma,\infty}([-1,1]^d)$, there exists a $g_{\epsilon} \in \mathcal{F}_{\text{DNN}}$ and $C_1>0$, such that $\|g_{\epsilon}-g^*\|_{\infty} \leq \epsilon$. 
		
		\item $H(\epsilon) \leq C_1 (log(\frac{1}{\epsilon}) +1)$ and $W(\epsilon), U(\epsilon) \leq C_1 \cdot \epsilon^{-\frac{d}{\gamma}}(\log (\frac{1}{\epsilon})+1)$, where $H$ denotes the number of layers, $W$ the number of weights, and $U$ the total number of hidden units.
	\end{enumerate}
\end{lemma}

The next lemma establishes the approximation error bound for the mixture density network. This bound is affected by the approximation error in Assumption \ref{assump2} through the parameter $\omega_1$. The specific selection of $\epsilon$ in the main proof is a tradeoff between the approximation error and the Rademacher complexity.

\begin{lemma} \label{lemma4}
	Suppose Assumptions \ref{assump2} and \ref{assump3} hold. Then for any $\epsilon>0$ and integer $G\ge 1$, there exists a set of DNN functions $\{(g_{g1},g_{g2},g_{g3})_{g=1}^{G}\}$ whose network architectures depend on $\epsilon$ such that
	\begin{enumerate}[(i)]
		\item $ \|f^* - f_T\|_{\infty} \leq C'_2 G^{4\omega_2} \epsilon  +O(G^{-\omega_1}) \;\; \textrm{ where } \; f_T= \sum_{g=1}^G g_{g1}(x) \frac{1}{\sqrt{2 \pi }g_{g3}(x)}e^{-\frac{(y-g_{g2}(x))^2}{2 g^2_{g3}(x)}}, $
		for some constant $C'_2>0$ that is independent of $\epsilon$;
		\item the number of hidden layer $H_{gj}$, the number of parameters $W_{gj}$, and the total number of hidden units $U_{gj}$ satisfy that 
		\begin{enumerate}[(a)]
			\item $H_{gj} \leq C_1 \{ \log(G/\epsilon) +1 \}$
			\item $W_{gj} \leq C_1 G^{\frac{d}{\gamma}} \ \epsilon ^{-\frac{d}{\gamma}}\{ \log (G / \epsilon)+1 \}$
			\item $U_{gj} \leq C_1 G^{\frac{d}{\gamma}} \ \epsilon ^{-\frac{d}{\gamma}}\{ \log (G / \epsilon)+1 \}$
		\end{enumerate}
	\end{enumerate}
	We remark that $H_{gj}$, $W_{gj}$, and $U_{gj}$ depend on $\epsilon$, as specified in (ii). 
\end{lemma}

\noindent
\textbf{Proof}:
By Lemma \ref{lemma2}, for any $\epsilon>0$ and $G\ge 1$, there exists a set of DNN functions $\{(g_{g1},g_{g2},g_{g3})_{g=1}^{G}\}$, such that,
\begin{align} \label{eqn:3func}
	\|g_{g1}-\alpha^*_g\|_{\infty} \leq \frac{\epsilon}{G}, \;\; \|g_{g2}-\mu^*_g\|_{\infty} \leq \frac{\epsilon}{G}, \;\; \|g_{g3}-\sigma^*_g\|_{\infty} \leq \frac{\epsilon}{G}. 
\end{align}
Besides, $H_{gj}$ and $W_{gj}$ of the DNN functions class satisfy that $H_{gj} \leq C_1 \{ \log(G/\epsilon) +1 \}$, $W_{gj} \leq C_1 G^{\frac{d}{\gamma}} \ \epsilon ^{-\frac{d}{\gamma}}\{ \log (G / \epsilon)+1 \}$, and $U_{gj} \leq C_1 G^{\frac{d}{\gamma}} \ \epsilon ^{-\frac{d}{\gamma}}\{ \log (G / \epsilon)+1 \}$. Therefore, 
\begin{align*}
	\|f^* - f_T\|_{\infty} = \left\| \sum_{g=1}^{G} \alpha^*_g(x) \frac{1}{\sqrt{2 \pi}\sigma^*_g(x)} e^{-\frac{(y-\mu^*_g(x))^2}{2 {\sigma^*_g}^2(x)}}+O(G^{-\omega_1}) - \sum_{g=1}^G g_{g1}(x) \frac{1}{\sqrt{2 \pi }g_{g3}(x)} e^{-\frac{(y-g_{g2}(x))^2}{2 g^2_{g3}(x)}}  \right\|_{\infty}  
\end{align*}
Since $|y| \leq 1$, $|\mu^*_g(x)| \leq C_2$, and $C_2^{-1} G^{-\omega_2} \leq \sigma^*_g(x) \leq C_2$, there exists a constant $\widetilde{C}_1$, such that 
\begin{align*}
	\left\| \frac{1}{\sqrt{2 \pi}\sigma^*_g(x)} e^{-\frac{(y-\mu^*_g(x))^2}{2 {\sigma^*_g}^2(x)}} \right\|_{\infty} \leq \widetilde{C}_1 G^{\omega_2}. 
\end{align*}
By \eqref{eqn:3func} and since every differentiable function with bounded gradient is Lipschitz, there exists a constant $\widetilde{C}_2$, such that
\begin{align*}
	\left\| \frac{1}{\sqrt{2 \pi}\sigma^*_g(x)} e^{-\frac{(y-\mu^*_g(x))^2}{2 {\sigma^*_g}^2(x)}}-\frac{1}{\sqrt{2 \pi }g_{g2}(x)}e^{-\frac{(y-g_{g2}(x))^2}{2 g^2_{g3}(x)}} \right\|_{\infty} \leq \widetilde{C}_2 \frac{\epsilon G^{4\omega_2}}{G}. 
\end{align*}
Applying the triangle inequality, we obtain that there exists a constant $C'_2>0$ such that,   
\begin{align*}
	\|f^* - f_T\|_{\infty} \leq  C'_2 G^{4\omega_2} \epsilon +O(G^{-\omega_1})
\end{align*}
This completes the proof of Lemma \ref{lemma4}.
\eop
\bigskip

The next lemma connects the metric entropy between the function class of mixture density network and the deep neural network (DNN). The transition between the two metric entropies introduces a multiplier of $G$ as the cost. 

\begin{lemma}  \label{lemma5}
	Suppose Assumptions \ref{assump2} and \ref{assump3} hold. Then there exists a constant $C_3>0$, such that 
	\begin{align*} 
		\log \mathcal{T}(\delta,\mathcal{F},\|\cdot\|_n)  \leq    G \log  \mathcal{T}\left( \frac{ \delta}{C_3 G^{3/2 + 3\omega_2}} ,\mathcal{F_{\text{DNN}}},\|\cdot\|_n \right)
	\end{align*}
\end{lemma}

\noindent
\textbf{Proof}:
We first establish the Lipschitz condition. We compute the partial derivative of 
\begin{align*}
	f(y|x)= \sum_{g=1}^G \alpha_g(x) \frac{1}{\sqrt{2 \pi} \sigma_g(x)}  e^{-\frac{(y-\mu_g(x))^2}{2 \sigma_g^2(x)}} \in \mathcal{F}, 
\end{align*}
with respect to $\alpha_g, \mu_g, \sigma_g$. Since $|y|\leq 1$, $0 \leq \alpha_g \leq 1$, $|\mu_g(x)| \leq C_2$, and $C_2^{-1} G^{\omega_2} \leq \sigma_g(x) \leq C_2$, there exists $C'_3>0$, which depends on $A_1,A_2,A_3$ and $A_4$, such that 
\begin{align*}
	\left| \frac{\partial f}{\partial \alpha_g} \right| &=  \frac{1}{\sqrt{2 \pi} \sigma_g(x)}  e^{-\frac{(y-\mu_g(x))^2}{2 \sigma_g^2(x)}} \leq C'_3 G^{\omega_2}\\ 
	\left| \frac{\partial f}{\partial \mu_g} \right| &= \frac{|y-\mu_g(x)|}{\sigma^2_g(x)}  \frac{\alpha_g}{\sqrt{2 \pi} \sigma_g(x)}  e^{-\frac{(y-\mu_g(x))^2}{2 \sigma_g^2(x)}} \leq C'_3 G^{2\omega_2}\\
	\left| \frac{\partial f}{\partial \sigma_g} \right| &=  \Big |\frac{(y-\mu_g(x))^2}{\sigma^3_g(x)} -\frac{1}{\sigma(x)} \Big| \frac{\alpha_g}{\sqrt{2 \pi} \sigma_g(x)}  e^{-\frac{(y-\mu_g(x))^2}{2 \sigma_g^2(x)}}  \leq C'_3 G^{3\omega_2}
\end{align*}
Then, for any two neural network models $f,g \in \mathcal{F}$, there exits a constant $C_3$, which depends on $C'_3$, such that 
\begin{align*}
	|f-g|^2 & \leq  \sum_{j=1}^G C^2_3 G^{2\omega_2} G |h^{(f)}_{\alpha j}(x)-h^{(g)}_{\alpha j}(x)|^2    \\
	|f-g|^2 & \leq  \sum_{j=1}^G C^2_3 G^{4\omega_2} G |h^{(f)}_{\mu j}(x)-h^{(g)}_{\mu j}(x)|^2    \\
	|f-g|^2 & \leq  \sum_{j=1}^G C^2_3 G^{6\omega_2} G |h^{(f)}_{\sigma j}(x)-h^{(g)}_{\sigma j}(x)|^2    
\end{align*}
where $h^{(f)}$ and $h^{(g)}$ are for the neural network $f$ and $g$, respectively, and we utilize the property that the relu, softmax and surplus functions are all 1-Lipschitz. 

Next, we apply Lemma A.6 of \cite{entropy_contraction}, and obtain that, 
\begin{align}
	\mathcal{T}(\delta,\mathcal{F},\|\cdot\|_n)  \leq  \left\{ \mathcal{T}\left( \frac{ \delta}{C_3   G^{3/2+3\omega_2}}  ,\mathcal{F_{\text{DNN}}},\|\cdot\|_n \right) \right\}^G.
\end{align}
Taking the logarithm transformation on both sides yields that
\begin{align}
	\log \mathcal{T}(\delta,\mathcal{F},\|\cdot\|_n)  \leq    G \log \mathcal{T}\left( \frac{ \delta}{C_3   G^{3/2+3\omega_2}} ,\mathcal{F_{\text{DNN}}},\|\cdot\|_n \right).
\end{align}
This completes the proof of Lemma \ref{lemma5}.
\eop
\bigskip

Since we consider the stationary time series in our setting, we cannot apply some traditional empirical inequalities, such as the Bernstein's inequality and Theorem 2.1 of \cite{bartlett2005local}, developed for i.i.d.\ data. The next lemma relaxes the i.i.d.\ constraint and enables the application of those inequalities in the setting of time series with $\beta$-mixing.

\begin{lemma} \label{lemma6}
	Suppose there are a sequence of stationary time series, $U = (X_1,X_2,...,X_T)$, with a common marginal distribution, and each $X_t \in \mathcal{R}^d$, $t=1,\ldots, T$. Suppose the $\beta-$mixing coefficient of the sequence satisfies that $\beta(p) \leq c_1 e^{-c_2p}$ for some $c_1,c_2$. Then there exists a sequence, $U_i^0 = (X^0_{ip+1},X^0_{ip+2},...,X^0_{ip+p})$, for $i \geq 0$, such that
	\begin{enumerate}[(i)]
		\item $U_i^0$ has the same distribution as $U_i = (X_{ip+1},X_{ip+2},...,X_{ip+p})$.
		\item The sequence $\{U^0_{2i}\}_{i \geq 0}$ is i.i.d., and so is  $\{U^0_{2i+1}\}_{i \geq 0}$.
		\item For any $i \geq 0$, $\prob(U_i \neq U_i^0) \leq \beta(p)$
		\item $\prob(|G_T(U)-G_T(U^0)| \neq 0) \leq p^{-1} T \beta(p)$, where $G_T(U) = \sqrt{T}\left\{ \frac{1}{T} \sum_{t=1}^T f(X_{t+1}|X_t) - \mathbb{E}f \right\}$, and $G_T(U^0) = \sqrt{T}\left\{ \frac{1}{T} \sum_{t=1}^T f(X^0_{t+1}|X^0_t) - \mathbb{E}f \right\}$.
	\end{enumerate}
\end{lemma}

\noindent
\textbf{Proof}:
The first three statements can be obtained directly from Lemma 4.1 of \cite{stationary_bound}. For the last statement, we have that, 
\begin{align*}
	\prob\left( |G_T(U)-G_T(U^0)| \neq 0 \right) \leq \prob\left( \frac{2 \|f\|_{\infty}}{\sqrt{T}} \sum_{t=1}^T \mathbf{1}_{\{X_t \neq X^0_t\}} \neq 0 \right)
	\leq \frac{T}{p}\prob\left( U_i \neq U_i^0 \right) \leq \frac{T}{p}\beta(p).
\end{align*}
This completes the proof of Lemma \ref{lemma6}.
\eop
\bigskip

The next lemma presents an uniform bound for the forward and the backward learner across all the choices of $b$. This convergence rate is important for establishing Theorem $\ref{thm4}$ and $\ref{thm5}$.

\begin{lemma} \label{lemma7}
	Suppose the conditions in Theorem \ref{thm4} hold. Then there exists a constant $c_0>1/2$, such that
	\begin{eqnarray*}
		\max_{1\le b\le B}  \int_{x}|\widehat{\varphi}^{(\ell)}(\mu_b|x)-\varphi^*(\mu_b|x)|^2\mathbb{F}(dx)=O_p(T^{-c_0}),\\
		\max_{1\le b\le B} \int_{x} |\widehat{\psi}^{(\ell)}(\nu_b|x)-\psi^*(\nu_b|x)|^2\mathbb{F}(dx)=O_p(T^{-c_0}),
	\end{eqnarray*}
	where $\mathbb{F}$ denotes the cumulative distribution function of $X_{1}$, and $\widehat{\varphi}^{(\ell)}$ and $\widehat{\psi}^{(\ell)}$ are the bounded functions estimated by the mixture density network under Assumption \ref{assump3}.  
\end{lemma}

\noindent
\textbf{Proof}:
By Theorem\ref{mdn_bound}, there exists a constants $c>0$, such that, with probability at least $1-O(T)$,
\begin{align*}
	\left\| \widehat{f}_{X_{t+1}|X_t}^{(\ell)}-f^*_{X_{t+1}|X_{t}} \right\|_2 \leq c d \left\{ G^{-\omega_1}+  G ^{\frac{\gamma+d}{2\gamma}+4\omega_2}T^{-\frac{\gamma}{2\gamma+d}} \log^3 (T G) \right\}
\end{align*}

For the given $X_{(\ell-1) n + 1},...,X_{\ell n}$, for $\widetilde{X}_{t+1}|X_t$ sampled from $\widehat{f}_{X_{t+1}|X_t}^{(\ell)}$, and a given function $h$ bounded by some constant $H_0$, we have that,
\begin{align} \label{transfer}
	\begin{split}
		& \frac{1}{M} \sum_{j=1}^M h(\widetilde{X}^{(j)}_{t+1}|X_t) - E_{f^*} h(X_{t+1}|X_t) \\ 
		= \; & \frac{1}{M} \sum_{j=1}^M h(\widetilde{X}^{(j)}_{t+1}|X_t)-E_{\widehat{f}} h(X_{t+1}|X_t)+E_{\widehat{f}} h(X_{t+1}|X_t) - E_{f^*} h(X_{t+1}|X_t)
	\end{split}
\end{align}
where $E_f$ denotes the expectation by using $f$ as the density function, and $E_{\widehat{f}}$ by using $\widehat{f}$ as the density. Since $|h(\widetilde{X}_{t+1}|X_t)| \leq H_0$, applying the Hoeffding bound, we have that, with probability at least $1-e^{-t^2/(2 M H_0^2)}-C_{10}/M-O(T^{-1})$,
\begin{align} \label{hoeffding}
	\sum_{j=1}^M \left\{ h(\widetilde{X}^{(j)}_{t+1}|X_t) - E_{\widehat{f}}  h(\widetilde{X}_{t+1}|X_t) \right\} \leq t.
\end{align}
Setting $t= M^{1/2} \log M$, and plugging (\ref{hoeffding}) into (\ref{transfer}), we obtain that, with probability as least $1-M^{-1/(2  H_0^2)}-C_{10}/M-O(T^{-1})$,
\begin{align*} 
	\begin{split}
		& \left| \frac{1}{M} \sum_{j=1}^M h(\widetilde{X}^{(j)}_{t+1}|X_t) - E_{f^*} [h(X_{t+1}|X_t)] \right|^2 \\ 
		\leq \; & \left| \frac{1}{M^{1/2}\log M}+E_{\widehat{f}} [h(X_{t+1}|X_t)] - E_{f^*} [h(X_{t+1}|X_t)] \right|^2 \\
		\leq \; & \left| \frac{1}{M^{1/2}\log M}+\int |h(y)| \frac{|\widehat{f}(y|X_t)-f^*(y|X_t)|}{f^*(y|X_t)}f^*(y|X_t) dy \right|^2  
	\end{split}
\end{align*}
Then by Cauchy–Schwarz inequality, we obtain that,
\begin{align*} 
	\begin{split}
		& \left| \frac{1}{M} \sum_{j=1}^M h(\widetilde{X}^{(j)}_{t+1}|X_t) - E_{f^*} [h(X_{t+1}|X_t)] \right|^2 \\
		\leq \; & \frac{2}{M  \log^2 M}+\frac{2}{ C^2} \|\widehat{f}(y|X_t)-f^*(y|X_t)\|^2_2 \int |h(y)|^2 f^*(y|X_t)  dy \\
		\leq \; & \frac{2}{M  \log^2 M}+\frac{2}{ C^2} \|\widehat{f}(y|X_t)-f^*(y|X_t)\|^2_2  {H_0^2}   
	\end{split}
\end{align*}
Taking the expectation of $X$ and $\mathop{\sup}_{h}$ on both sides, and using the result from Theorem \ref{mdn_bound}, we obtain that, with probability as least $1-(\kappa_1 T^{\kappa_2})^{-1/(2  H_0^2)} - C_{10}/(\kappa_1 T^{\kappa_2}) - O(T^{-1})$,
\begin{align*} 
	\begin{split}
		\mathop{\sup}_{h} \mathbb{E}_X \Big | \frac{1}{M} \sum_{j=1}^M h(\widetilde{X}^{(j)}_{t+1}|X_t) - E_{f^*} [h(X_{t+1}|X_t)] \Big |^2 
		\leq \frac{2}{M \log^2 M}+ \mathop{\sup}_{h} \frac{2 H_0^2}{C^2} \|\widehat{f}-f^*\|^2_2 
		= o(T^{-1/2}),     
	\end{split}
\end{align*}
where the last inequality holds due to the conditions of Theorem \ref{thm4}, where $M = \kappa_1 T^{\kappa_2}$ for some $\kappa_1 >0$, $\kappa_2 \geq 1/2$, and the convergence rate for $\widehat{f}_{X_{t+1}|X_t}^{(\ell)}$ is $o(T^{-1/4})$. Setting the $h$ function as the conditional characteristic functions $\widehat{\varphi}^{(\ell)}(\mu_b|x)$ and $\widehat{\psi}^{(\ell)}(\mu_b|x)$, both of which are upper bounded, then there exists a constant $c_0 \geq 1/2$, such that, with probability as least $1-(\kappa_1 T^{\kappa_2})^{-1/(2  H_0^2)} - C_{10}/(\kappa_1 T^{\kappa_2}) - O(T^{-1})$, 
\begin{eqnarray*}
	\max_{1\le b\le B}  \int_{x}|\widehat{\varphi}^{(\ell)}(\mu_b|x)-\varphi^*(\mu_b|x)|^2\mathbb{F}(dx) & = & O_p(T^{-c_0}),\\
	\max_{1\le b\le B} \int_{x} |\widehat{\psi}^{(\ell)}(\nu_b|x)-\psi^*(\nu_b|x)|^2\mathbb{F}(dx) & = & O_p(T^{-c_0}).
\end{eqnarray*}
This completes the proof of Lemma \ref{lemma7}.
\eop

\change{
	\subsection{Additional Numerical Results}
	\label{sec:add_numericals}
	
	\subsubsection{Sensitivity analysis}
	\label{sec:appex_sensi}
	
	We conduct a sensitivity analysis to investigate the performance of the proposed test under different choices of the hyper-parameters, including the number of data chunks $L$, the number of pseudo samples $M$ from the forward and backward generators, the largest number of lags $Q$ considered in the test, as well as the number of hidden units $U$. Each time we vary one hyper-parameter, while we fix the rest at the default values. We consider Model 1, i.e., the VAR model, in Section \ref{sec:sim}, with $T = 500$. Figure \ref{fig:sensitivity} reports the percentage of times the null hypothesis is rejected out of 500 data replications at the significance level $\alpha=0.05$. It is seen that the test results are fairly stable across different values of the varying hyper-parameter, except that the power increases slightly as $L$ increases. Overall, we conclude that the proposed test is not overly sensitive to the choice of these parameters, as long as they are in a reasonable range.
	
	\begin{figure}[t!]
		\centering
		\includegraphics[width=1\linewidth]{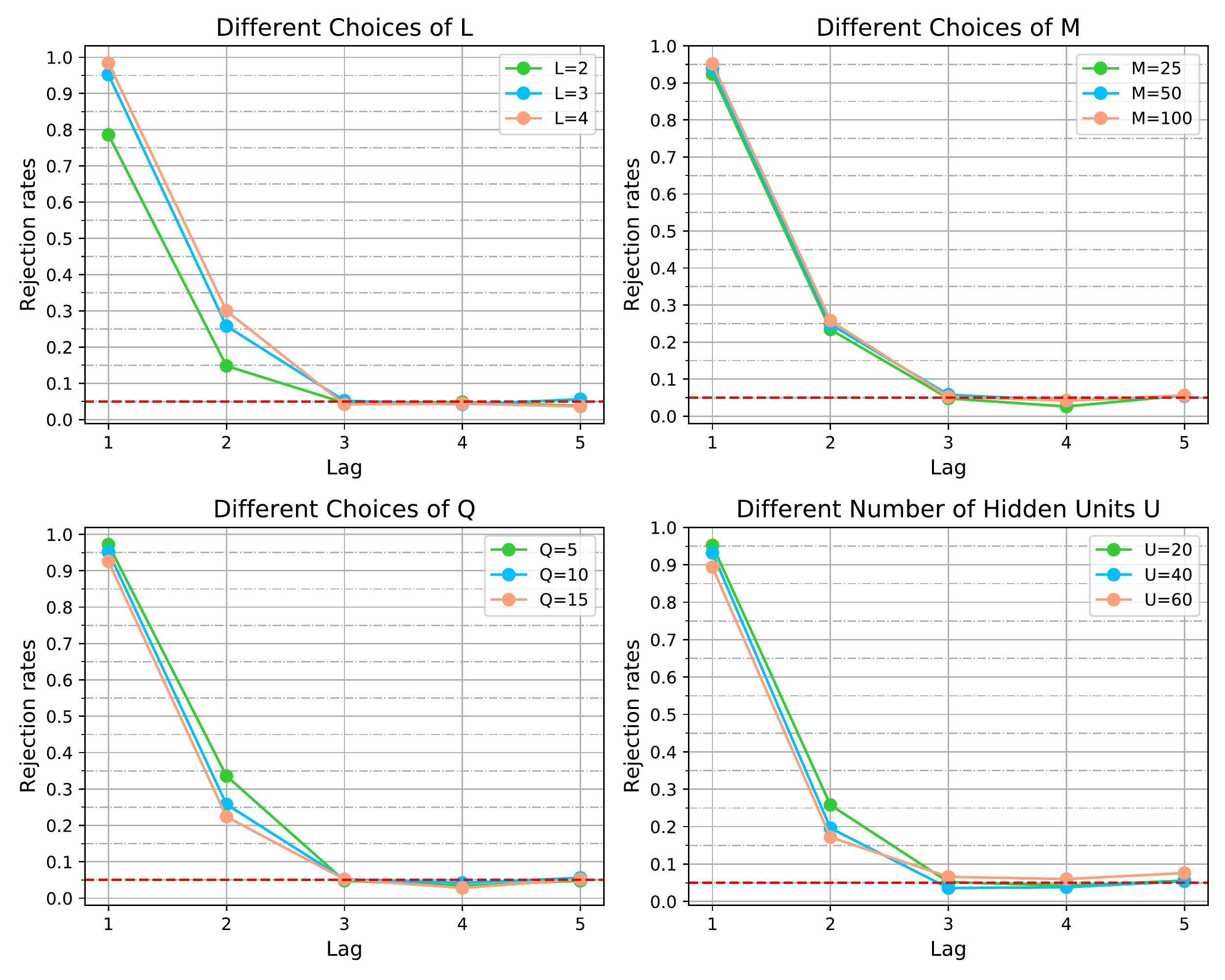}
		\caption{Sensitivity analysis: the percentage of times the null hypothesis is rejected out of 500 data replications at the significance level $\alpha=0.05$ under difference values of the hyper-parameters $L$, $M$, $Q$, and $U$.}
		\label{fig:sensitivity}
	\end{figure}

	\subsubsection{Empirical evidence of Theorem \ref{mdn_bound}} 
	\label{sec:appex_simbound}
	
	Theorem \ref{mdn_bound} establishes the error bound of the MDN estimator, and is critical for the consistency of the proposed test. We carry out an additional simulation to further study the result of Theorem \ref{mdn_bound} empirically. Specifically, we consider a simple version of Model 1, i.e., a one-dimensional AR(0.5) time series model, in Section \ref{sec:sim}. Since AR(0.5) is Gaussian, MDN can exactly approximate the true density function of the data generating process, so the approximation error is constant zero. As a result, $\|\widehat{f}_{X_{t+1}|X_t}-f^*_{X_{t+1}|X_t}\|_2$ only contains the estimation error. According to Theorem \ref{mdn_bound}, $\|\widehat{f}_{X_{t+1}|X_t}-f^*_{X_{t+1}|X_t}\|_2$ should converge to zero at the rate of $O(T^{-1/2})$ when the number of mixture components $G$ is fixed, and at the rate of $O(G^{1/2})$ when the time series length $T$ is fixed. So we fix $G=1$ and vary $T$ in the first case, whereas we fix $T=2000$ and vary $G$ in the second case, and we record the corresponding estimation error. Figure \ref{fig:sim_bound} reports the average estimation error across 50 data replications when $T$ or $G$ varies, all in the log scale. It is seen that the slopes from the empirical results agree well with the theoretical values. 
	
	\begin{figure}[t!]
		\centering
		\begin{subfigure}[b]{0.47\textwidth}
			\centering
			\includegraphics[width=1\linewidth]{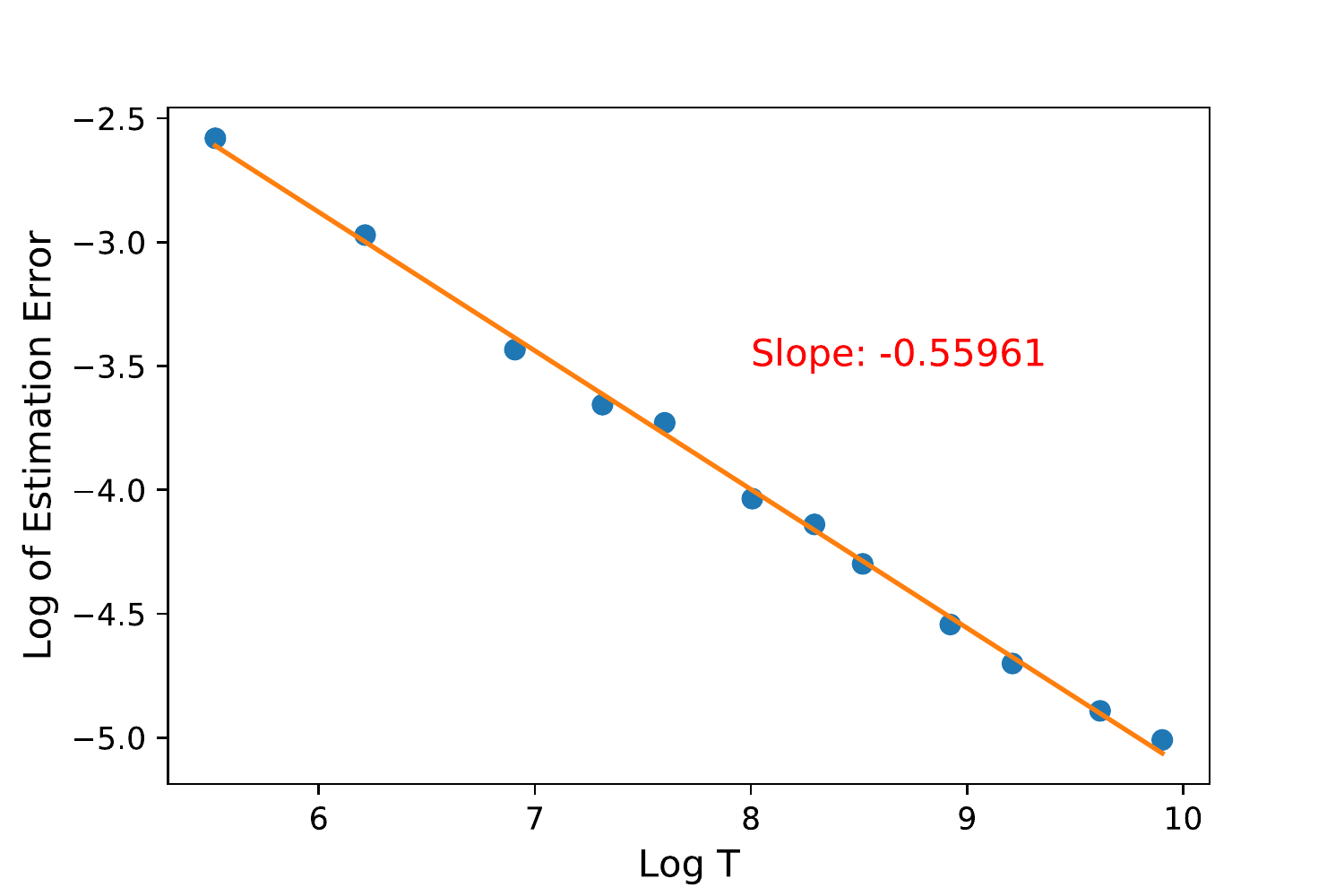}
		\end{subfigure}
		\hfill
		\begin{subfigure}[b]{0.47\textwidth}
			\centering
			\includegraphics[width=1\linewidth]{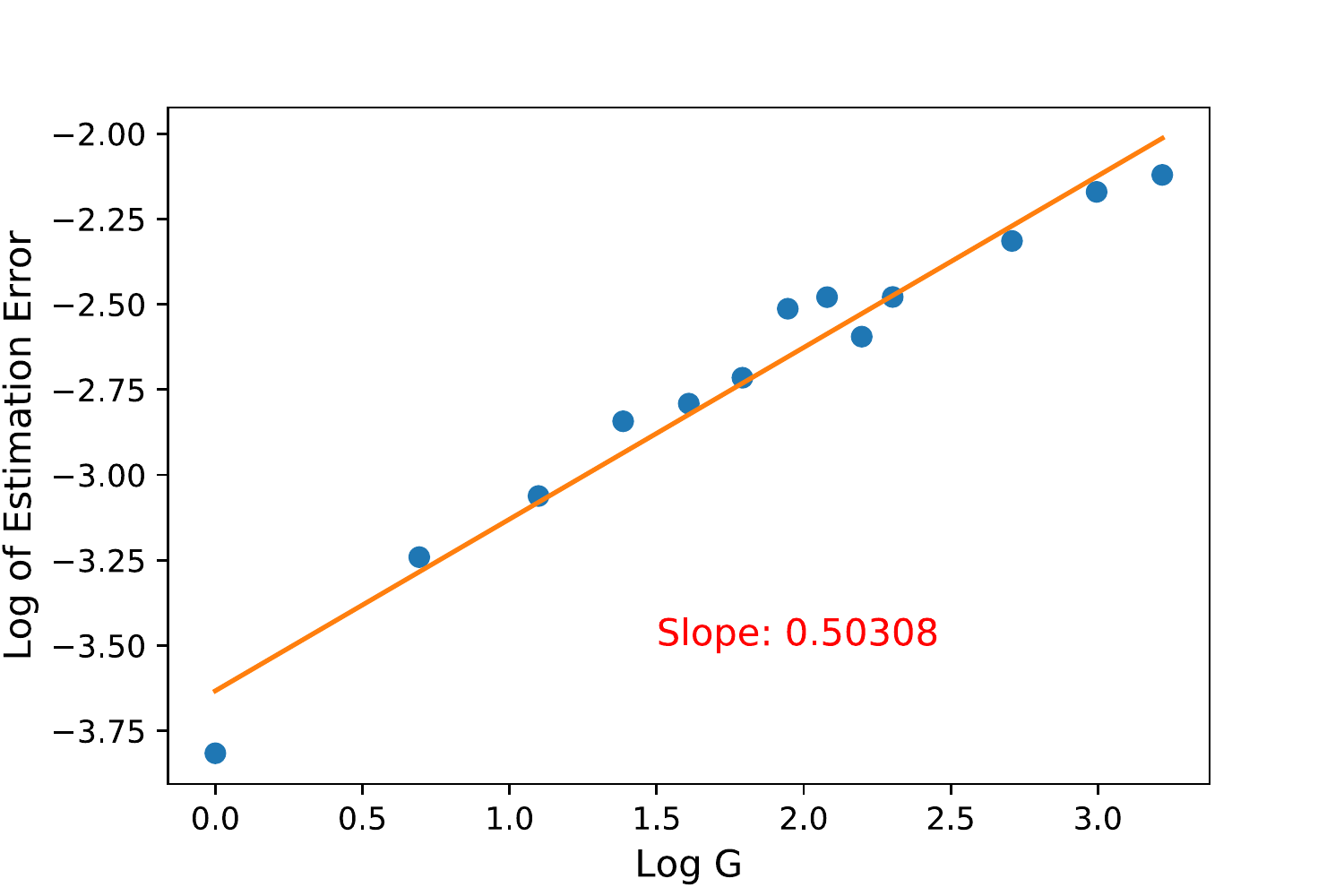}
		\end{subfigure}
		\caption{Average estimation error $\|\widehat{f}_{X_{t+1}|X_t}-f^*_{X_{t+1}|X_t}\|_2$ across 50 data replications, as a function of the time series length $T$, or the number of mixture components $G$, all in the log scale.}
		\label{fig:sim_bound}
	\end{figure}

	\subsubsection{Contributing components for the proposed test}
	\label{sec:appex_contrib}
	
	There are two key components in our proposed test: the MDN learner, and the doubly robust test statistic. We carry out additional simulations to evaluate the contributions of these components. 
	
	First, we compare the MDN learner with the local polynomial regression, in terms of the empirical convergence rate of the estimation error. We adopt Model 1, the VAR model, of Section \ref{sec:sim}, with $d=3$. Figure \ref{fig:sim_bound_new} reports the average estimation error $\|\widehat{f}_{X_{t+1}|X_t}-f^*_{X_{t+1}|X_t}\|_2$ across 50 data replications as the time series length $T$ varies. The left panel is based on the MDN learner, and the right panel is based the local polynomial regression. It is seen that the convergence rate of the MDN is much faster than at of the local polynomial regression. 
	
	\begin{figure}[t!]
		\centering
		\begin{subfigure}[b]{0.47\textwidth}
			\centering
			\includegraphics[width=1\linewidth]{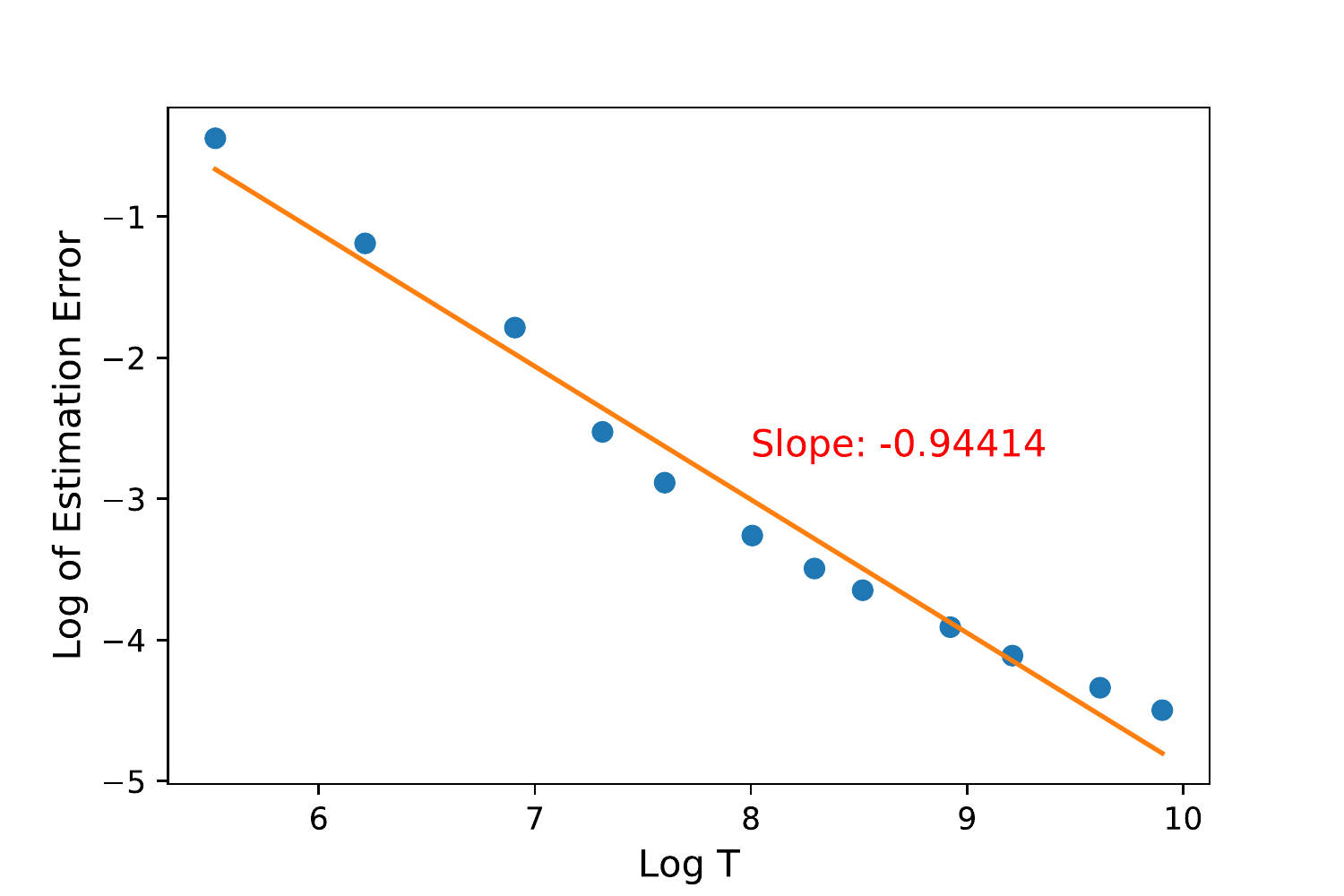}
		\end{subfigure}
		\hfill
		\begin{subfigure}[b]{0.47\textwidth}
			\centering
			\includegraphics[width=1\linewidth]{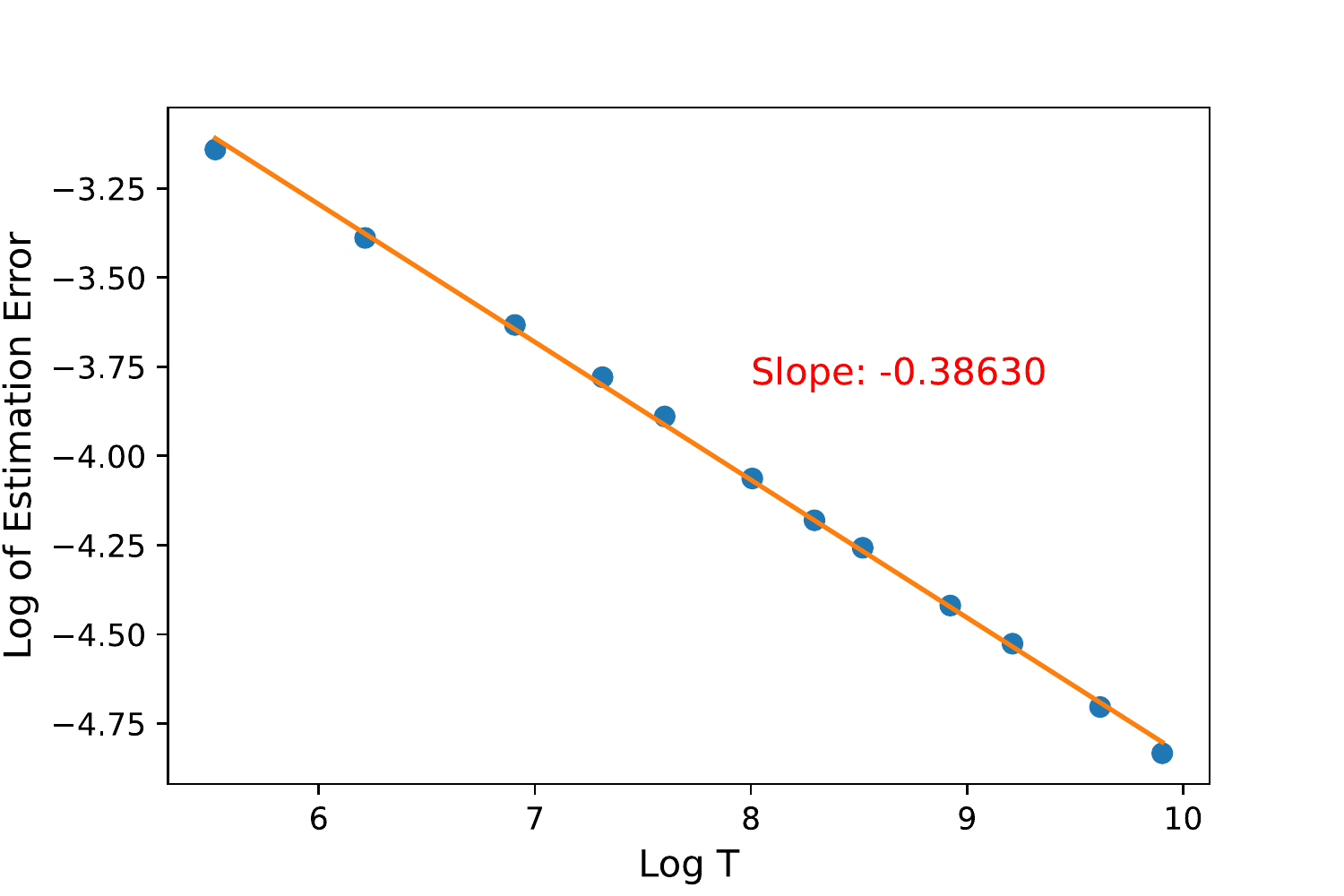}
		\end{subfigure}
		\caption{Average estimation error $\|\widehat{f}_{X_{t+1}|X_t}-f^*_{X_{t+1}|X_t}\|_2$ across 50 data replications, as a function of the time series length $T$, all in the log scale. The left panel is based on the MDN learner, and the right panel is based the local polynomial regression.}
		\label{fig:sim_bound_new}
	\end{figure}
	
	Next, we compare our doubly robust test statistic in \eqref{eqn:S.qmunu} with the test statistic in \eqref{eqn:Stilde.qmunu} that is not doubly robust, but we use the MDN learner in both cases. We again adopt Model 1, the VAR model, of Section \ref{sec:sim}, with $d=3$. Table \ref{tab:double_robust} reports the percentage of times out of 500 data replications when the null hypothesis is rejected under the significance level $\alpha=0.05$, with the varying time series length $T = \{500, 1000, 1500\}$. The true order of the Markov model is $K = 3$. It is seen that the test statistic in \eqref{eqn:Stilde.qmunu} works well in this example as well. However, as we discuss in Section \ref{sec:test-stat}, the test statistic in \eqref{eqn:Stilde.qmunu} does not have the desirable theoretical guarantee, since the bias of this test statistic has the same order of magnitude as that of the estimator of conditional characteristic function. By contrast, the doubly robust test statistic can decay to zero at a faster rate than the convergence rate of the individual estimator of CCF. 
	
	\begin{table}[b!]
		\centering
		\caption{Percentage of times out of 500 data replications when the null hypothesis is rejected under the significance level $\alpha = 0.05$. The true order of the Markov model is $K = 3$. Two test statistics \eqref{eqn:S.qmunu} and \eqref{eqn:Stilde.qmunu} are compared, while the MDN learner is used in both cases.} 
		\label{tab:double_robust}
		\begin{tabular}{|c|c|c|c|c|c|c|} \hline
			Order $k$ & Test statistic & 1  & 2 & 3 & 4 & 5\\ \hline
			T = 500 & \eqref{eqn:S.qmunu} & 0.952 & 0.258 & 0.052 & 0.042 & 0.056 \\ \cline{2-7}
			& \eqref{eqn:Stilde.qmunu} & 0.952 & 0.360 & 0.034 & 0.036 & 0.032 \\ \hline
			T = 1000 & \eqref{eqn:S.qmunu} & 1.000 & 0.856 & 0.042 & 0.044 & 0.044 \\ \cline{2-7}
			& \eqref{eqn:Stilde.qmunu} & 1.000 & 0.928 & 0.064 & 0.050 & 0.046 \\ \hline
			T = 1500 & \eqref{eqn:S.qmunu} & 1.000 & 0.992 & 0.060 & 0.058 & 0.048 \\ \cline{2-7} 
			& \eqref{eqn:Stilde.qmunu} & 1.000 & 1.000 & 0.054 & 0.042 & 0.042 \\ \hline
		\end{tabular}
	\end{table}
}

\end{document}